\documentclass[conference]{IEEEtran}

\PassOptionsToPackage{bookmarks={false}}{hyperref}
\usepackage{multirow}
\usepackage[table]{xcolor}
\usepackage{cite}
\usepackage{siunitx}
\usepackage[pdftex]{graphicx}
\graphicspath{{../pdf/}{../jpeg/}}
\DeclareGraphicsExtensions{.pdf,.jpeg,.png}
\usepackage{algorithmic}
\usepackage{array}
\usepackage{floatrow}
\usepackage{fixltx2e}
\usepackage[table]{xcolor}
\usepackage{stfloats}
\usepackage{url}
\usepackage{times}
\usepackage{color}
\usepackage{verbatim}
\usepackage{hyperref}
\usepackage{subfiles}
\usepackage{caption}
\usepackage{subcaption}
\usepackage{lipsum}
\usepackage{xr}
\usepackage{cite}
\usepackage{export}
\usepackage{nameref}
\usepackage{zref-xr}
\usepackage{adjustbox}
\usepackage{amsmath, amsthm, amssymb}
\usepackage[ansinew]{inputenc}
\usepackage{mathtools}
\usepackage{esvect}
\usepackage{layouts}
\usepackage{xspace}	
\usepackage{siunitx}
\usepackage{floatrow}
\usepackage{soul}
\usepackage{blindtext}
\usepackage{tabularx}
\usepackage{etoolbox}
\usepackage[font=small,skip=0pt]{caption}
\usepackage{authblk}

\makeatletter
\patchcmd{\@classz}
  {\CT@row@color}
  {\oldCT@column@color}
  {}
  {}
\patchcmd{\@classz}
  {\CT@column@color}
  {\CT@row@color}
  {}
  {}
\patchcmd{\@classz}
  {\oldCT@column@color}
  {\CT@column@color}
  {}
  {}
\makeatother

\usepackage{setspace}
\zxrsetup{toltxlabel}
\zexternaldocument*{appendix}
\hypersetup{
    colorlinks=true,
    linkcolor=red,
    filecolor=magenta,      
    urlcolor=cyan,
}

\renewcommand{\baselinestretch}{.96}
\definecolor{fxnote}{rgb}{0.8000,0.0000,0.0000}

\renewcommand{\paragraph}[1]{\textbf{#1}}

\newcommand{\bench}[1]{\textit{#1}}
\newcommand{\solo}[0]{3DR~Solo\xspace}
\newcommand{\red}[1]{\textcolor{red}{#1}}

\newcommand{\Sec}[1]{Section~\ref{#1}}
\newcommand{\Fig}[1]{Figure~\ref{#1}}

\IEEEoverridecommandlockouts
\date{}
\title{MAVKit: A Platform for Hardware-in-the-Loop Micro-Aerial Vehicle (MAV) Benchmarking}

\title{MAVKit: A Closed Loop Hardware and Software Platform for Micro-Aerial Vehicle Benchmarking}

\title{The Role of Compute in Aerial Agents:\\Using a Closed Loop Hardware and Software Platform for Micro-Aerial Vehicle Benchmarking}

\title{MAVBench: Micro Aerial Vehicle Benchmarking}

\author[$\dagger$]{Behzad Boroujerdian*\thanks{* These two authors contributed equally.}}
\author[$\dagger$]{Hasan Genc*}
\author[$\ddagger$]{Srivatsan Krishnan}
\author[$\dagger$]{Wenzhi Cui} 
\author[$\mp$]{Aleksandra Faust}
\author[$\dagger\ddagger\S$]{Vijay Janapa Reddi}

\affil[ ]{ }
\affil[$\dagger$]{The University of Texas at Austin}
\affil[$\ddagger$]{Harvard University}
\affil[$\mp$]{Google Brain}
\affil[$\S$]{Google}
\affil[ ]{ }
\affil[ ]{\texttt{\url{https://github.com/harvard-edge/MAVBench}}}

\begin{document}

\maketitle

\thispagestyle{empty}

\begin{abstract}
Unmanned Aerial Vehicles (UAVs) are getting closer to becoming ubiquitous in everyday life. Among them, Micro Aerial Vehicles (MAVs) have seen an outburst of attention recently, specifically in the area with a demand for autonomy. A key challenge standing in the way of making MAVs autonomous is that researchers lack the comprehensive understanding of how performance, power, and computational bottlenecks affect MAV applications. MAVs must operate under a stringent power budget, which severely limits their flight endurance time. As such, there is a need for new tools, benchmarks, and methodologies to foster the systematic development of autonomous MAVs. In this paper, we introduce the ``MAVBench'' framework which consists of a closed-loop simulator and an end-to-end application benchmark suite. A closed-loop simulation platform is needed to probe and understand the intra-system (application data flow) and inter-system (system and environment) interactions in MAV applications to pinpoint bottlenecks and identify opportunities for hardware and software co-design and optimization. In addition to the simulator, MAVBench provides a benchmark suite, the first of its kind, consisting of a variety of MAV applications designed to enable computer architects to perform characterization and develop future aerial computing systems. Using our open source, end-to-end experimental platform, we uncover a hidden, and thus far unexpected compute to total system energy relationship in MAVs. Furthermore, we explore the role of compute by presenting three case studies targeting performance, energy and reliability. These studies confirm that an efficient system design can improve MAV's battery consumption by up to 1.8X. 

\end{abstract}


\section{Introduction}

Unmanned aerial vehicles (a.k.a drones) are becoming an important part of our technological society. With myriad use cases, such as in sports photography~\cite{drone-sports1}, surveillance~\cite{drone-surveillance}, disaster management, search and rescue~\cite{disaster-drone,nepal-earthquake-drone}, transportation and package delivery~\cite{Amazondel:online,CSD-Amazon-Drone-patents,drone-package-delivery-google}, and more, these unmanned aerial vehicles are on the cusp of demonstrating their full potential. 

\begin{figure}[t!]
  \includegraphics[trim=0 0 0 -10, clip, width=1.0\columnwidth]{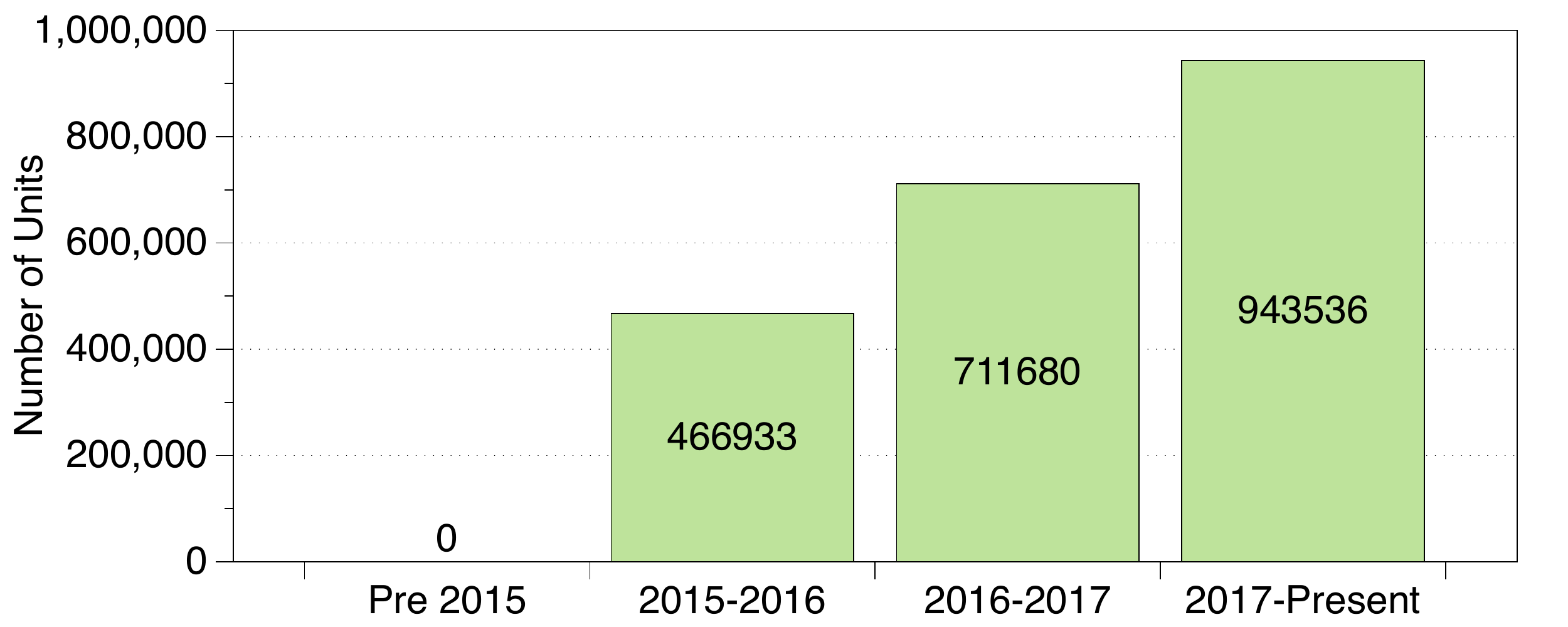}
  \caption{Rapidly growing interest in UAVs. Data mined from FAA vehicle registration. The number of FAA registrations increased by 2X over the past two years, and it is rapidly growing. The FAA projects that by 2021 the number will exceed 4M units~\cite{FAA:online}.}
  \label{fig:faa-registrations}
\end{figure}

Hence, drones are rapidly increasing in number. Between 2015, when the U.S. Federal Aviation Administration (FAA) first required every owner to register their drone, and 2017, the number of drones has grown by over 200\%. At the time of writing, the FAA indicates that there are over 900,000 drones registered with the FAA drone registry database (\Fig{fig:faa-registrations}). By 2021, the FAA expects this number will exceed 4 million units~\cite{faa-2021}. Such an upward trend can be explained by the new opportunities that unmanned aerial vehicles are enabling.


The growth and significance of this emerging domain of autonomous agents call for architects attention. Challenges such as low endurance (how long the drone can last in the air) and small battery capacities for drones demand hardware and system architects' attention. The limited on-board energy budget manifests itself in the limited endurance and range of drones. This can be seen in various off-the-shelf commercial drones where endurance is typically less than 20 minutes, and flight range is about 15 miles~\cite{CSD-Amazon-Drone-patents}. To practically deploy drones, both their endurance and range must be improved. 


In this paper, we investigate and show the role of computing given the endurance and range challenges. For example, we show how a powerful compute subsystem can be deployed to mitigate the problem of limited endurance. The drone's compute subsystem dictates how fast a drone can maneuver, fly, and efficiently finish its mission. Hence, a computing subsystem that takes a long time to do path planning while the drone is hovering in the air, results in the inefficient consumption of energy. Furthermore, a more powerful compute subsystem can lead to more intelligent decision making (e.g., shorter paths to take). It is important to note that enabling intelligence on drones is challenging because of the computational power, size, weight, and cooling limitations.


To enable research and investigation, the foremost challenge to address is the lack of systematic benchmarks and infrastructure for research. To address this shortcoming, we introduce MAVBench, the first of its kind, a platform for the holistic evaluation of aerial agents, involving a closed-loop simulation framework and a benchmark suite. MAVBench facilitates the integrated study of performance and energy efficiency of not only the compute subsystem in isolation but also the compute subsystem's dynamic and runtime interactions with the simulated micro-aerial vehicle (MAV) as a whole.


The goals of MAVBench, which builds on top of AirSim~\cite{Airsim_paper}, are to faithfully capture all the interactions a real MAV encounters in the field and to ensure reproducible runs across experiments, starting from the software layers down to the hardware layers. Our simulation setup uses a hardware-in-the-loop configuration that can enable hardware and software architects to perform co-design studies to optimize system performance by considering the entire vertical application stack running on top of it, which includes the Robotics Operating System (ROS) and complete applications. It reports a variety of quality-of-flight (QoF) metrics, such as the performance, power consumption, and trajectory statistics of the drone.

MAVBench provides an application suite covering a variety of popular applications of micro aerial vehicles. We constructed five distinct and widely used applications: \bench{Scanning}, \bench{Package Delivery}, \bench{Aerial Photography}, \bench{3D Mapping} and \bench{Search and Rescue.} MAVBench applications are whole, comprised of holistic end-to-end application dataflows found in a typical real world drone application. These applications' dataflows are comprised of several state-of-the-art computational kernels, such as object detection~\cite{yolo16,hog}, occupancy map generation~\cite{octomap}, motion planning~\cite{ompl}, localization~\cite{orbslam2,vins-mono}, which we integrated together to create complete applications. MAVBench has the ability to ``plug and play'' different computational kernels to study trade-offs for the same task.


MAVBench enables us to investigate, understand and quantify the power and performance demands of typical MAV applications from the underlying compute subsystem. More specifically, it allows us to answer a very fundamental question: \emph{what is the role of computing in autonomous MAVs?} 

We quantitatively demonstrate via simulations and measurements that compute has a significant impact on how efficiently the drone uses its energy while flying, affecting mission times and overall energy consumption. An off-the-shelf MAV, such as the DJI~Matrice~\cite{dji-matrice} or \solo~\cite{solo3DR}, consumes between 300~W to 400~W for its rotors, as much power as a typical data center server, with an average endurance that is typically less than 20 minutes. Compute on average consumes less than 5\% of that total system power.  A state-of-the-art compute platform like the Nvidia TX2 consumes about 10~W on average. However, the compute performed onboard by the TX2 can affect flight mission time by as much as 2X. We conduct frequency and core scaling experiments on the TX2 to demonstrate how the \emph{compute performance affects the drone's velocity, which in turns impacts its mission time, and consequentially the drone's total system energy consumption.}

Taking advantage of our platform, we provide three case studies targeting performance, energy and reliability. The performance case study examines a sensor-cloud architecture for drones where the computation is distributed across the edge and the cloud. Such an architecture shows a reduction in the drone's overall mission time by as much as 50\% when the cloud support is enabled. The energy case study, targets Octomap~\cite{octomap}, a computationally intensive kernel that is at the heart of some of the MAVBench applications and demonstrates how approximations in its internal representation, and thus compute, can enable safe flight while improving overall energy consumption. Last by not least, our reliability case study investigates the impact of sensor noise on the performance of one of our applications, namely package delivery, showing a performance degradation of up to 90\% while in presence of high depth image noise. All three case studies demonstrate the potential ways in which MAVBench can be used for architecture and systems research. In general, the platform allows a variety of hardware and software (co-)design studies.


In summary, we make the following contributions:

\begin{itemize}
    \item We present a closed-loop simulation framework that enables hardware and software architects to perform performance and power optimization studies that are relevant to computer system design and architecture.
    \item We introduce an end-to-end benchmark suite, comprised of several workloads and their corresponding state-of-the-art kernels. These workloads represent popular real-world use cases of MAVs.
    \item We uncover the role of computing and its relationship with flight time and endurance for unmanned MAVs.
    \item We present case studies to concretely explore and emphasize compute's impact on performance, energy and reliability of MAV systems.

\end{itemize}

The rest of the paper is organized as follows. 
~\Sec{sec:background} provides basic background about Micro Aerial vehicles, the reasons for their prominence amongst UAVs, and the challenges system designers face.  \Sec{sec:simulation} describes the MAVBench closed-loop simulation platform. \Sec{sec:mavbench} introduces the MAVBench benchmark suite and describes the computational kernels and full-system stack it implements. \Sec{sec:char} uses MAVBench to uncover an interesting relationship between compute and the MAV's total system energy consumption. \Sec{sec:case-study} presents three case studies further examining the impact of compute stack on performance, energy and reliability of MAV systems. \Sec{sec:related} discusses related work, and \Sec{sec:conclusion} summarizes and concludes the paper.
       
\section{Background}
\label{sec:background}

We provide a brief background about unmanned aerial vehicles (UAVs). 
In \Sec{sec:mav}, we discuss the most ubiquitous and growing segment of UAVs, i.e., Micro Aerial Vehicles (MAVs) and its different flight-wing types. In  \Sec{sec:constraints}, we present the overall system level constraints facing MAVs.

\subsection{Micro Aerial Vehicles (MAV)}
\label{sec:mav}
There is no single established standard to categorize all drones. However, typically, a UAV is classified as a ``Micro UAV'' if its weight is less than 2~\si{\kilo\gram}, and it operates within a radius of 5~\si{\kilo\meter}. MAVs' small size increases their accessibility and affordability by shortening their ``development and deployment time,'' as well as reducing the cost of ``prototyping and manufacturing''~\cite{Zhang2017207}. Furthermore, their small size coupled with their ability to move flexibly empowers them with the agility and maneuverability necessary for various applications, such as sports photography, indoor mapping, surveillance, etc.

MAVs come in different shapes and sizes. A key distinction is their wing type. On one end of the spectrum, MAVs have fixed wings. On the other end, MAVs have rotor wings. Fixed wing MAVs, as their names suggest, have fixed winged airframe. Their wing structure is deployed for taking-off, navigation, and landing. These MAVs typically require (small) runways for taking-off and landing. Due to the aerodynamics of their wings, they are capable of gliding in the air, which improves their ``endurance'' (i.e., how long they last in the air).

Rotor wing MAVs are becoming the dominant kind. They take off vertically, land vertically, and move with more agility compared to their fixed-wing counterparts. They do not require constant forward airflow movement over their wings from external sources since they generate their own thrust using rotors. Such capabilities enhance their benefits in constrained environments, especially indoors, such as in buildings and warehouses where there are many tight spaces and corners.

Although MAVs enjoy the aforementioned advantages, their complex mechanical (rotors/payload etc) and electrical subsystem (battery, processors) limit their endurance, and as such present unique challenges for system architects and engineers.

\begin{figure}[t!]
\centering
    \begin{subfigure}{.49\columnwidth}
    \centering
    \includegraphics[trim=0 0 0 0, clip, width=0.9\columnwidth]{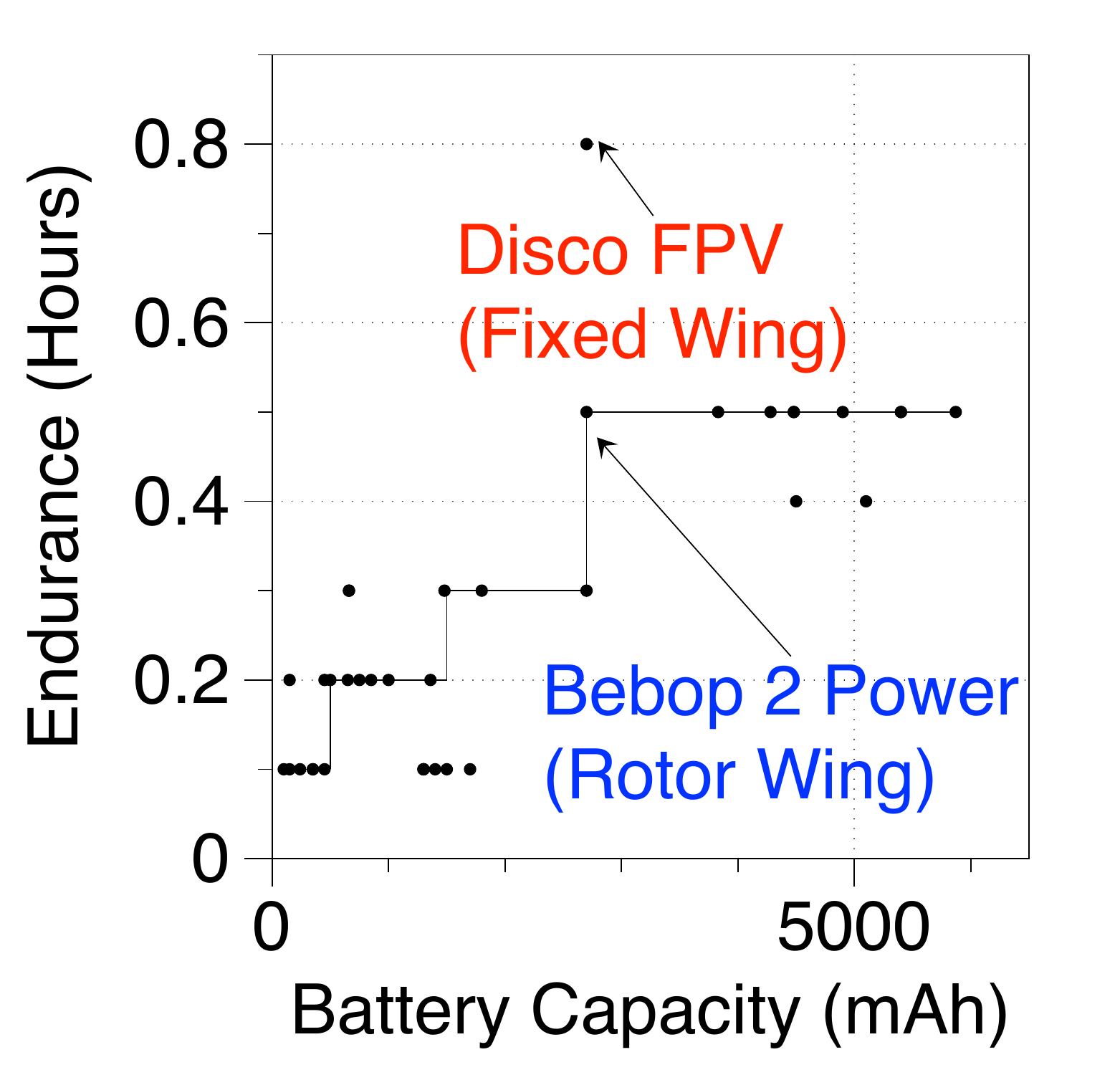}
   \vspace{-5pt}
   \caption{Flight endurance time plotted against total battery capacity.}
    \label{fig:battery_capacity_vs_endurance}
    \end{subfigure}
    \hfill
    \begin{subfigure}{.49\columnwidth}
    \centering	
    \includegraphics[trim=0 0 0 0, clip, width=0.9\columnwidth]{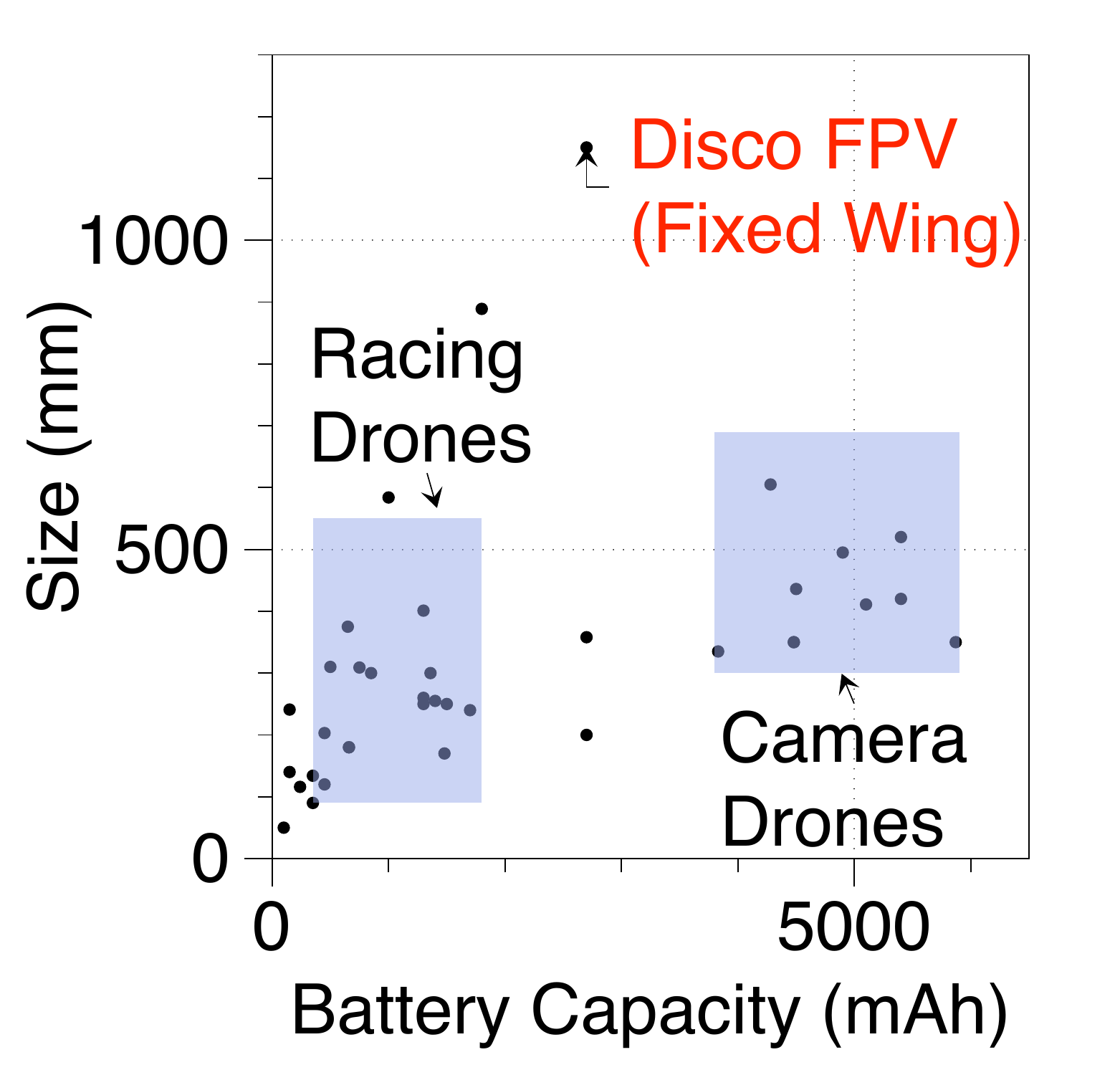}
    \vspace{-5pt}
    \caption{Drone size plotted against total battery capacity.}
    \label{fig:battery_capacity_vs_size}
    \end{subfigure}
\caption{MAVs based on battery capacity and size. Endurance is important for MAVs to be useful in the real-world. However, their small size limits the amount of on-board battery capacity.}
\label{fig:tradeoff}
\end{figure}

\subsection{MAV Constraints}
\label{sec:constraints}

MAVs are tightly constrained on their resources. These constraints typically have to do with the limitations in the mechanical subsystems (rotors/payload size etc.) and the electrical subsystem (battery/processor etc.). For example, in package delivery, the payload size (i.e., the package) affects the mechanical subsystem, requiring more thrust from the rotors and this, in turn, affects the electrical subsystem by demanding more energy from the battery source. Comprehending these constraints is crucial to understand how to optimize the system. The biggest of the constraints as they relate to computer system design are performance and energy.


\paragraph{Performance Constraints:} MAVs are required to meet various real-time constraints. For example, a drone flying at a high speed looking for an object requires fast detection kernels. Such a task is challenging in nature for large-sized drones that are capable of carrying high-end computing systems, and they are virtually impossible on smaller sized MAVs. Hence, the stringent real-time requirements dictate the type of compute engines that can be put on these MAVs.

\paragraph{Energy Constraints:} The amount of battery capacity on board plays an important role in the type of applications MAVs can perform. Battery capacity has a direct correlation with the endurance of these vehicles. To understand this relationship, we show the most popular MAVs available in the market and compare their battery capacity to their endurance. As \Fig{fig:battery_capacity_vs_endurance} shows, higher battery capacity translates to higher endurance. We see a step function trend, i.e., for classes of MAVs that has similar battery capacity, they have similar endurance. On top of this observation, we also see that for the same battery capacity, a fixed wing has longer endurance compared to rotor based MAVs. For instance, in \Fig{fig:battery_capacity_vs_endurance}, we see that the Disco FPV (Fixed wing) has higher endurance compared to the Bebop 2 Power (Rotor wing) even though they have a similar amount of battery capacity.
Note that the size of MAV also has a correlation with battery capacity as shown in \Fig{fig:battery_capacity_vs_size}. 

\section{Closed-loop Simulation}
\label{sec:simulation}
\setstcolor{red}

We discuss a closed-loop simulation environment for simulating and studying MAVs. In \Sec{sec:components}, we describe the components of a MAV, how they interact and why the intra- and inter-system interactions are important to capture. In \Sec{sec:setup}, we show how our setup captures these components and their interactions in a closed-loop setup. In \Sec{sec:energy}, we describe how the simulator models energy consumption, in addition to the functional and performance data described in the earlier section. In \Sec{sec:knobs}, we describe the knobs that our simulator supports to enable exploratory studies, and in \Sec{sec:accuracy}, we describe the limitations of our current setup and opportunities for future enhancements. 
 
\begin{figure}[t!]
\centering
\includegraphics[trim=0 0 0 -25, clip, width=0.9\linewidth]{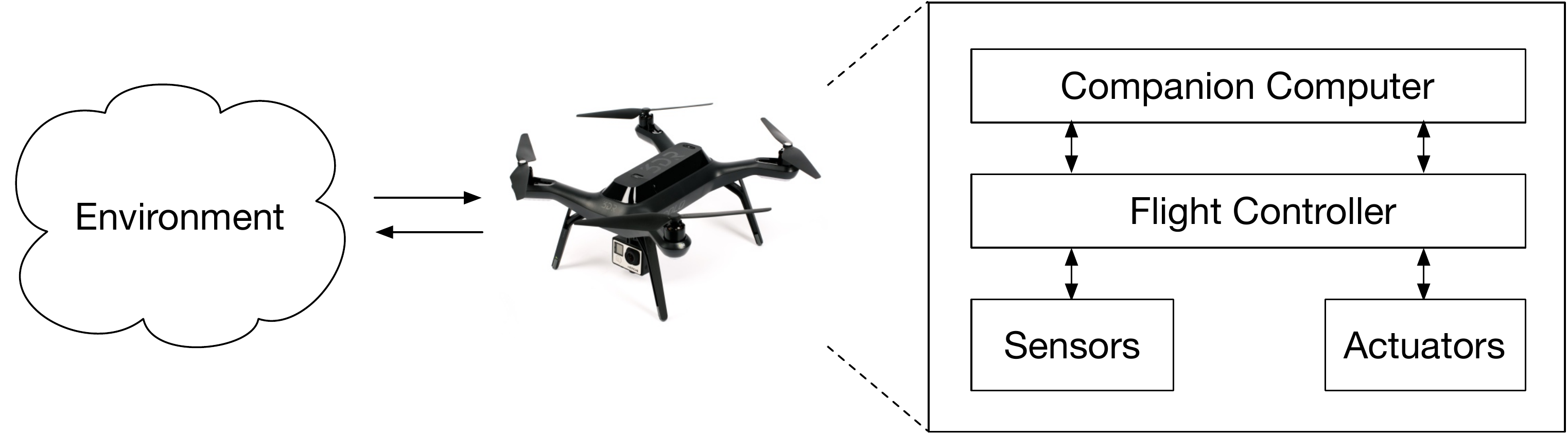}
\caption{Closed-loop data flow in a MAV. Information flows from sensors collecting environment data into the MAV's compute system, down into the actuators and back to the environment.}
\label{fig:Aerial_agent_data_flow}
\end{figure}

\subsection{MAV System Components Dissected}
\label{sec:components}

Closed-loop operation is an integral component of autonomous MAVs. In such systems, the data flows in a (closed) loop, starting from the environment, going through the MAV and back to the environment as shown in \Fig{fig:Aerial_agent_data_flow}. The process involves sensing the environment (Sensors), interpreting it and making decisions (Compute), and finally navigating within or modifying the environment (Actuators) in a loop. 

\paragraph{Sensors:} Sensors are responsible for capturing the state associated with the agent and its surrounding environment. 
To enable intelligent flights, MAVs must be equipped with a rich set of sensors capable of gathering various forms of data such as depth, position, and orientation. For example, \mbox{RGB-D} cameras can be utilized for determining obstacle distances and positions. The number and the type of sensors are highly dependent on the workload requirements and the compute capability of on board processors for their interpretations.

\paragraph{Flight Controller (Compute):} The flight controller (FC) is an autopilot system responsible for the MAV's stabilization and conversion of high-level to low-level actuation commands. 
While they themselves come with basic sensors, such as gyroscopes and accelerometers, they are also used as a hub for incoming data from other sensors such as GPS, sonar, etc. For command conversions, FCs take high-level flight commands such as``take-off" and lower them to a series of instructions understandable by actuators such as rotors. FCs use light-weight processors such as the ARM Cortex-M3 32-bit RISC core for the aforementioned tasks.    
 
\paragraph{Companion Computer (Compute):} The companion computer is a powerful compute unit (compared to the FC) that is responsible for the processing of the high level, computationally intensive tasks such as vision processing. Not all MAVs come equipped with companion computers, rather these are typically an add-on option for more processing. NVIDIA's TX2 is a representative example with significantly more compute capability than an FC.

{\paragraph{Actuators:} Actuators allow agents to react to their surroundings and further modify them. They range anywhere from rather simple rotors to robotic arms capable of grasping and lifting objects. Similar to sensors, their type and number is a function of the workload and processing power on board.

\subsection{Simulation Setup}
\label{sec:setup}

We present how our setup, shown in \Fig{fig:end-to-end}, maps to the various components corresponding to a MAV's operation.

\begin{figure}[t!]
\centering
\includegraphics[trim= 10 10 10 10, clip, width=0.75\columnwidth]{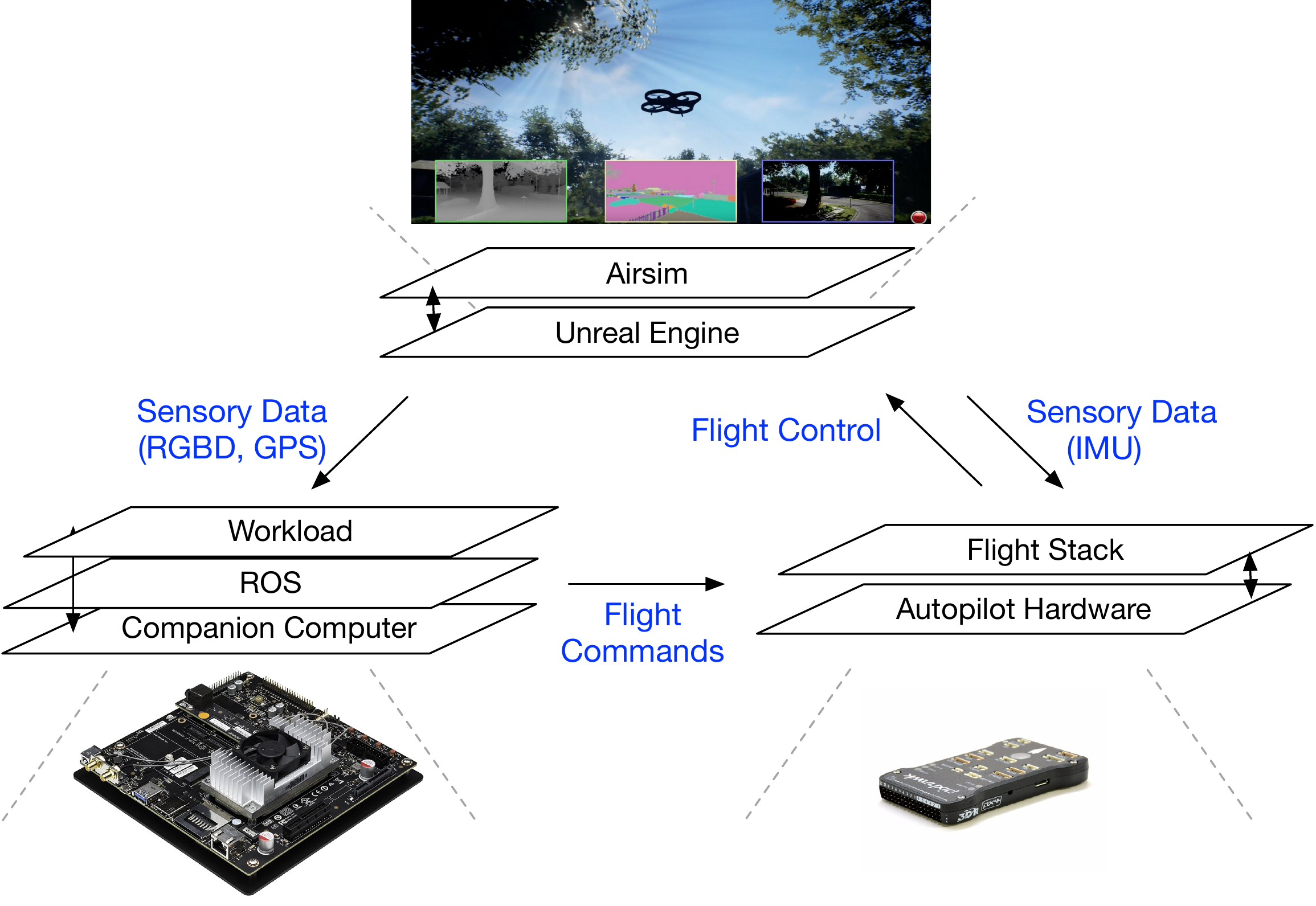}
\caption{Architectural overview of our closed-loop simulation. 
}
\label{fig:end-to-end}
\end{figure}

\paragraph{Environments, Sensors and Actuators:} Environments, sensors and actuators are simulated with the help of a game engine called Unreal~\cite{GameEngi70:online}. With a physics engine at its heart, it ``provides the ability to perform accurate collision detection as well as simulate physical interactions between objects within the world''~\cite{PhysicsS8:online}. Unreal provides a rich set of environments such as mountains, jungles, urban setups, etc. to simulate.

To capture a MAV's dynamics and kinematics through its actuators' behavior and its sensory modules, we used AirSim, an open-source Unreal based plug-in from Microsoft~\cite{Airsim:online}. We limit our sensors and actuators to the ones realistically deployable by MAVs, such as \mbox{RGB-D} cameras and IMUs. Unreal and Airsim run on a powerful computer (host) capable of physical simulation and rendering. Our setup uses an Intel Core i7 CPU and a high-end NVIDIA GTX 1080 Ti GPU.  

\paragraph{Flight Controller:} AirSim supports various flight controllers that can be either hardware-in-the-loop or completely software-simulated. For our experiments, we chose the default software-simulated flight controller provided by AirSim. However, AirSim also supports other FCs, such as the Pixhawk~\cite{Pixhawk:online}, shown in black in \Fig{fig:end-to-end} which runs the PX4~\cite{PX4Archi7:online} software stack. AirSim supports any FC which can communicate using MAVLINK, a widely used micro aerial vehicle message marshaling library~\cite{mavlinkm68:online}. 

\paragraph{Companion Computer:} We used an NVIDIA Jetson TX2~\cite{TX2}, a high-end embedded platform from Nvidia with 256 Pacal CUDA cores GPU and a Quad ARM CPU; however, the flexibility of our setup allows for swapping this embedded board with others such as x86 based Intel Joule~\cite{joule:online}. TX2 communicates with Airsim and also FC via Ethernet.

\paragraph{ROS:} Our setup uses the popular Robot Operating System (ROS) for various purposes such as low-level device control and inter-process communication~\cite{ROSorgPo80:online}.
Robotic applications typically consist of many concurrently-running processes that are known as ``nodes.'' For example one node might be responsible for navigation, another for localizing the agent and a third for object detection. ROS provides peer-to-peer communication between nodes, either through blocking ``service'' calls, or through non-blocking FIFOs (known as the Publisher/Subscriber paradigm). 

\paragraph{Workloads:} Our workloads runs within the ROS runtime on TX2. They are extensively discussed in \Sec{sec:benchmarks}.

\paragraph{Putting It All Together:} To understand the flow of data, we walk the reader through a simple workload where the MAV is tasked to detect an object and fly toward it. The object (e.g. a person) and its environment (e.g. urban) are modeled in the Unreal Engine. As can be seen in \Fig{fig:end-to-end}, the MAV's sensors (e.g. accelerometer and \mbox{RGB-D} Camera), modeled in Airsim, feed their data to the flight controller (e.g. physics data to PX4) and the companion computer (e.g. visual and depth to TX2) using MAVLink protocol. The kernel (e.g. object detection), running within the ROS runtime environment on the companion computer, is continuously invoked until the object is detected. Once so, flight commands (e.g. move forward) are sent back to flight controller, where they get converted to a low level rotor instruction stream flying the MAV closer to the person. 

\subsection{Energy Simulation and Battery Model}
\label{sec:energy}

We extended the AirSim simulation environment with an energy and a battery model. Our energy model is a function of the velocity and acceleration of the MAV~\cite{energyaware}. The higher the velocity or acceleration, the higher the amount of energy consumption. Velocity and acceleration values are sampled continuously, their associated power calculated and integrated for capturing the total energy consumed by the agent.

\newcommand{\norm}[1]{\left\lVert#1\right\rVert}

We used a parametric power estimation model proposed in \cite{3DR-energy-model}. The formula for estimating power $P$ is described below:%
\begin{equation}
\begin{aligned}
P = \begin{bmatrix}
		\beta_{1} \\
		\beta_{2} \\
		\beta_{3}
	\end{bmatrix}^{T}
    \begin{bmatrix}
		\norm{\vec{v}_{xy}} \\
		\norm{\vec{a}_{xy}} \\
		\norm{\vec{v}_{xy}}\norm{\vec{a}_{xy}}
	\end{bmatrix}
    +
    \begin{bmatrix}
		\beta_{4} \\
		\beta_{5} \\
		\beta_{6}
	\end{bmatrix}^{T}
    \begin{bmatrix}
		\norm{\vec{v}_{z}} \\
		\norm{\vec{a}_{z}} \\
		\norm{\vec{v}_{z}}\norm{\vec{a}_{z}}
	\end{bmatrix}
    \\
    +
    \begin{bmatrix}
		\beta_{7} \\
		\beta_{8} \\
		\beta_{9}
	\end{bmatrix}^{T}
    \begin{bmatrix}
		m \\
		\vec{v}_{xy} \cdot \vec{w}_{xy} \\
		1
	\end{bmatrix}
\end{aligned}
\label{eqn:power}
\end{equation}

In the Equation~\ref{eqn:power}, $\beta_{1}$, ..., $\beta_{9}$ are constant coefficients determined based on the simulated drone. $\vec{v}_{xy}$ and $\vec{a}_{xy}$ are the horizontal speed and acceleration vectors whereas  $\vec{v}_{z}$ and $\vec{a}_{z}$ are the corresponding vertical values. $m$ is the mass and $\vec{w}_{xy}$  is the vector of wind movement. 

We have a battery model that implements a coulomb counter approach~\cite{coulomb-counter}. The simulator calculates how many coulombs (product of current and time) have passed  through the drone's battery over every cycle. This is done by calculating the power and the voltage associated with the battery. The real-time voltage is modeled as a function of the percentage of the remaining coulomb in the battery as described in ~\cite{battery-model}.

\subsection{Simulation Knobs and Extensions}
\label{sec:knobs}

With the help of Unreal and AirSim, our setup exposes a wide set of knobs. Such knobs enable the study of agents with different characteristics targeted for a range of workloads and conditions. For different environments, the Unreal market provides a set of maps free or ready for purchase. Furthermore, by using Unreal programming, we introduce new environmental knobs, such as (static) obstacle density, (dynamic) obstacle speed, and so on. In addition, Unreal and AirSim allow for the MAV and its sensors to be customized. For example, the cameras' resolution, their type, number and positions all can be tuned according to the workloads' need.   

Our simulation environment can be extended. For the compute on the edge, the TX2 can be replaced with other embedded systems or even micro-architectural simulators, such as gem5. Sensors and actuators can also be extended, and various noise models can be introduced allowing for reliability studies. 

\subsection{Simulation Fidelity and Limitations}
\label{sec:accuracy}

The fidelity of our end-to-end simulation platform is subject to different sources of error, as it is with any simulation setup. The major obstacle is the \textit{reality gap}---i.e., the difference between the simulated experience and the real world. This has always posed a challenge for robotic systems. The discrepancy results in difficulties where the system developed via simulation does not function identically in the real world. 

To address the reality gap, we iterate upon our simulation components and discuss their fidelity and limitations. Specifically, this involves (1) simulating the environment, (2) modeling the drone's sensors and flight mechanics, and last but not least (3) evaluating the compute subsystem itself.

First, the Unreal engine provides a high fidelity environment. By providing a rich toolset for lighting, shading, and rendering, photo-realistic virtual worlds can be created. In prior work~\cite{unrealcv}, authors examine photorealism by running a Faster-RCNN model trained on PASCAL in an Unreal generated map. The authors show that object detection precision can vary between 1 and 0.1 depending on the elevation and the angle of the camera. Also, since Unreal is open-sourced, we pro grammatically emulate a range of real-world scenarios. For example, we can set the number of static obstacles and vary the speed of the dynamic ones to fit the use case.   

Second, AirSim provides high fidelity models for the MAV, its sensors and actuators. Embedding these models into the environment in a real-time fashion, it deploys a physics engine running with 1000~\si{\hertz}.  As the authors discuss in~\cite{Airsim_paper}, the high precision associated with the sensors, actuators and their MAV model, allows them to simulate a Flamewheel quadrotor frame equipped with a Pixhawk~v2 with little error. 

Flying a square-shaped trajectory with sides of length 5~\si{\meter} and a circle with a radius of 10~\si{\meter},  AirSim achieves  0.65~\si{\meter} and 1.47~\si{\meter} error, respectively. Although they achieve high precision, the sensor models, such as the ``camera lens models,'' ``degradation of GPS signal due to obstacles,'' ``oddities in camera,'' etc. can benefit from further improvements. 

Third, as for the compute subsystem itself, our hardware has high fidelity since we use off-the-shelf embedded platforms for both the companion computer and the flight controller. As for the software, it is crucial to note ROS is widely used and adopted as the \textit{de facto} OS in the robotics research community.

\section{Benchmark Suite} 
\label{sec:mavbench}

\begin{figure*}[t]
\centering
\includegraphics[height=1.65in, keepaspectratio]{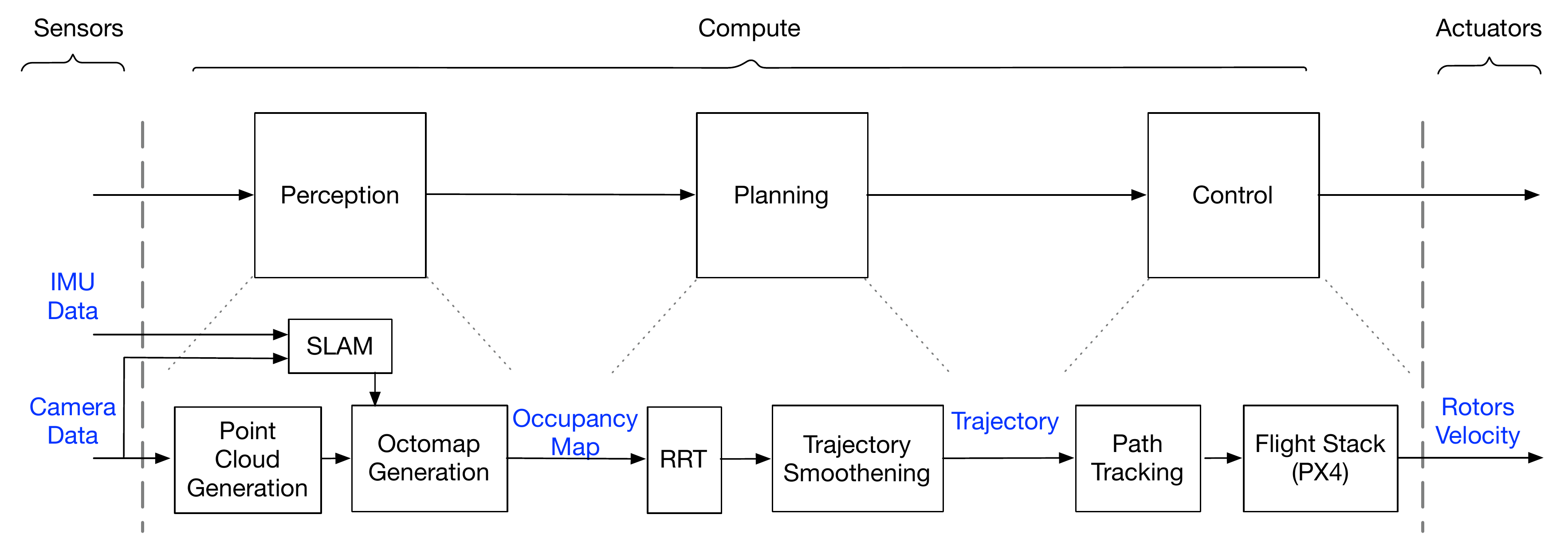}
\caption{High-level application pipeline for a typical MAV application. The upper row presents a universal pipeline that all our MAVBench applications follow, which involves \emph{perception}, \emph{planning} and \emph{control}. The lower row presents how a specific workload in MAVBench (e.g. package delivery) maps to the universal high-level application pipeline.}
\label{fig:software_pipeline}
\end{figure*}

To quantify the power and performance demands of typical MAV applications, we created a set of workloads that we compiled into a benchmark suite. Our benchmarks run on top of our closed-loop simulation environment. The suite aims to cover a wide range of representative applications. Each workload is an \textit{end-to-end} application that allows us to study the kernels' impact on the whole application as well as to investigate the interactions and dependencies between kernel. 

By providing holistic end-to-end applications instead of only focusing on individual kernels, MAVBench allows for the examination of kernels' impacts and their optimization at the application level. This is a lesson learned from Amdahl's law, which recognizes that the true impact of a component's improvement needs to be evaluated globally rather than locally.

In \Sec{sec:sw_pipeline}, we present a high level software pipeline associated (though not exclusive) to our workloads. In \Sec{sec:benchmarks}, we present functional summaries of the workloads in MAVBench, their use cases, and mappings from each workload to the high level software pipeline. In \Sec{sec:kernels}, we describe in details of the prominent computational kernels that are incorporated into our workloads, and finally in \Sec{sec:QoF} we provide a short discussion regarding the Quality-of-Flight (QoF) metrics to evaluate MAV applications.

The MAVBench workloads have different computational kernels, as shown in Table~\ref{kernel_makeup}. MAVBench aims at being comprehensive by (1) selecting applications that target different robotic domains (robotics in hazardous areas, construction, etc.) and (2) choosing kernels (e.g. point cloud, RRT) common across a range of applications, not limited to our benchmark-suite. The computational kernels (OctoMaps, RTT, etc.) that we use in the benchmarks are the building blocks of many robotics applications and hence they are platform agnostic. 



\subsection{Application Dataflow} \label{sec:sw_pipeline}

\begin{figure*}[t]
	\centering
	\begin{subfigure}[t]{1.385in}
		\centering
		\includegraphics[width=\textwidth, height=1in]{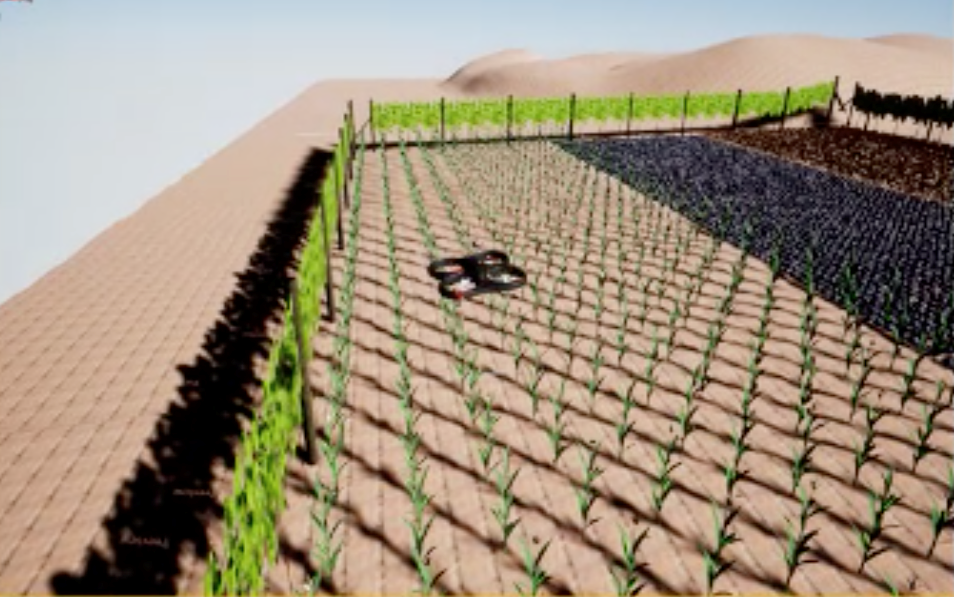}
		\caption{Scanning.}\label{fig:benchmarks:scanning}
	\end{subfigure}
	\hfill
	\begin{subfigure}[t]{1.385in}
		\centering
		\includegraphics[width=\textwidth, height=1in]{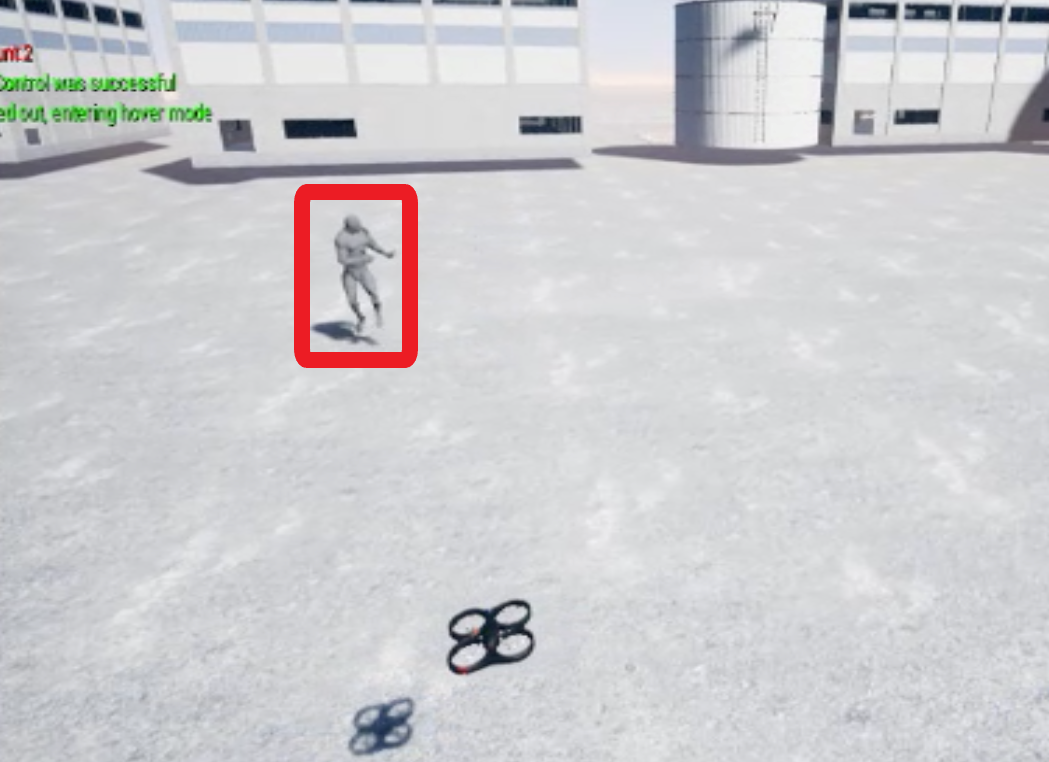}
		\caption{Aerial Photography.}\label{fig:benchmarks:aerial-photo}
	\end{subfigure}
	\hfill
    \begin{subfigure}[t]{1.385in}
		\centering
		\includegraphics[width=\textwidth, height=1in]{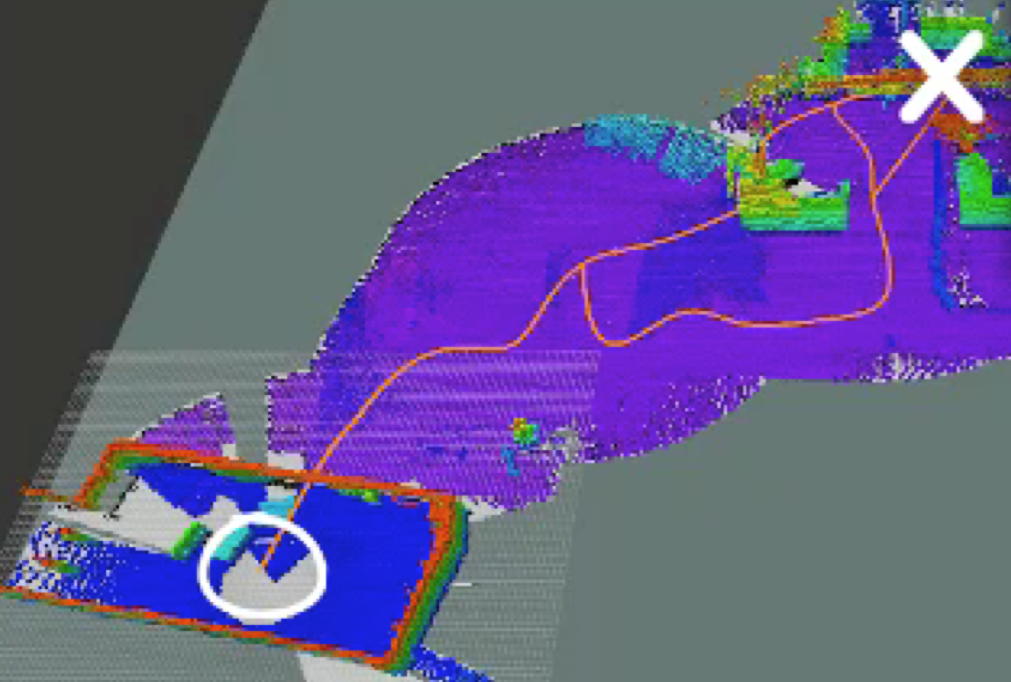}
		\caption{Package Delivery.}\label{fig:benchmarks:package-delivery}
	\end{subfigure}
	\hfill
    \begin{subfigure}[t]{1.385in}
		\centering
		\includegraphics[width=\textwidth, height=1in]{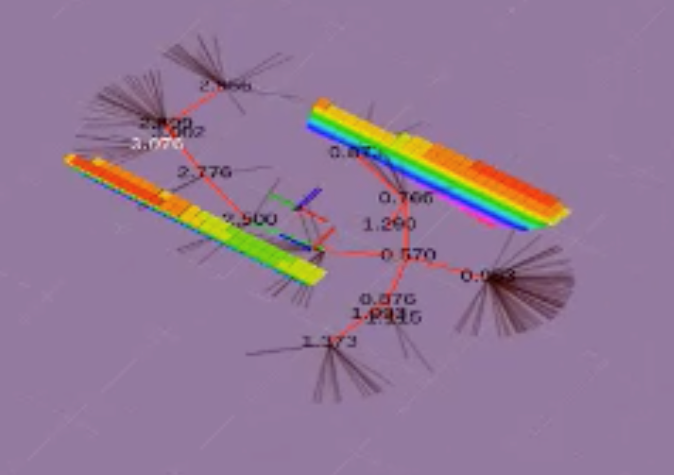}
		\caption{3D Mapping.}\label{fig:benchmarks:3D-mapping}
	\end{subfigure}
	\hfill
    \begin{subfigure}[t]{1.385in}
		\centering
		\includegraphics[width=\textwidth, height=1in]{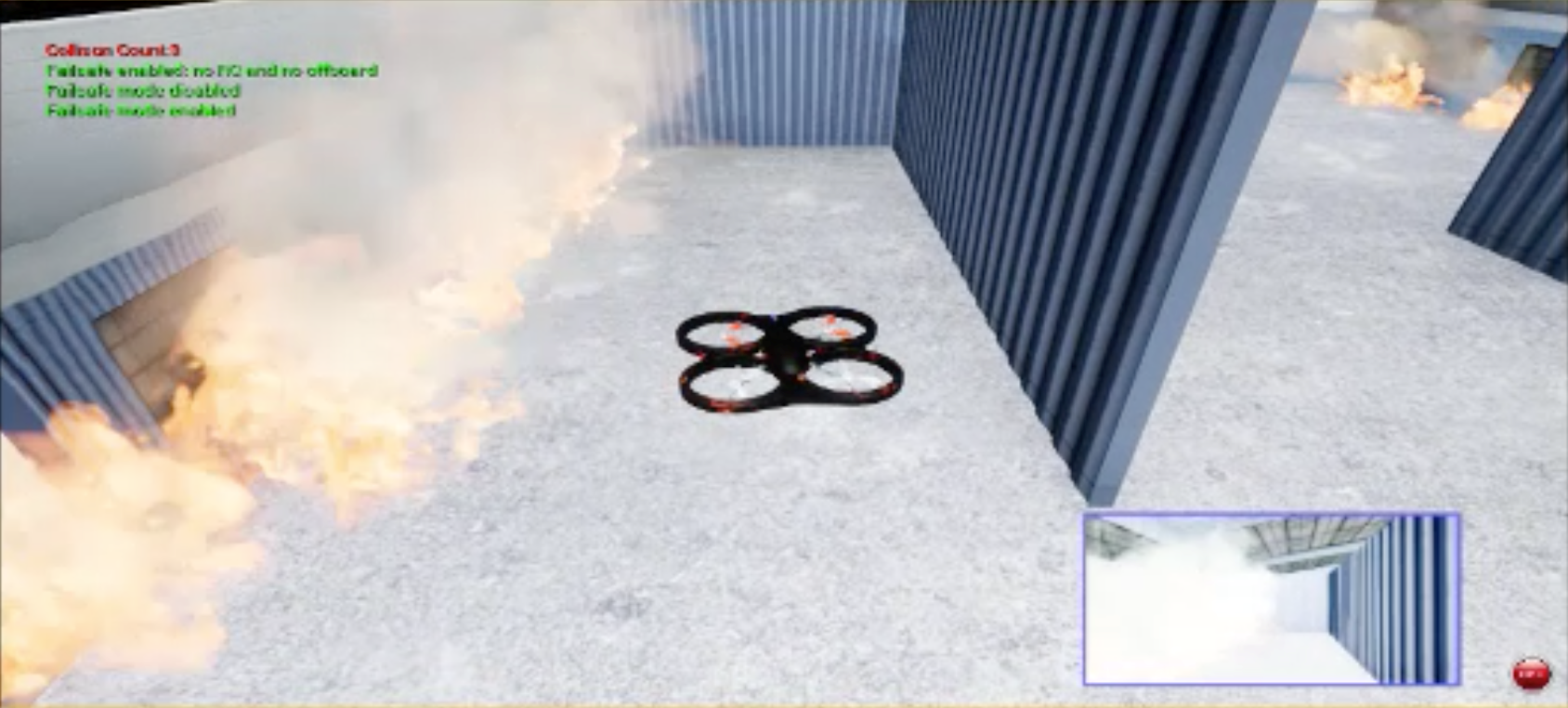}
        \caption{Search and Rescue.}\label{fig:benchmarks:search-and-rescue}
	\end{subfigure}
    \vspace{-5pt}
    \caption{MAVBench workloads. Each workload is an end-to-end application targeting both industry and research use cases. All figures are screenshots of a MAV executing a workload within its simulated environment. Fig.~\ref{fig:benchmarks:package-delivery} shows a MAV planning a trajectory to deliver a package. Fig.~\ref{fig:benchmarks:3D-mapping} shows a MAV sampling its environment in search of unexplored areas to map.}
    \label{fig:bench_screenshot}
\end{figure*}

\begin{figure}[t!]
	\centering
	\begin{subfigure}[t]{\columnwidth}
		\centering
        \vspace{-5pt}
		\includegraphics[width=\columnwidth]{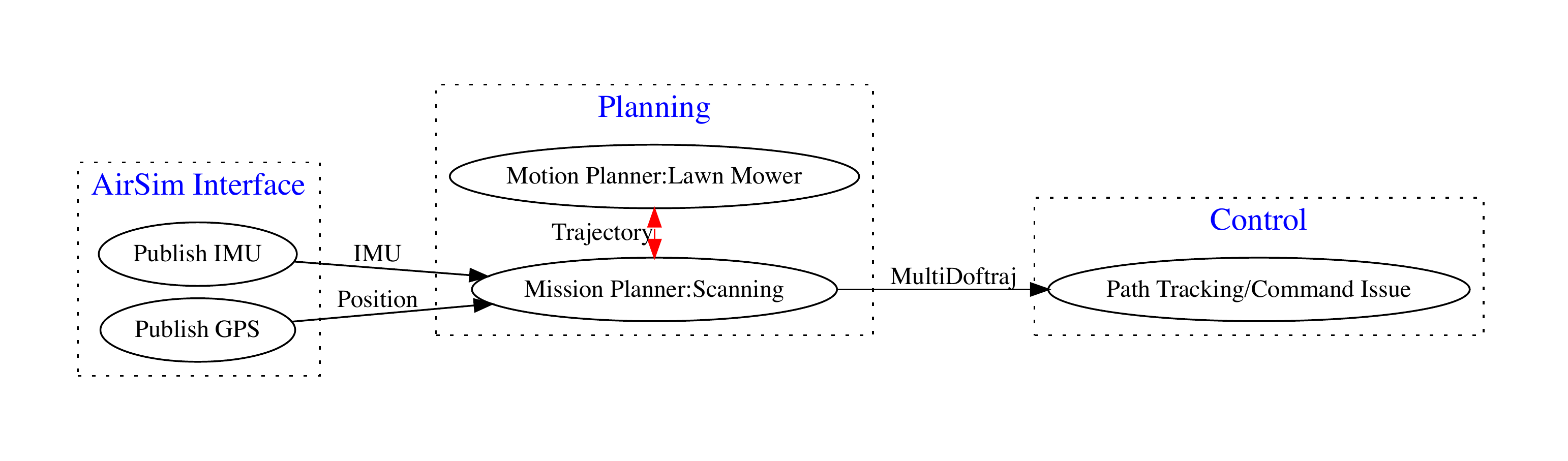}
        \vspace{-20pt}
\caption{Scanning.}\label{fig:benchmarks:data-flow:scanning}
	\end{subfigure}
    \begin{subfigure}[t]{\columnwidth}
		\centering
		\includegraphics[width=\columnwidth]{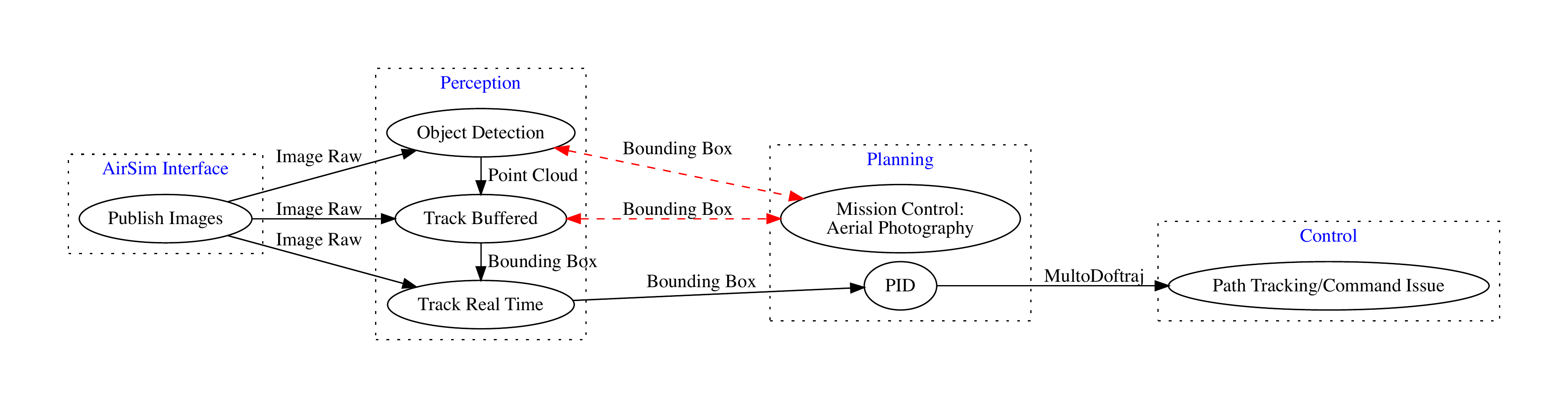}
        \vspace{-20pt}
		\caption{Aerial Photography.}\label{fig:benchmarks:data-flow:aerial_photography}
	\end{subfigure}
	\begin{subfigure}[t]{\columnwidth}
		\centering
		\includegraphics[width=\columnwidth] {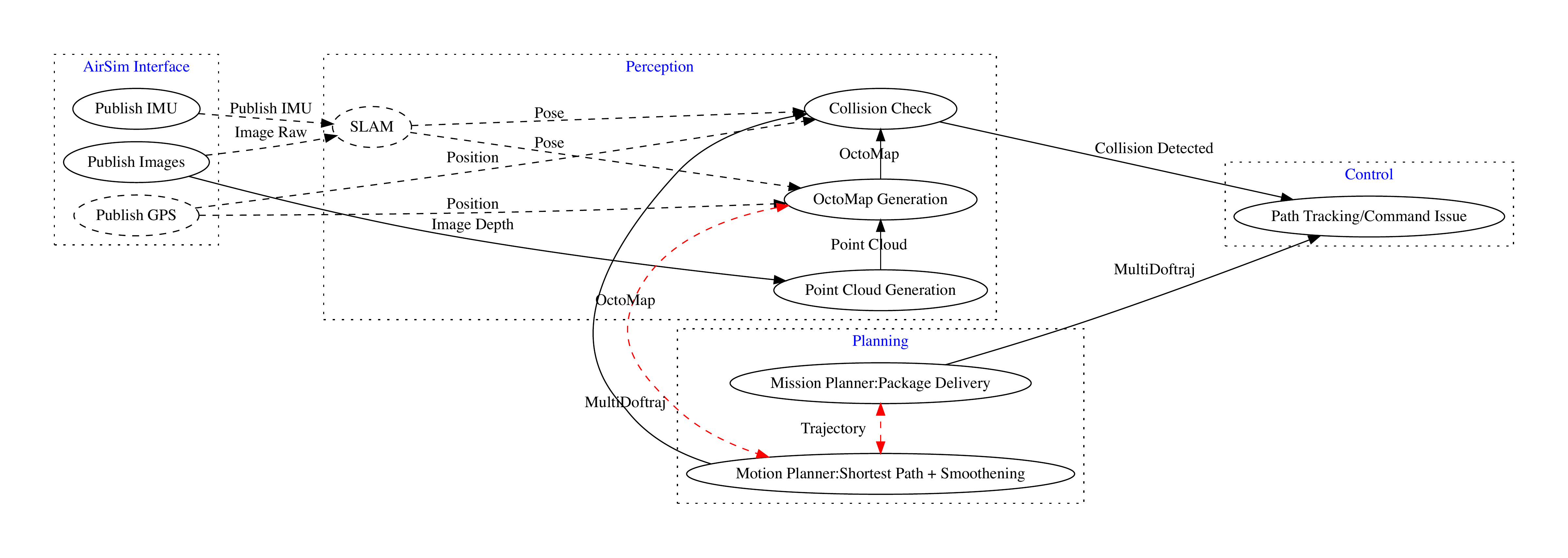}
        \vspace{-20pt}
		\caption{Package Delivery.}\label{fig:benchmarks:data-flow:package_deilvery}
	\end{subfigure}
    \begin{subfigure}[t]{\columnwidth}
		\centering
		\includegraphics[width=\columnwidth]{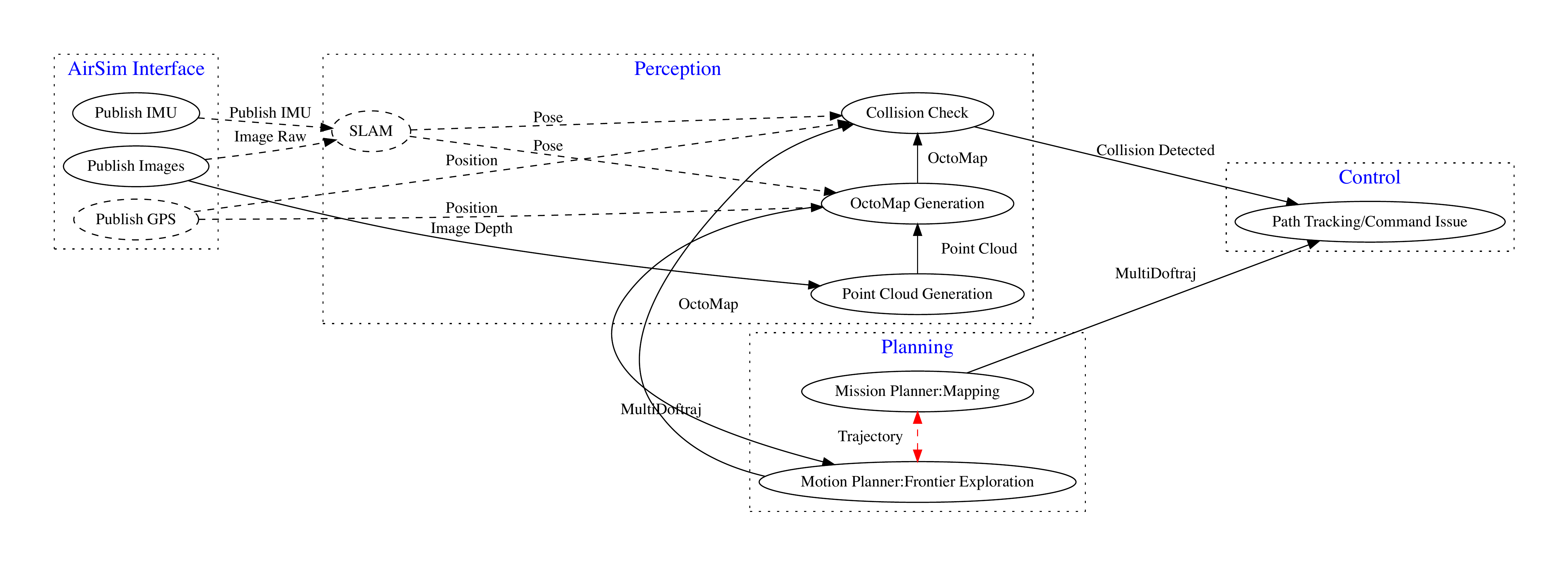}
        \vspace{-20pt}
		\caption{3D Mapping.}\label{fig:benchmarks:data-flow:mapping}
	\end{subfigure}
    \begin{subfigure}[t]{\columnwidth}
		\centering
		\includegraphics[width=\columnwidth]{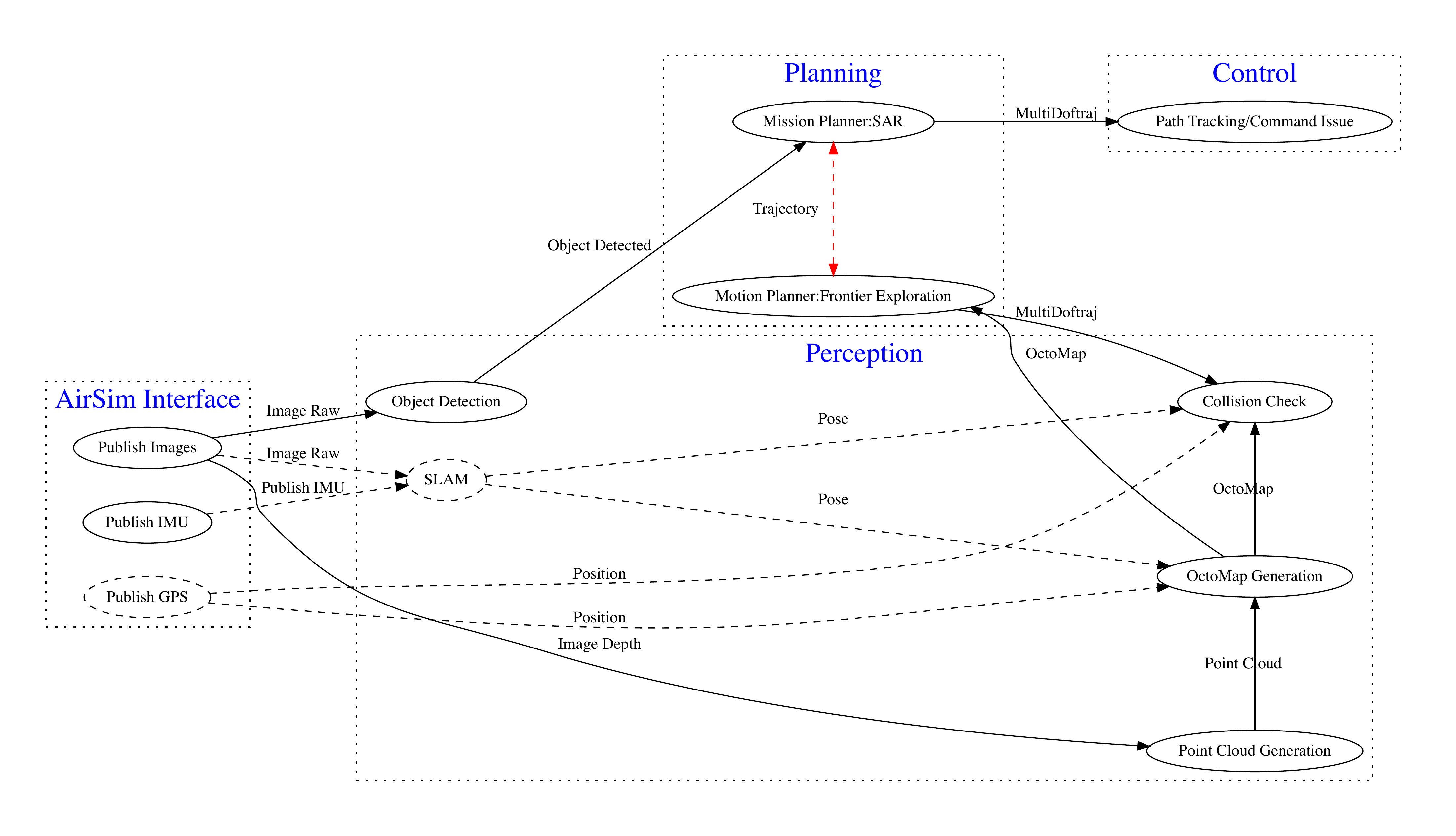}
		\caption{Search and Rescue.}\label{fig:benchmarks:data-flow:sar}
	\end{subfigure}
    \caption{Application dataflows. Nodes are denoted with a circle and communication between the nodes is captured with an arrow. If the communication paradigm between the nodes is of a subscriber/publisher kind, the arrows are filled and black, whereas in the case of the client/server paradigm, they are dotted and red. Nodes with a subscriber/publisher communication paradigm or without any at all run in parallel. Dotted lines denote various localization techniques.}
    \label{fig:benchmarks_data_flow}
\end{figure}

There are three fundamental processing stages in each application: \emph{Perception}, \emph{Planning} and \emph{Control}. In the perception stage, the sensory data is processed to extract relevant states from the environment and the drone. This information is fed into the next two stages (i.e., planning and control). Planning ``plans'' flight motions and forwards them to the actuators in the control subsystem. \Fig{fig:software_pipeline} summarizes this high-level software pipeline, which each of our workloads embody.

\textbf{Perception:}
It is defined as ``the task-oriented interpretation of sensor data''~\cite{Handbook_robotic}. Inputs to this stage, such as sensory data from cameras or depth sensors, are fused to develop an elaborate model in order to extract the MAV's and its environment's relevant states (e.g. the positions of obstacles around the MAV). This stage may include tasks such as Simultaneous Localization and Mapping (SLAM) that enables the MAV to infer its position in the absence of GPS data.

\textbf{Planning:}
 Planning generally involves generating a \textit{collision-free} path to a target using the output of the perception (e.g. a occupancy map of obstacles in the environment). In short, this step involves first generating a set of possible paths to the target, such as by using the probabilistic roadmap (PRM) algorithm, and then choosing an optimal one among them using a path-planning algorithm, such as A*.

\textbf{Control:}
This stage is about following a desired path, which is absorbed from the previous stage, while providing a set of guarantees such as feasibility, stability and robustness~\cite{tech_problem}. In this stage, the MAV's kinematics and  dynamics are considered, such as by smoothening paths to avoid high-acceleration turns, and then, finally, the flight commands are generated (e.g. by flight controllers such as the PX4) while ensuring the aforementioned guarantees are still respected.

\subsection{Benchmarks} \label{sec:benchmarks}
The MAVBench benchmark suite consists of five workloads, each equipped with the flexibility to configure its computational kernel composition (described later in \Sec{sec:kernels}).
The following section sheds light on their functional summary. In addition, the inner workings of these workloads are explained in terms of the three-stage high-level application pipeline. \Fig{fig:bench_screenshot} presents screenshots of these different workloads. The application dataflows are shown in \Fig{fig:benchmarks_data_flow}.

\textbf{Scanning:} In this simple though popular use case, a MAV scans an area specified by its width and length while collecting sensory information about conditions on the ground. It is a common agricultural use case. For example, a MAV may fly above a farm to monitor the health of the crops below. To do so, the MAV first uses GPS sensors to determine its location (Perception). Then, it plans an energy efficient ``lawnmower path'' over the desired coverage area, starting from its initial position (Planning). Finally, it closely follows the planned path (Control). While in-flight, the MAV can collect data on ground conditions using on-board sensors, such as cameras or LIDAR.


\textbf{Aerial Photography:} Drone aerial photography is an increasingly popular use of MAVs for entertainment, as well as businesses. In this workload, we design the MAV to follow a moving target with the help of computer vision algorithms. The MAV uses a combination of object detection and tracking algorithms to identify its relative distance from a target (Perception). Using a PID controller, it then plans motions to keep the target near the center of the MAV's camera frame (Planning), before executing the planned motions (Control).

\textbf{Package Delivery:} In this workload, a MAV navigates through an obstacle-filled environment to reach some arbitrary destination, deliver a package and come back to its origin. Using a variety of sensors such as RGBD cameras or GPS, the MAV creates an occupancy map of its surroundings (Perception). Given this map and its desired destination coordinate, it plans an efficient collision-free path. To accommodate for the feasibility of maneuvering, the path is further smoothened to avoid high-acceleration movements (Planning), before finally being followed by the MAV (Control). While flying, the MAV continuously updates its internal map of its surroundings to check for new obstacles, and re-plans its path if any such obstacles obstruct its planned trajectory.

\textbf{3D Mapping:} With use cases in mining, architecture, and other industries, this workload instructs a MAV to build a 3D map of an unknown polygonal environment specified by its boundaries. To do so, as in package delivery, the MAV builds and continuously updates an internal map of the environment with both ``known'' and ``unknown'' regions (Perception). Then, to maximize the highest area coverage in the shortest time, the map is sampled and a heuristic is used to select an energy efficient (i.e. short) path with a high exploratory promise (i.e. with many unknown areas along the edges) (Planning). Finally, the MAV closely follows this path (Control), until the entire area has been mapped.


\textbf{Search and Rescue:} MAVs are promising vehicles for search-and-rescue scenarios where victims must be found in the aftermath of a natural disaster. For example, in a collapsed building due to an earthquake, they can accelerate the search since they are capable of navigating difficult paths by flying over and around obstacles. In this workload, a MAV is required to explore an unknown area while looking for a target such as a human.  For this workload, the \textit{3D Mapping} application is augmented with an object detection machine-learning-based algorithm in the perception stage to constantly explore and monitor its environment, until a human target is detected.

\subsection{Benchmark Kernels} \label{sec:kernels}

The MAVBench workloads incorporate numerous computational kernels that can be grouped under the three pipeline stages described earlier in Section~\ref{sec:sw_pipeline}. Table~\ref{kernel_makeup} shows the kernel make up of MAVBench's workloads and their corresponding time profile (measured at 2.2 GHz, 4 cores enabled mode of Jetson TX2). MAVBench is equipped with multiple implementations of each computational kernel. For example, MAVBench comes equipped with both YOLO and HOG detectors that can be used interchangeably in workloads with object detection. The user can determine which implementations to use by setting the appropriate parameters. Furthermore, our workloads are designed with a ``plug-and-play'' architecture that maximizes flexibility and modularity, so the computational kernels described below can easily be replaced with newer implementations designed by researchers in the future.

\renewcommand{\arraystretch}{1.15}
\begin{table*}[]
\centering
\caption{MAVBench applications and their kernel make up time profile in $ms$. The application suite, as a whole, exercises a variety of different computational kernels across the perception, planning and control stages, depending on their use case. Furthermore, within each of the kernel computational domain, applications have the flexibility to choose between different kernel implementations.}
\label{kernel_makeup}
\resizebox{\columnwidth}{!}{
\begin{tabular}{c|c|c|c|c|c|c|c|c|c|c|c|c|c|}
\cline{2-14}
\multirow{3}{*}{}                                                                           & \multicolumn{8}{c|}{\textbf{Perception}}                                                                                                                                                                                                                                                                                                                                                                                                                                                                      & \multicolumn{4}{c|}{\textbf{Planning}}                                                                                                                                                                                                                                                                    & \textbf{Control}                                                                                 \\ \cline{2-14} 
                                                                                            & \multirow{2}{*}{\textit{\begin{tabular}[c]{@{}c@{}}Point Cloud\\ Generation\end{tabular}}} & \multirow{2}{*}{\textit{\begin{tabular}[c]{@{}c@{}}Occupancy Map\\ Generation\end{tabular}}} & \multirow{2}{*}{\textit{\begin{tabular}[c]{@{}c@{}}Collision\\ Check\end{tabular}}} & \multirow{2}{*}{\textit{\begin{tabular}[c]{@{}c@{}}Object\\ Detection\end{tabular}}} & \multicolumn{2}{c|}{\textit{\begin{tabular}[c]{@{}c@{}}Object\\ Tracking\end{tabular}}} & \multicolumn{2}{c|}{\textit{Localization}} & \multirow{2}{*}{\textit{PID}} & \multirow{2}{*}{\textit{\begin{tabular}[c]{@{}c@{}}Smoothened\\ Shortest Path\end{tabular}}} & \multirow{2}{*}{\textit{\begin{tabular}[c]{@{}c@{}}Frontier\\ Exploration\end{tabular}}} & \multirow{2}{*}{\textit{\begin{tabular}[c]{@{}c@{}}Smoothened \\Lawn Mowing\end{tabular}}} & \multirow{2}{*}{\textit{\begin{tabular}[c]{@{}c@{}}Path Tracking/\\ Command Issue\end{tabular}}} \\ \cline{6-9}
                                                                                            &                                                                                            &                                                                                              &                                                                                     &                                                                                      & \textit{Buffered}                          & \textit{Real Time}                         & \textit{GPS}        & \textit{SLAM}        &                               &                                                                                              &                                                                                          &                                                                                 &                                                                                                  \\ \hline
\multicolumn{1}{|c|}{\textbf{Scanning}}                                                     &                                                                                            &                                                                                              &                                                                                     &                                                                                      &                                            &                                            &                     &                      &                               &                                                                                              &                                                                                          & 89                                                                              & 1                                                                                                \\ \hline
\multicolumn{1}{|c|}{\textbf{\begin{tabular}[c]{@{}c@{}}Aerial\\ Photography\end{tabular}}} &                                                                                            &                                                                                              &                                                                                     & 307                                                                                  & 80                                         & 18                                         & 0                   &                      & 0                             &                                                                                              &                                                                                          &                                                                                 & 1                                                                                                \\ \hline
\multicolumn{1}{|c|}{\textbf{\begin{tabular}[c]{@{}c@{}}Package\\ Delivery\end{tabular}}}   & 2                                                                                          & 630                                                                                          & 1                                                                                   &                                                                                      &                                            &                                            & 0                   & 55                   &                               & 182                                                                                          &                                                                                          &                                                                                 & 1                                                                                                \\ \hline
\multicolumn{1}{|c|}{\textbf{\begin{tabular}[c]{@{}c@{}}3D\\ Mapping\end{tabular}}}         & 2                                                                                          & 482                                                                                          & 1                                                                                   &                                                                                      &                                            &                                            &              0       & 46                   &                               &                                                                                              & 2647                                                                                     &                                                                                 & 1                                                                                                \\ \hline
\multicolumn{1}{|c|}{\textbf{\begin{tabular}[c]{@{}c@{}}Search and\\ Rescue\end{tabular}}}  & 2                                                                                          & 427                                                                                          & 1                                                                                   & 271                                                                                  &                                            &                                            &               0      & 45                   &                               &                                                                                              & 2693                                                                                     &                                                                                 & 1                                                                                                \\ \hline
\end{tabular}
}
\end{table*}
\renewcommand{\arraystretch}{1}

\paragraph{Perception Kernels:} These are the computational kernels that allow a MAV application to interpret its surroundings.

\textit{Object Detection:} Detecting objects is an important kernel in numerous intelligent robotics applications. So, it is part of two MAVBench workloads: \textit{Aerial Photography} and \textit{Search and Rescue}. MAVBench comes pre-packaged with the YOLO~\cite{yolo16} object detector, and the standard OpenCV implementations of the HOG~\cite{hog} and Haar people detectors.

\textit{Tracking:}
It attempts to follow an instance of an object as it moves across a scene. 
This kernel is used in the \textit{Aerial Photography} workload. MAVBench comes pre-packaged with a C++ implementation~\cite{kcf-c++} of a KCF~\cite{kcf} tracker.

\textit{Localization:} MAVs require a method of determining their position. There are many ways that have been devised to enable localization, using a variety of different sensors, hardware, and algorithmic techniques. MAVBench comes pre-packaged with multiple localization solutions that can be used interchangeably for benchmark applications. Examples include a simulated GPS, visual odometry algorithms such as ORB-SLAM2~\cite{orbslam2}, and VINS-Mono~\cite{vins-mono} and these are accompanied with ground-truth data that can be used when a MAVBench user wants to test an application with perfect localization data.


\textit{Occupancy Map Generation:} Several MAVBench workloads, like many other robotics applications, model their environments using internal 3D occupancy maps that divide a drone's surroundings into occupied and unoccupied space. Noisy sensors are accounted for by assigning probabilistic values to each unit of space.
In MAVBench we use OctoMap~\cite{octomap} as our occupancy map generator since it provides updatable, flexible and compact 3D maps.

\paragraph{Planning Kernels} Our workloads comprise several motion-planning techniques, from simple ``lawnmower" path planning to more sophisticated sampling-based path-planners, such as RRT~\cite{rrt} or PRM~\cite{prm} paired with the A*~\cite{astar} algorithm. We divide MAVBench's path-planning kernels into three categories: \textit{shortest-path planners}, \textit{frontier-exploration planners}, and \textit{lawnmower path planners}. The planned paths are further smoothened using the \textit{path smoothening} kernel. 


\textit{Shortest Path:} Shortest-path planners attempt to find collision-free flight trajectories that minimize the MAV's traveling distance. MAVBench comes pre-packaged with OMPL~\cite{ompl}, the Open Motion Planning Library, consisting of many state-of-the-art sampling-based motion planning algorithms. These algorithms provide collision-free paths from an arbitrary start location to an arbitrary destination. 

\textit{Frontier Exploration:} 
Some applications in MAVBench incorporate collision-free motion-planners that aim to efficiently ``explore'' all accessible regions in an environment, rather than simply moving from a single start location to a single destination as quickly as possible.
For these applications, MAVBench comes equipped with the official implementation of the exploration-based ``next best view planner''~\cite{nbvplanner}.

\textit{Lawnmower:} Some applications do not require complex, collision-checking path planners. For example, agricultural MAVs are frequently tasked with flying over farms in a simple, lawnmower pattern, where the high-altitude of the MAV means that obstacles can be assumed to be nonexistent. For such applications, MAVBench comes with a simple path-planner that computes a regular pattern for covering rectangular areas.

\textit{Path Smoothening:} The motion planners discussed earlier return piecewise trajectories that are composed of straight lines with sharp turns. However, sharp turns require high accelerations from a MAV, consuming high amounts of energy (i.e., battery capacity). Thus, we use this kernel to convert these piecewise paths to smooth, polynomial trajectories that are more efficient for a MAV to follow.

\paragraph{Control Kernels} The control stage of the pipeline enables the MAV to closely follow its planned motion trajectories in an energy-efficient, stable manner. 

\textit{Path Tracking:} MAVBench applications produce trajectories that have specific positions, velocities, and accelerations for the MAV to occupy at any particular point in time. However, due to mechanical constraints, the MAV may drift from its location as it follows a trajectory, due to small but accumulated errors. So, MAVBench includes a computational kernel that guides MAVs to follow trajectories while repeatedly checking and correcting the error in the MAV's position.

\subsection{Quality-of-Flight (QoF) Metrics}
\label{sec:QoF}

Various figures of merits can be used to measure a drone's mission quality. While some of these metrics are universally applicable across applications, others are specific to the application under inquiry. On the one hand, for example, a mission's overall time and energy consumption are almost universally of concern. On the other hand, the discrepancy between a collected and ground truth map or the distance between the target's image and the frame center are specialized metrics for 3D mapping and aerial photography respectively. MAVBench platform collects statistics of both sorts; however, this paper mainly focuses on time and energy due to their universality. 
\section{The Role of Compute in MAVs}
\label{sec:char}

In this section, we discuss how compute affects MAV systems. At the high level, compute plays a crucial role both in the overall mission time and total energy consumption of such systems. First, we discuss each effect by providing relevant  theoretical background and supplement the discussion with a microbenchmark. Then, we analyze MAVBench as a set of representative applications in which such effects can manifest.   



\subsection{Compute and Flight Time Relationship}

Compute can play an important role in reducing the drone's mission time by increasing the mission's average velocity. Concretely, we identify that the reduction in hover time and the increase in maximum allowed velocity are the two major ways with which more compute can contribute to a higher average velocity. Here, we shed light on these different ways.

\paragraph{Hover Time Reduction:} Hover time and the average velocity have an inverse relationship, namely, the more drone spends time on hovering, the lower its average velocity. Similar to an idling CPU, a hovering drone is unfavorable since it is not working toward its mission, but yet wasting its limited energy. A hovering drone typically is waiting for its mission planning stage to make a decision (e.g. deciding on a path to follow). \emph{More compute power can help reduce hovering (e.g by reducing the decision making time),} hence decreasing the negative effect hovering has on the average velocity. 
\begin{figure}[t!]
\centering
    \begin{subfigure}{.49\columnwidth}
    \centering
    \includegraphics[trim=0 0 0 0, clip, width=1.0\columnwidth]{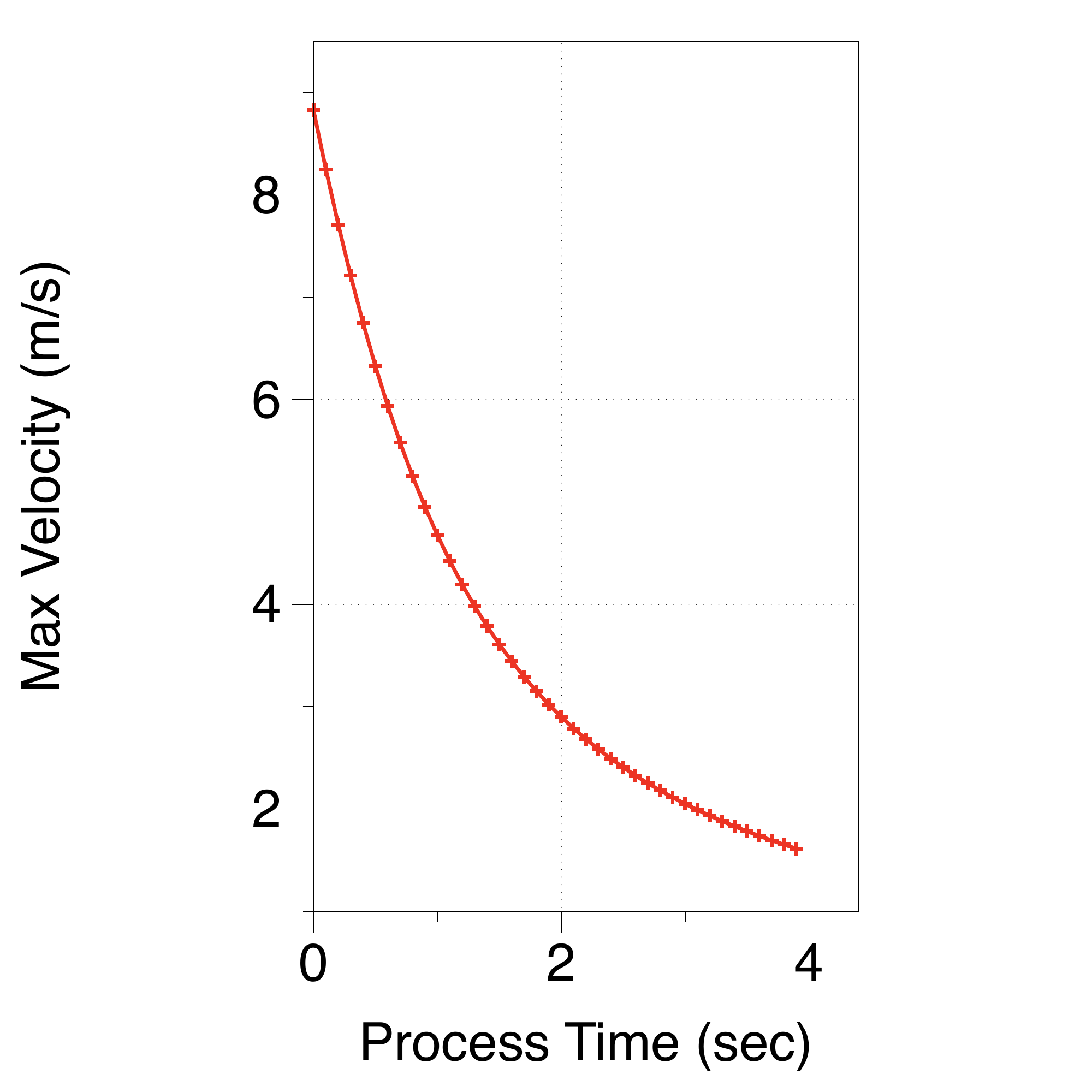}
    \caption{Theoretical max velocity.}
    \label{fig:process-time-velocity}
    \end{subfigure}
    \begin{subfigure}{.49\columnwidth}
    \centering
   \includegraphics[trim=0 0 0 0, clip, width=1.0\columnwidth]{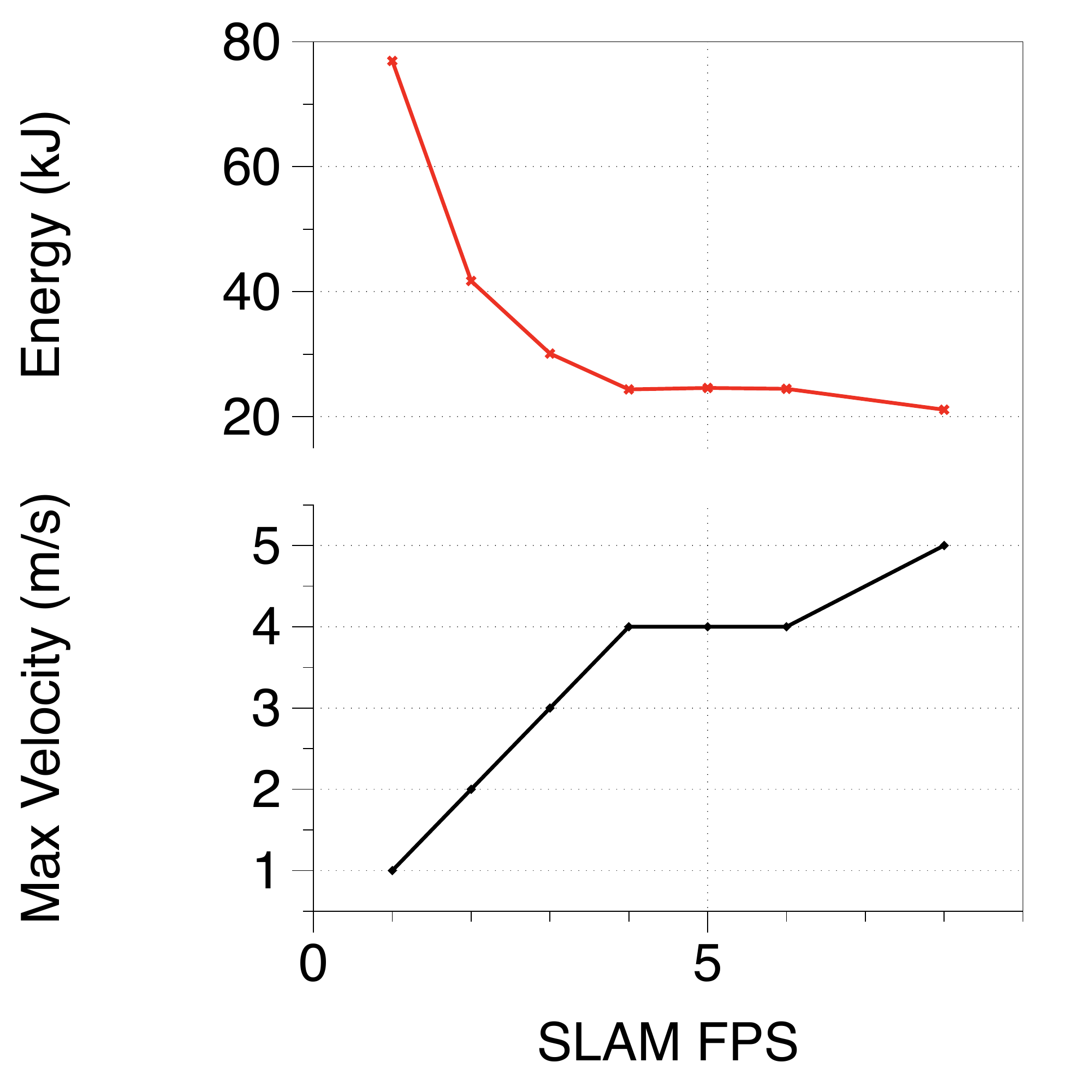}
    \caption{Measured max velocity.}
    \label{fig:slam-velocity-energy}
    \end{subfigure}
\vspace{-5pt}
\caption{\emph{(a)} Theoretical relationship between processing time and maximum velocity. \emph{(b}) Relationship between SLAM throughput (FPS) and maximum velocity and energy of UAVs.}
\end{figure}

\paragraph{Max Velocity Increase, Collision Avoidance Effect:} The maximum velocity of the drone is not only mechanically bounded, but also compute bounded. For a given flight velocity, a collision-free flight is only possible if the drone can process its surrounding fast enough to react to it. Therefore, \emph{a higher velocity requires a faster processing capability}. The collision avoidance task can be rather compute intensive exercising various stages of the pipeline starting from the pixel processing in the perception stage and ending with the command issue in control. In order to guarantee collision avoidance, a  drone's maximum velocity is determined based on the aforementioned pixel to response time. Equation~\ref{eq:runtime-compute-bound} specifies the components involved in setting this velocity where $\delta_{t}$, d, $a_{max}$ and v denote process time, required stopping distance, maximum acceleration limit of the drone and maximum velocity~\cite{high-speed-nav}. As \Fig{fig:process-time-velocity} shows, our simulated drone, in theory, is bounded by the max velocity anywhere between 8.83 to 1.57 m/s given a pixel to response time of the range 0 to 4~seconds. 

\begin{equation}
\label{eq:runtime-compute-bound}
v_{max} = a_{max}(~\sqrt[]{\delta{t}^2 + 2\frac{d}{a_{max}}} - \delta{t})
\end{equation}
\setlength{\belowcaptionskip}{-1ex}

\paragraph{Max Velocity Increase, Localization Failure Effect:} The faster the speed of the drone, the higher the likelihood of its localization failure because the environment changes rapidly around a fast drone. Kernels such as SLAM which help localize the drone by tracking a set of points/features through successive camera frames struggle to keep up with the rate of these changes. It is important to note that localization failures can have catastrophic effects such as permanent loss or spending of extra time (for example by backtracking) for re-localization. 

Minimizing or avoiding localization related failure scenarios is highly favorable, if not necessary. To examine the relationship between the compute, maximum velocity and localization failure, we devised a micro-benchmark in which the drone was tasked to follow a predetermined circular path of the radius 25~meters. For the localization kernel, we used ORB-SLAM2 and to emulate different compute powers, we inserted a sleep in the kernel. We swept different velocities and sleep times and bounded the failure rate to 20\%. As \Fig{fig:slam-velocity-energy} shows, higher FPS values, i.e. more compute, allows for a higher maximum velocity for a bounded failure rate. 

\begin{figure}[!t]
\centering
    \begin{subfigure}{.49\columnwidth}
    \centering
    \includegraphics[trim=0 0 0 0, clip, width=1.0\columnwidth]{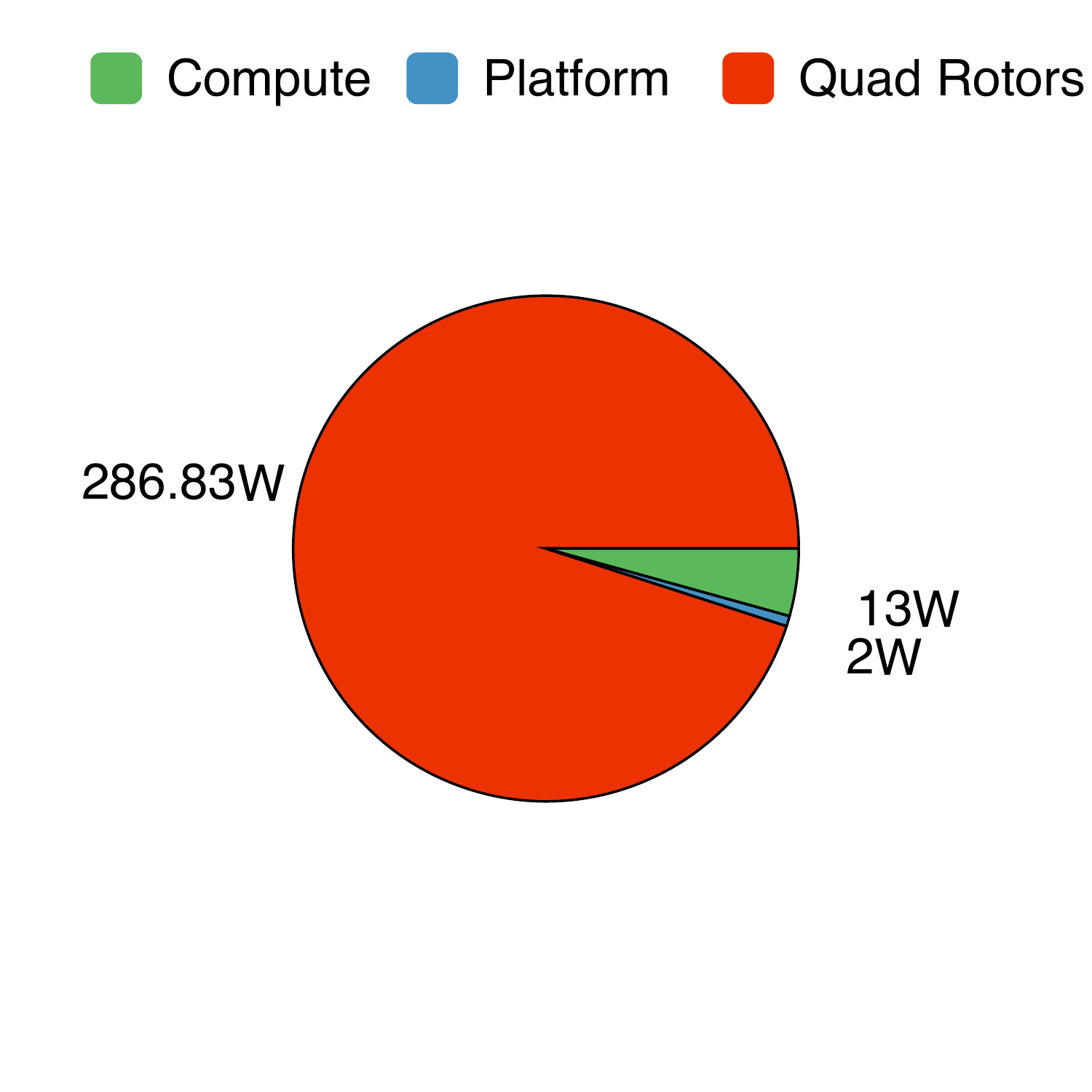}
    \caption{Measured power breakdown.}
    \label{fig:SOLO-power-breakdown}
    \end{subfigure}
    \begin{subfigure}{.49\columnwidth}
    \centering
    \includegraphics[trim=0 0 0 0, clip, width=1.0\columnwidth]{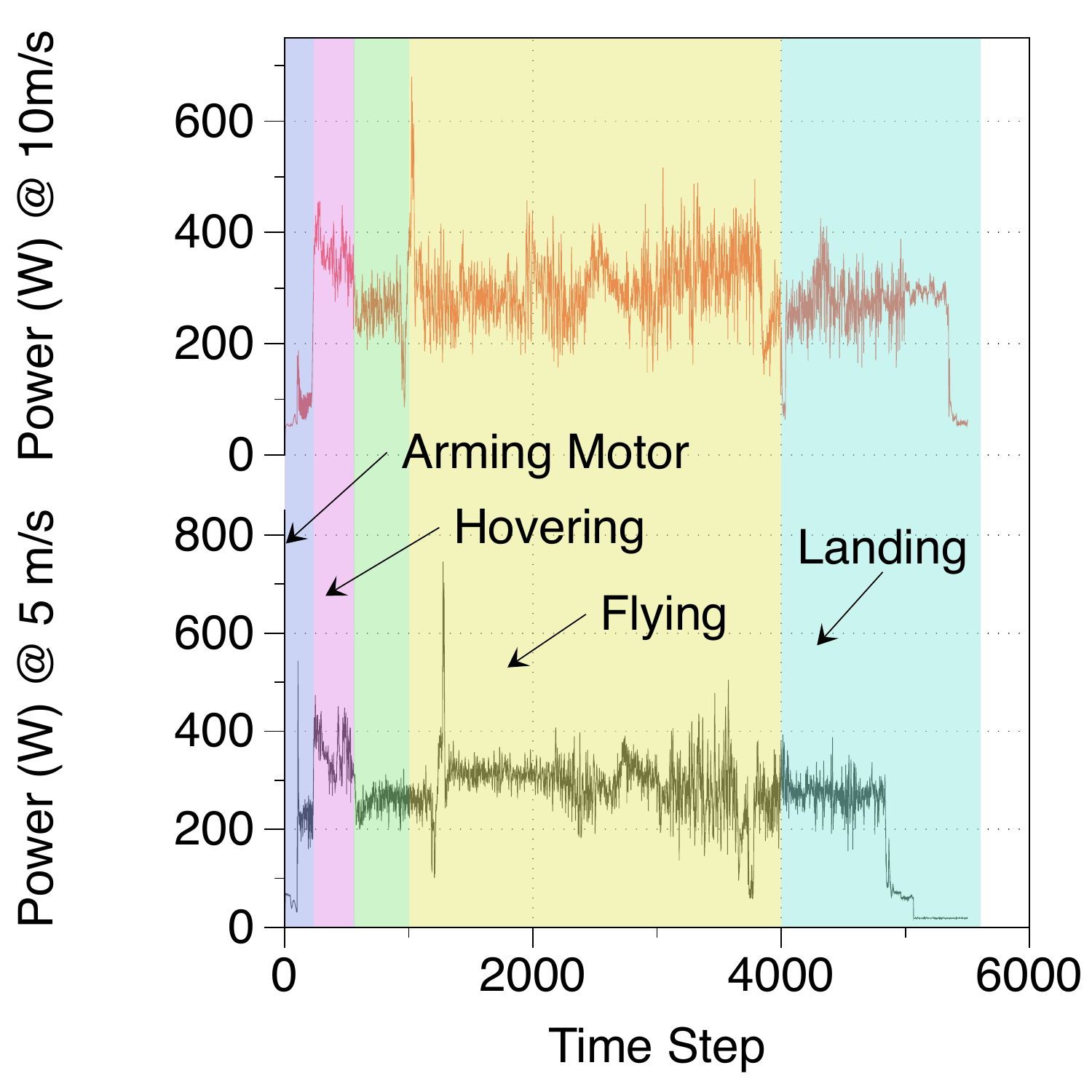}
    \caption{Measured mission power.}
    \label{fig:drone-power-time-series}
    \end{subfigure}
\caption{\emph{(a)} Measured power consumption of a flying 3DR Solo. \emph{(b)} Total measured power consumption while the drone is "Flying" at two different steady-state velocities. The power consumption is severely dominated by the quad rotors by 20X.}
\label{SOLO_power_breakdown}
\end{figure}
\subsection{Compute and Energy Relationship}
The compute subsystem can also have a significant role in reducing total MAV energy consumption. To understand this, first we present the power distribution associated with \solo~\cite{solo3DR}, a popular off-the-shelf MAV. To measure power, we attach a wattmeter known as Eagle~Tree Systems eLogger~V4~\cite{eLoggerV4} to the \solo's battery during flight. The wattmeter allows us to collect data over time at 50~Hz while the drone flies. We command the drone to fly for fifty seconds and pull the data off of the wattmeter after the drone lands.


As \Fig{SOLO_power_breakdown} shows, the majority of the power consumption is dedicated to rotors (locomotion) and the compute only occupies a small portion of the entire pie and its role seemingly trivial. Although small in quantity, compute can have a grand effect on the system's power. This is because by reducing the mission time (as explained in the previous section), more compute power can, in fact, reduce the bigger portion of the pie, namely rotors energy consumption (due to a shorter flight). Note that although more compute can lead to more energy consumption of the compute subsystem, the reduction in rotor's energy can easily outweigh such an increase.  

We profiled the mission time and the energy associated with the aforementioned microbenchmark. As the bottom plot in \Fig{fig:slam-velocity-energy} shows, higher compute capability results in increased SLAM FPS and hence a reduction in mission time by allowing for faster velocity. The reduced mission results in reduced total system energy, as the top plot in \Fig{fig:slam-velocity-energy} shows. By increasing processing speed by 5X, we were able to reduce the drone's energy consumption by close to 4X.

\subsection{MAVBench Workloads Compute vs. Flight Time vs. Energy}
\label{sec:operating-spoints}
We use our benchmark suite as a representative set of applications to examine the effect of compute on MAV systems. To analyze this effect, we conducted sensitivity analysis to core and frequency scaling of the TX2 board. TX2 has two sets of cores, namely a \textit{HMP Dual Denver cores} and a \textit{Quad ARM A57}. We turned off the Denver cores so that the indeterminism caused by process to core mapping variations across runs would not affect our results.
Average velocity, mission, and energy values of various operating points are profiled and presented as heat maps (\Fig{fig:benchmarks:OPA:scanning}---\Fig{fig:benchmarks:OPA:ap}) for a DJI Matrice 100 drone. \emph{In general, compute can improve mission time and lower energy consumption by as much as 5X.}

\paragraph{Scanning:}
We observe trivial differences for velocity, endurance and energy across all three operating points (\Fig{fig:benchmarks:OPA:scanning:velocity}, \Fig{fig:benchmarks:OPA:scanning:time}, and \Fig{fig:benchmarks:OPA:scanning:energy}). This is despite seeing a 3X boost in the motion planning kernel, i.e. lawn mower planning, which is its bottleneck (\Fig{fig:kernel-breakdown}). The trivial effect of compute on this application is because planning is done once at the beginning of the mission and its overhead is amortized over the rest of the mission time. For example the overhead of planning for a 5 minute flight is less than .001\%.  






\paragraph{Package Delivery}:  As compute scales with the number of cores and/or frequency values, we observe a reduction of up to 84\% and 82\% for the mission time and energy consumption, respectively (\Fig{fig:benchmarks:OPA:pd:time}, and \Fig{fig:benchmarks:OPA:pd:energy}). The sequential bottlenecks i.e. motion planning and OctoMap generation kernel are sped up by frequency scaling to enable the observed improvements. There does not seem to be a clear trend with core scaling, concretely between 3 and 4 cores. We conducted investigation and determine that such anomalies are caused by the non-real-time aspects of ROS, AirSim and the TCP/IP protocol used for the communication between the companion computer and the host. We achieve up to 2.9X improvement in OctoMap generation and that leads to maximum velocity improvement. It is important to note that although we also gain up to 9.2X improvements for the motion planning kernel, the low number of re-plannings and its short computation time relative to the entire mission time render its impact trivial. Overall the aforementioned improvements translate to up to 4.8X improvement in the average velocity. Therefore, mission time and the MAV's total energy consumption are reduced.

{
\begin{figure}[t!]
    	\centering
    	\begin{subfigure}[t!]{.3\columnwidth}
    	\centering
    		\includegraphics[width=\columnwidth]{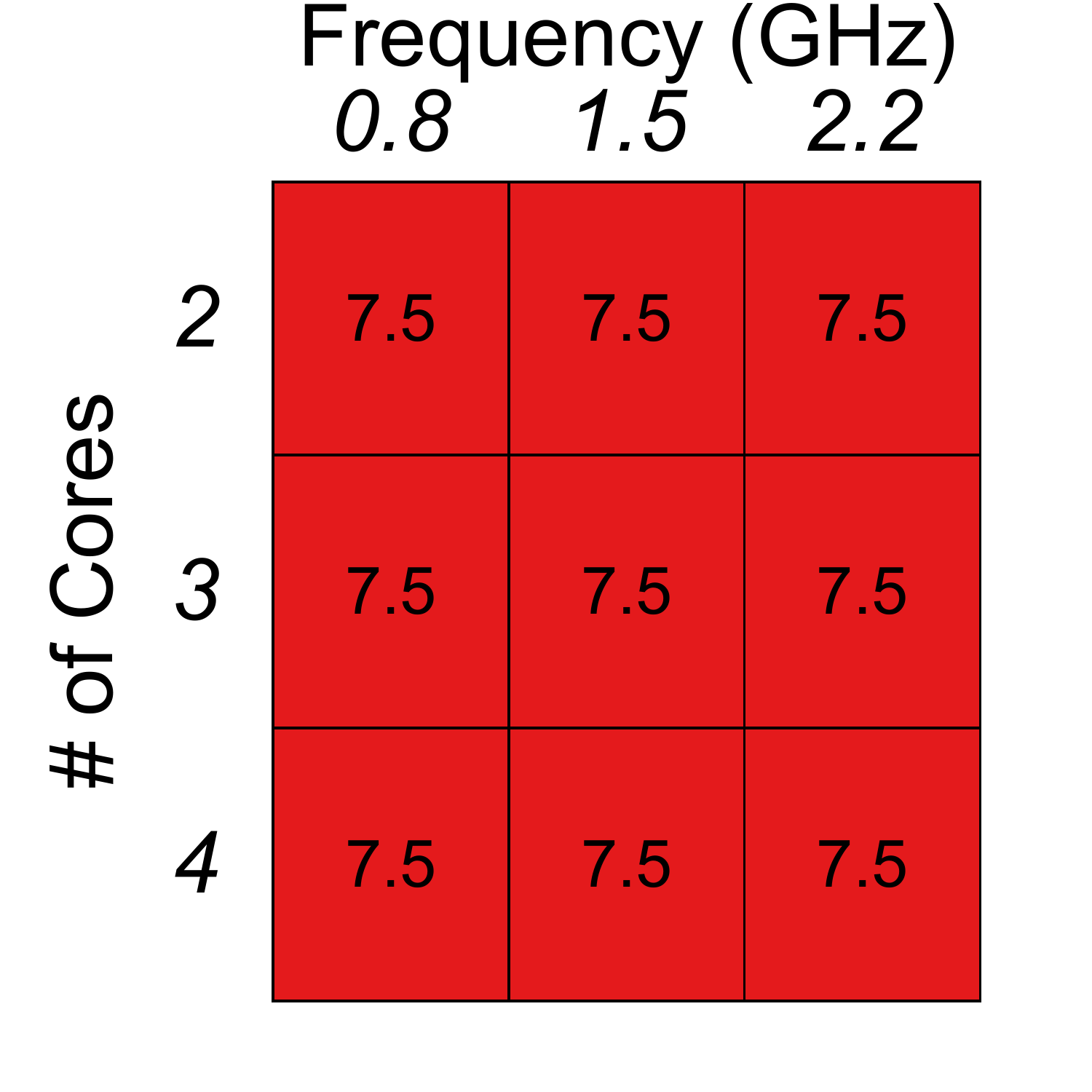}
    		\caption{Velocity (m/s)}
            \label{fig:benchmarks:OPA:scanning:velocity}
    	\end{subfigure}
        \begin{subfigure}[t!]{.3\columnwidth}
    	\centering 
    \includegraphics[width=\columnwidth]{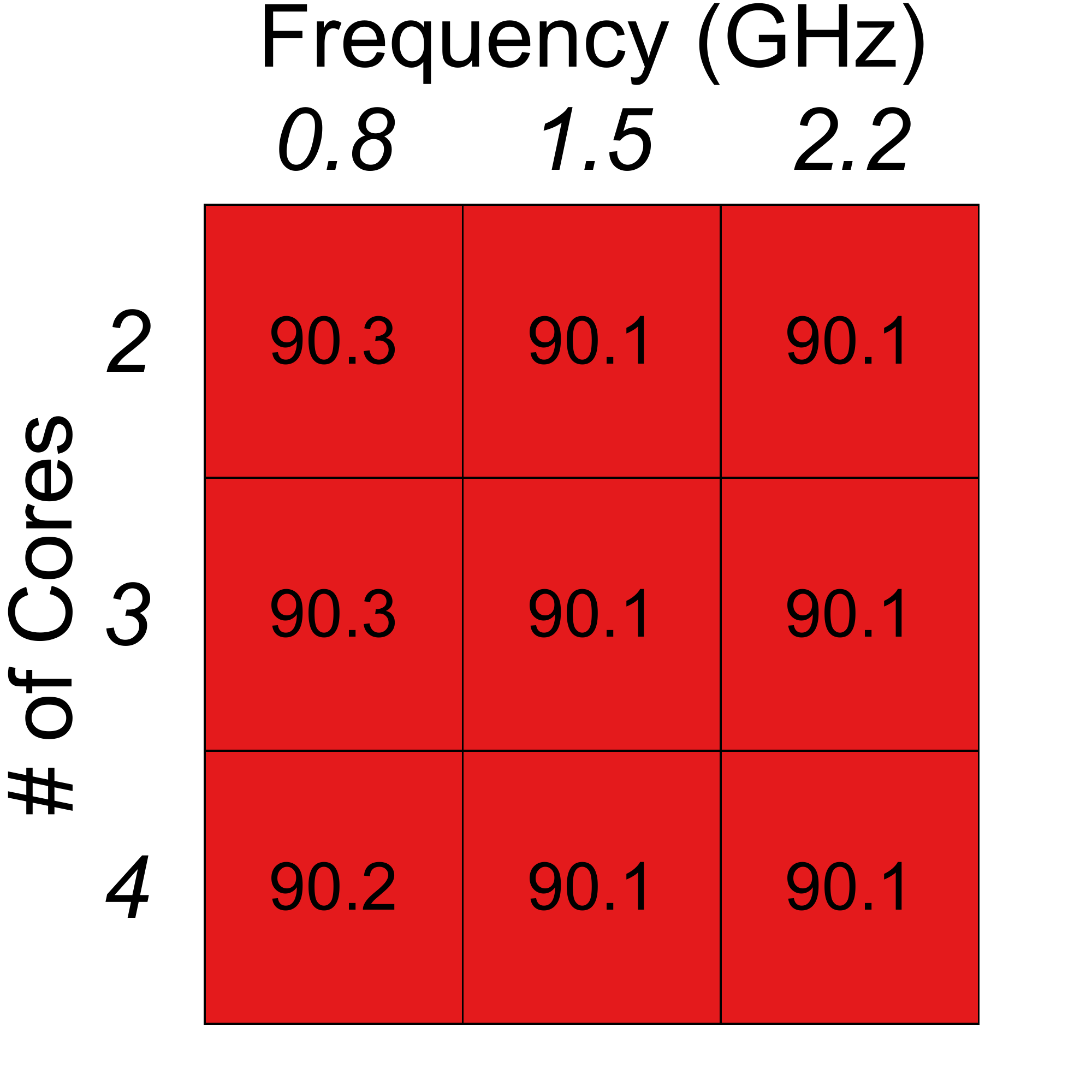}
    \caption{Mission Time (s)}
    \label{fig:benchmarks:OPA:scanning:time}
    \end{subfigure}
    \begin{subfigure}[t!]{.3\columnwidth}
    \centering
    \includegraphics[width=\columnwidth] {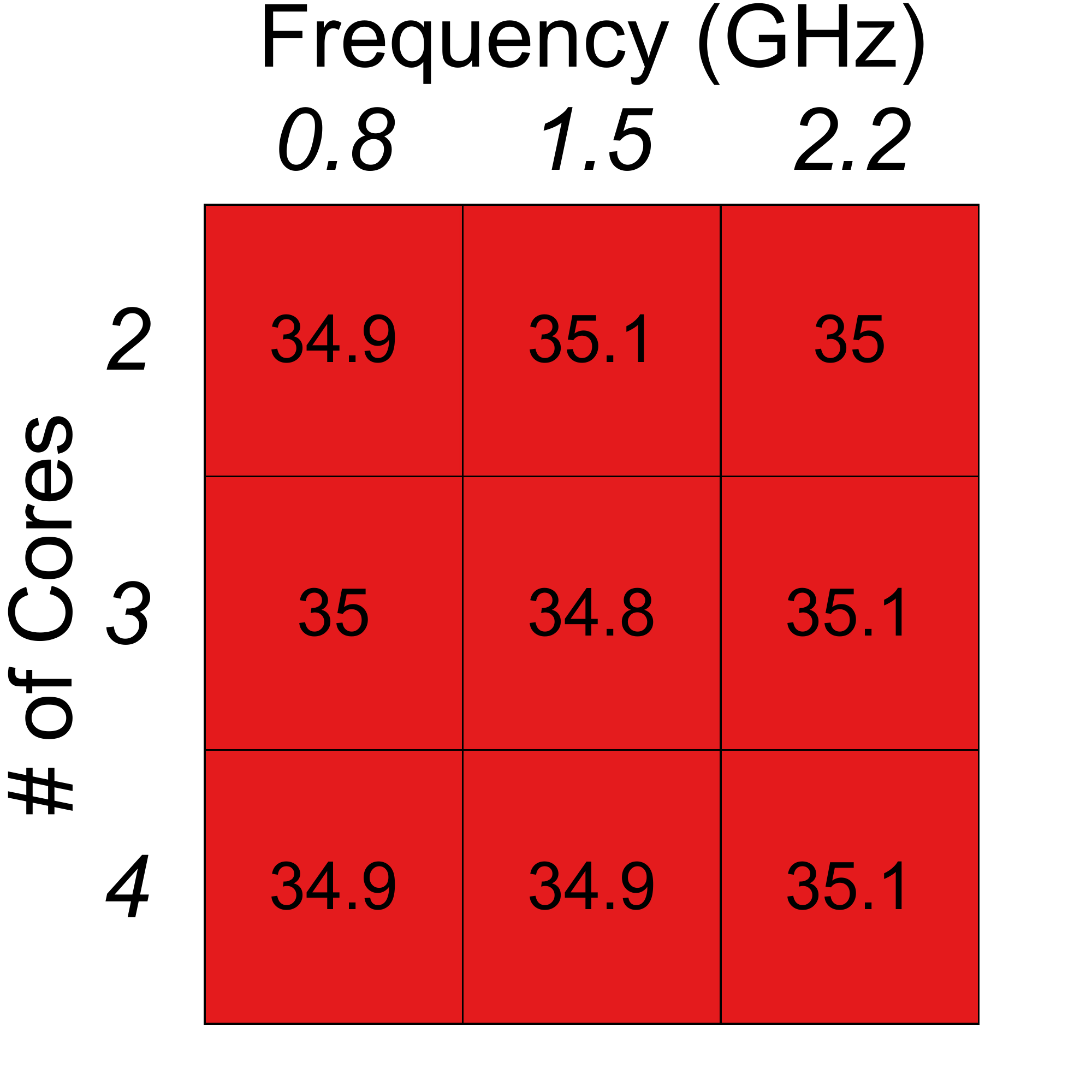}
    \caption{Energy (kJ)}
    \label{fig:benchmarks:OPA:scanning:energy}
    \end{subfigure}
    \caption{Scanning.}
    \label{fig:benchmarks:OPA:scanning}
    \end{figure}%
   \begin{figure}[t!]
    \centering
    \begin{subfigure}[t!]{.3\columnwidth}
    \centering
    \includegraphics[width=\columnwidth]{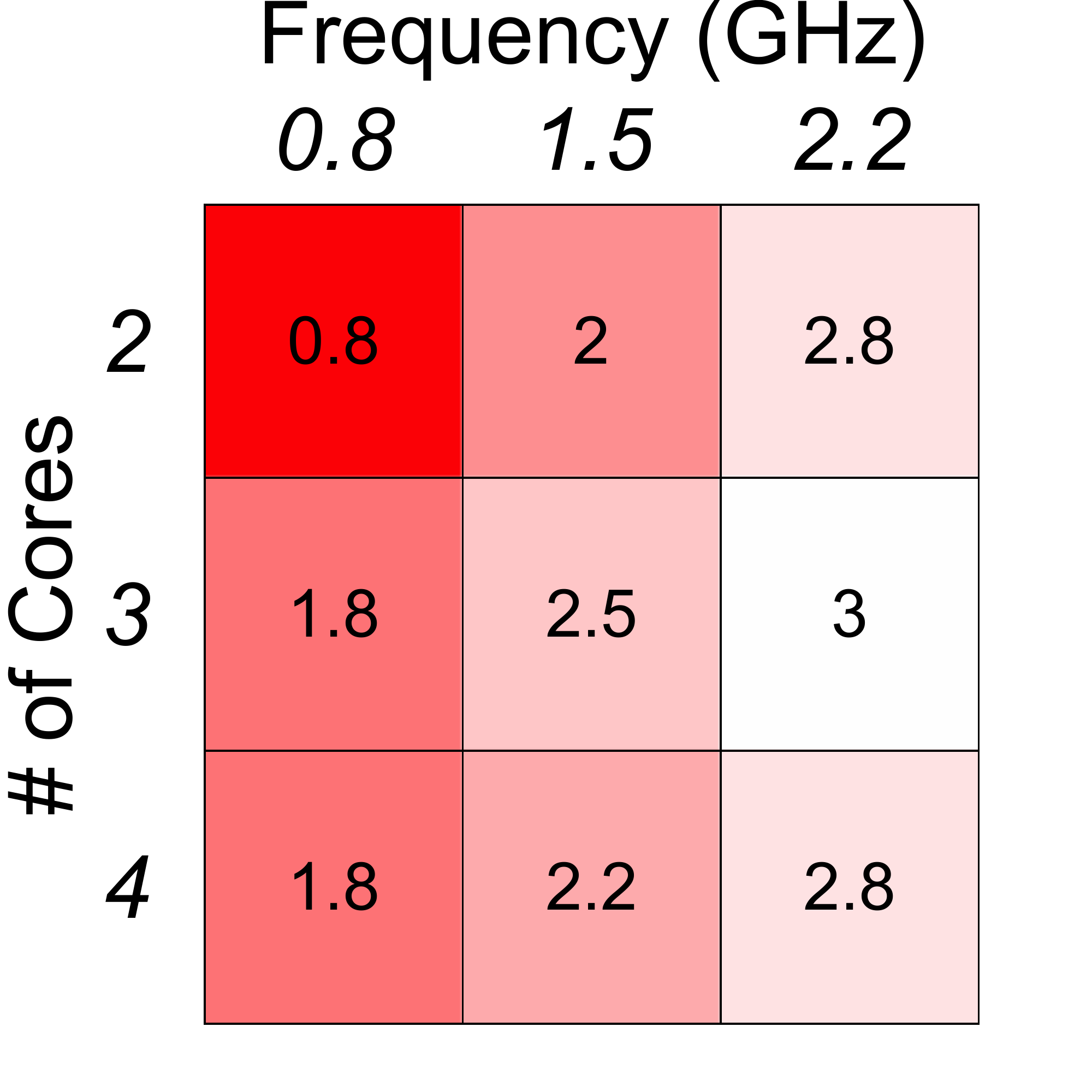}
    \caption{Velocity (m/s)}
     \label{fig:benchmarks:OPA:pd:velocity}
    \end{subfigure}
    \begin{subfigure}[t!]{.3\columnwidth}
    \centering
    \includegraphics[width=\columnwidth]{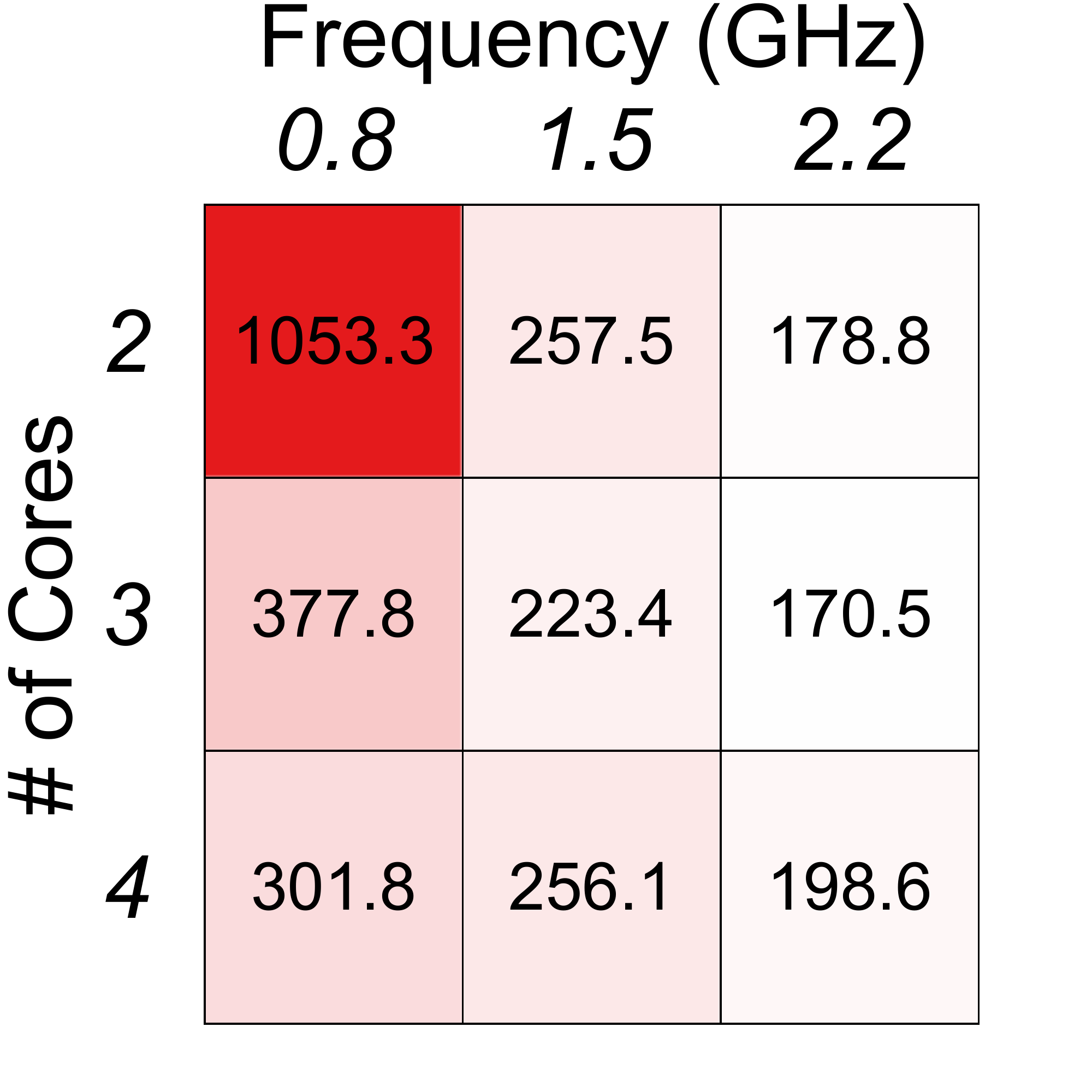}
    \caption{Mission Time (s)}
    \label{fig:benchmarks:OPA:pd:time}
    \end{subfigure}
    \begin{subfigure}[t!]{.3\columnwidth}
    \centering
    \includegraphics[width=\columnwidth] {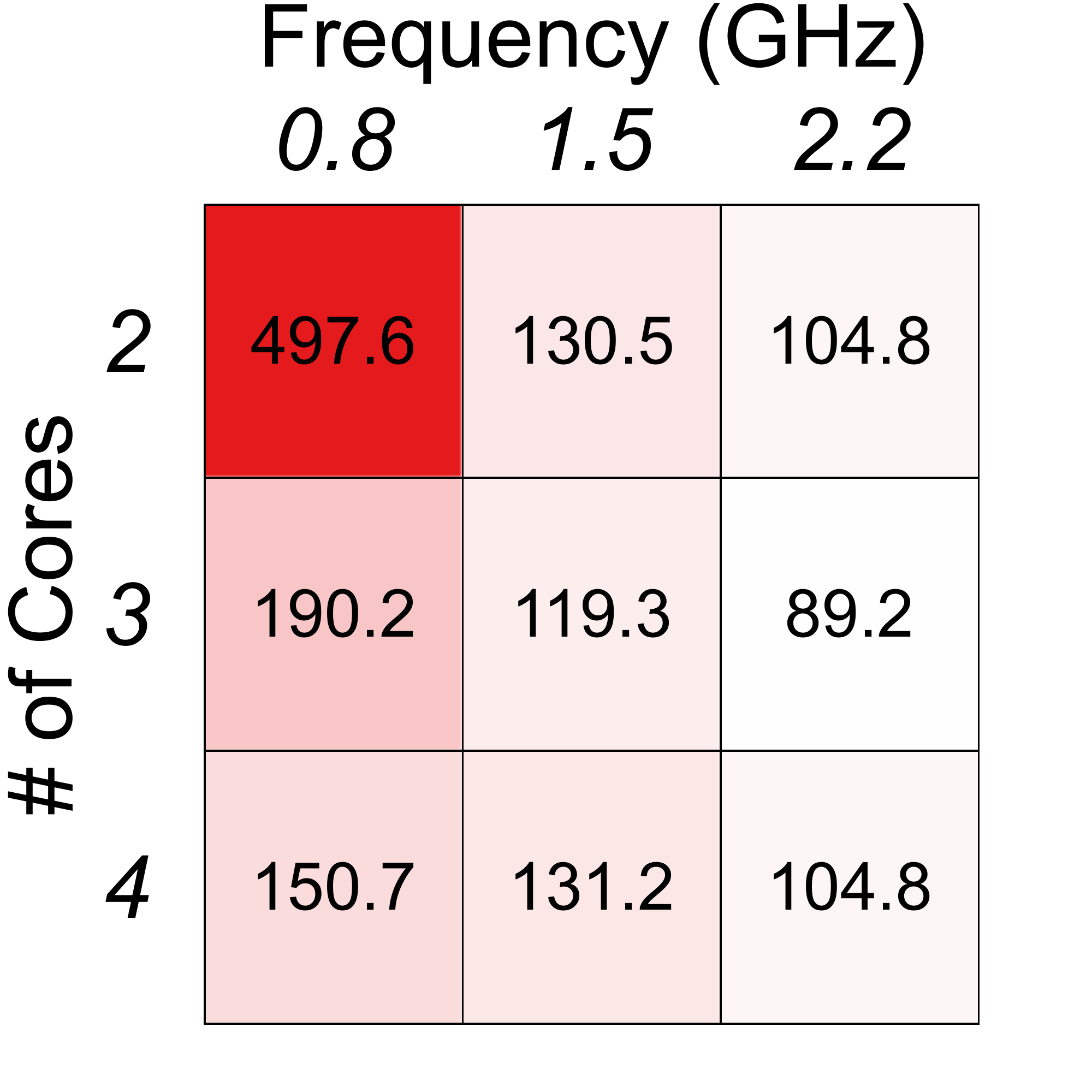}
    \caption{Energy (kJ)}
     \label{fig:benchmarks:OPA:pd:energy}
    \end{subfigure}
    \caption{Package Delivery.}
    \label{fig:benchmarks:OPA:pd}
    \end{figure}%
    \begin{figure}[t!]
    \centering
    \begin{subfigure}[t!]{.3\columnwidth}
    \centering
    \includegraphics[width=\columnwidth]{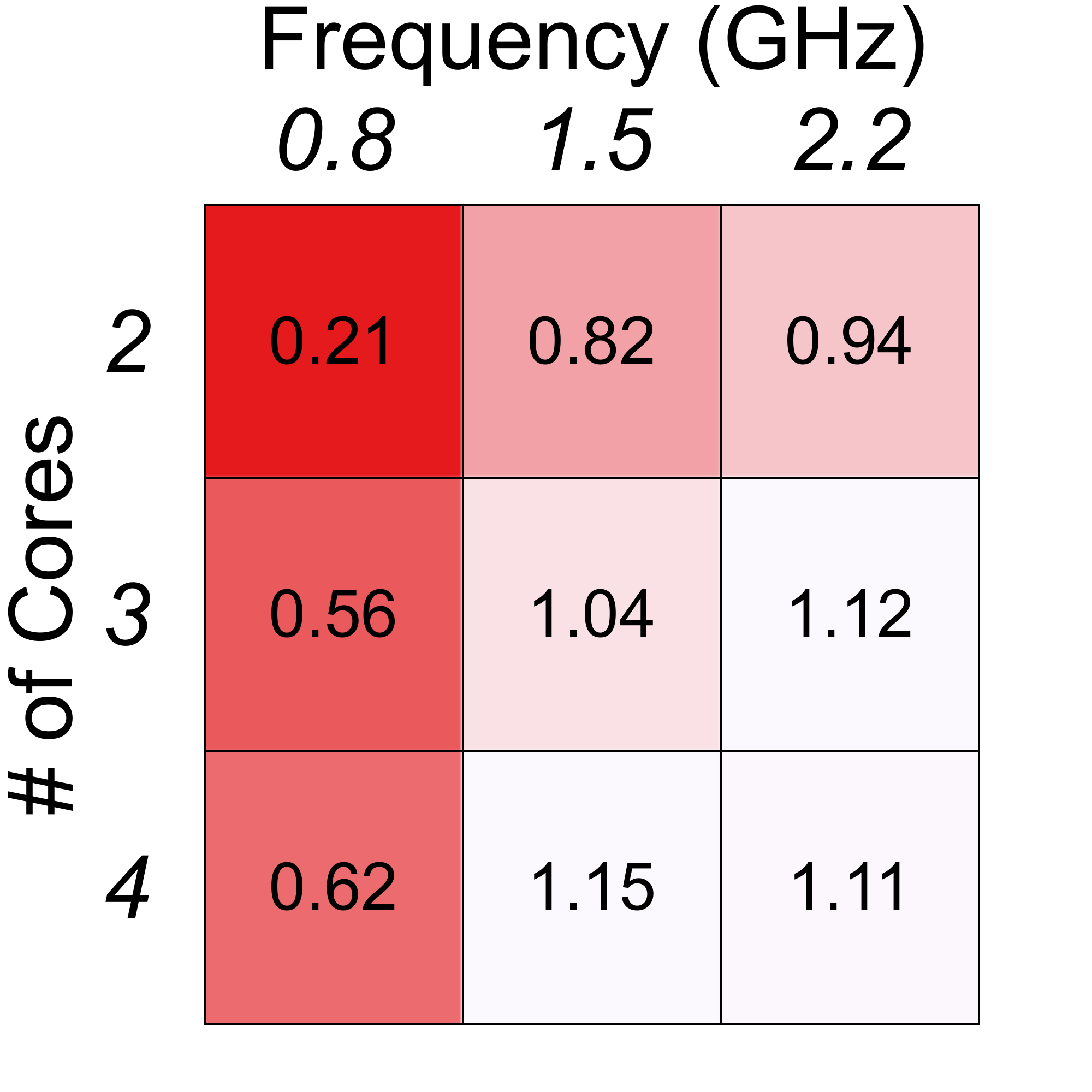}
    \caption{Velocity (m/s)}
     \label{fig:benchmarks:OPA:mapping:velocity}
    \end{subfigure}
    \begin{subfigure}[t!]{.3\columnwidth}
    \centering
    \includegraphics[width=\columnwidth]{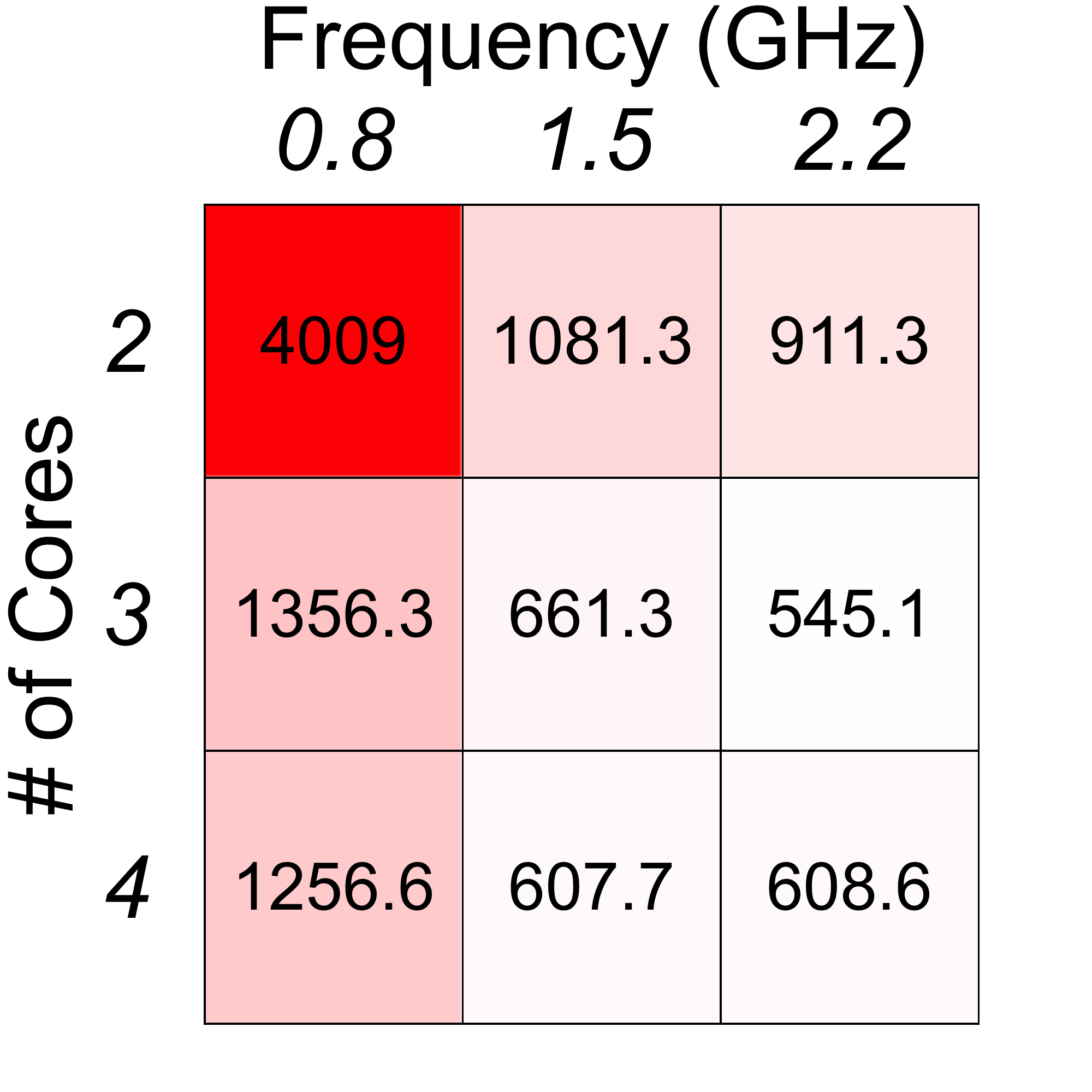}
    \caption{Mission Time (s)}
    \label{fig:benchmarks:OPA:mapping:time}
    \end{subfigure}
    \begin{subfigure}[t!]{.3\columnwidth}
    \centering
    \includegraphics[width=\columnwidth] {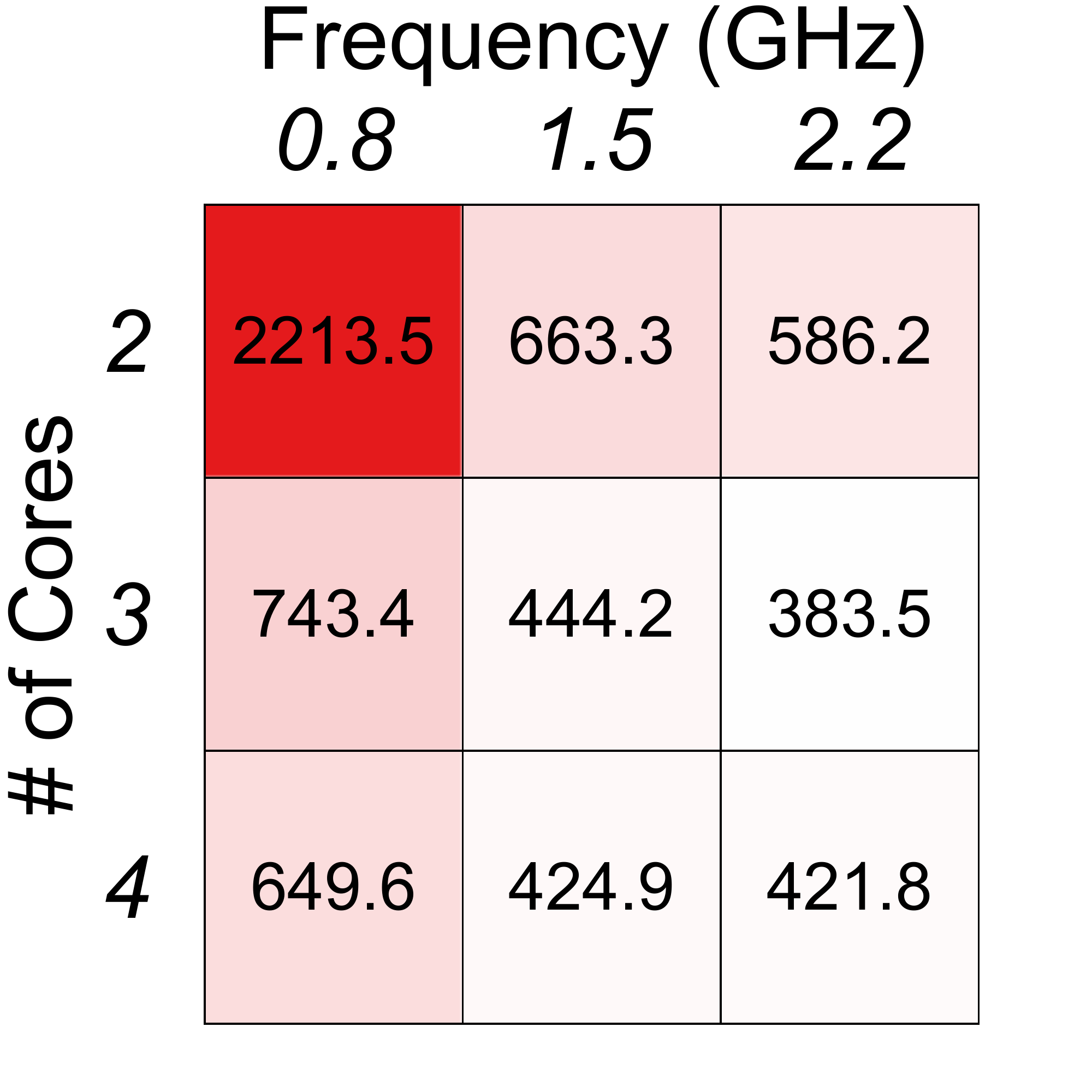}
    \caption{Energy (kJ)}
     \label{fig:benchmarks:OPA:mapping:energy}
    \end{subfigure}
    \caption{Mapping.}
    \label{fig:benchmarks:OPA:mapping}
    \end{figure}%
    \begin{figure}[t!]
    \centering
    \begin{subfigure}[t!]{.3\columnwidth}
    \centering
    \includegraphics[width=\columnwidth]{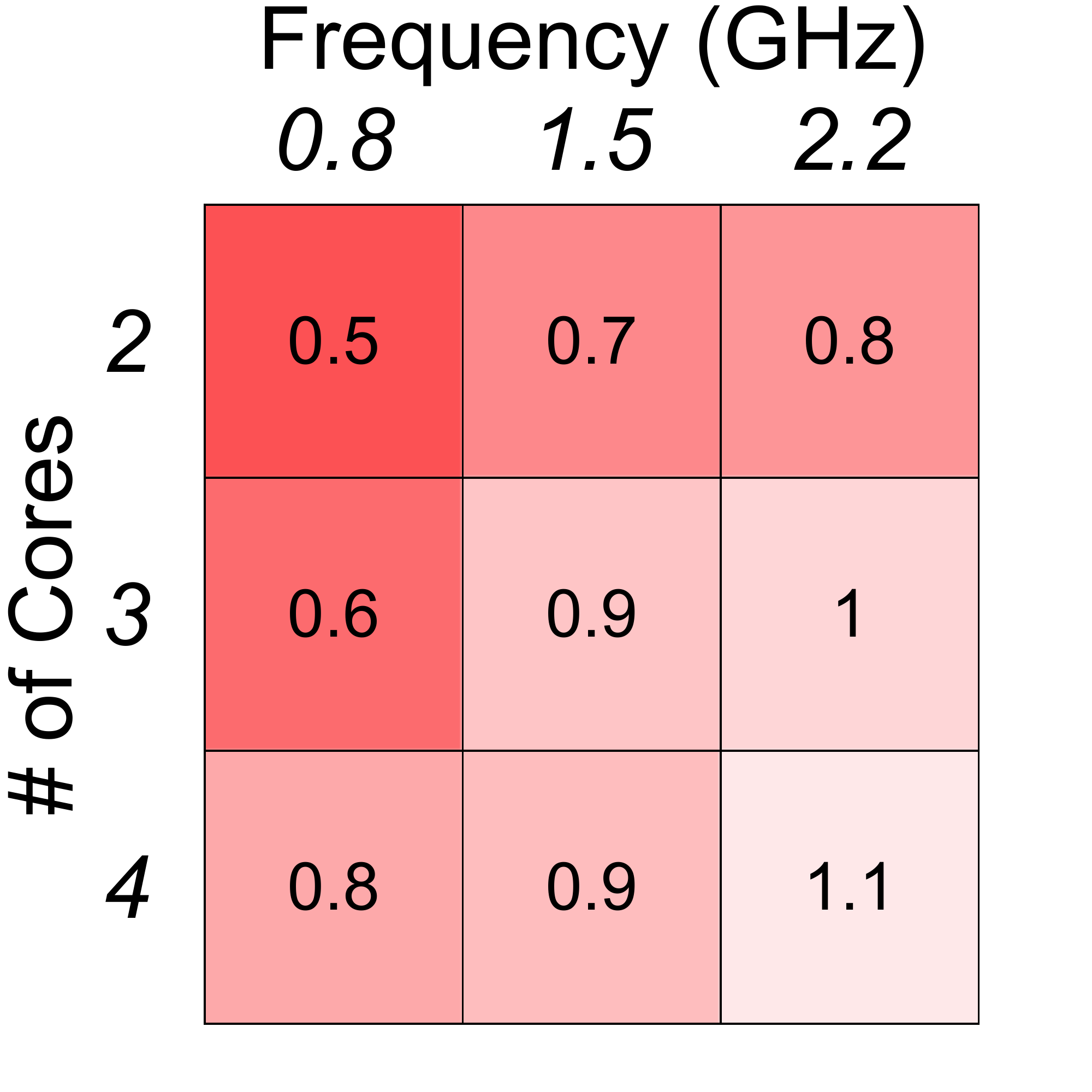}
    \caption{Velocity (m/s)}
     \label{fig:benchmarks:OPA:sar:velocity}
    \end{subfigure}
    \begin{subfigure}[t!]{.3\columnwidth}
    \centering
    \includegraphics[width=\columnwidth]{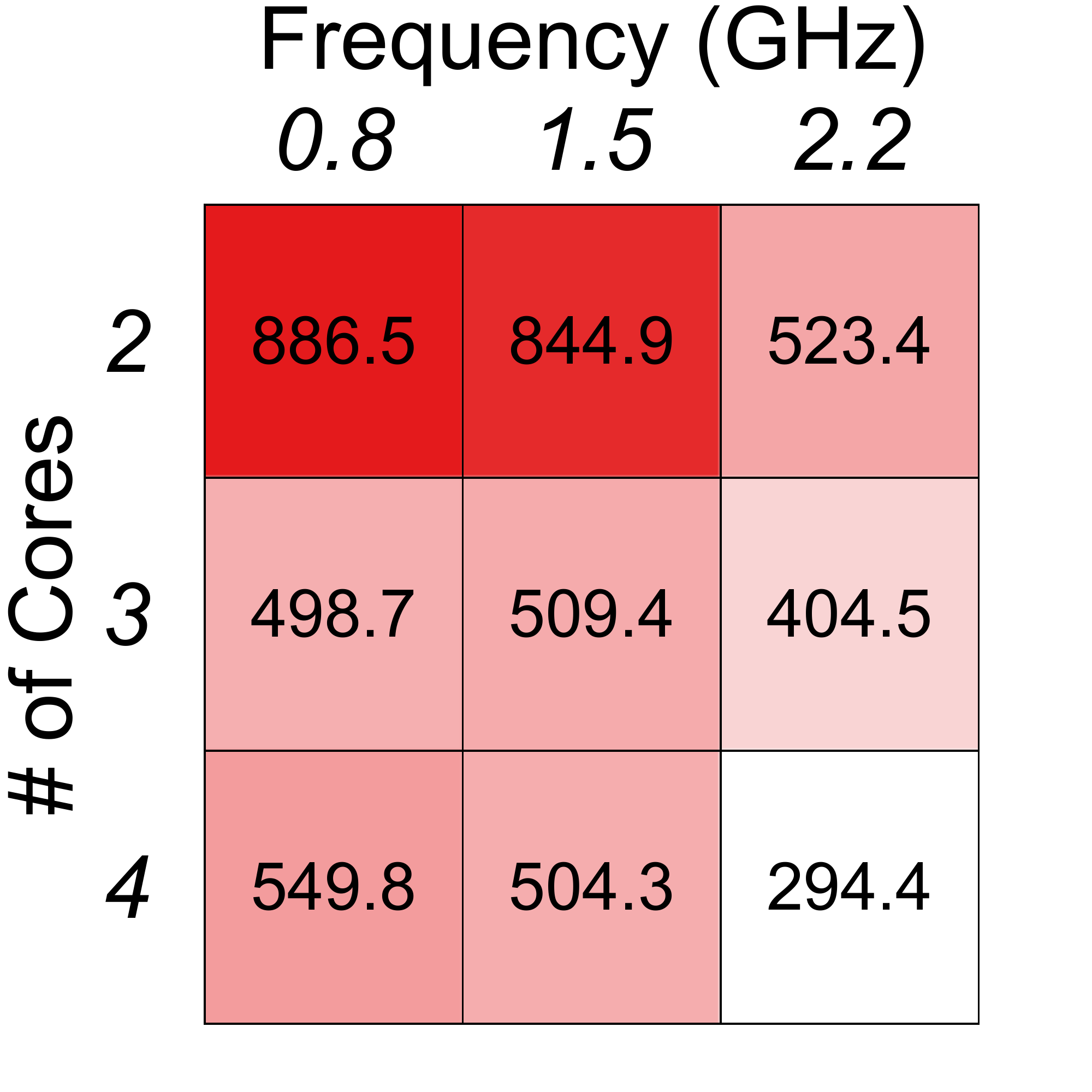}
    \caption{Mission Time (s)}
    \label{fig:benchmarks:OPA:sar:time}
    \end{subfigure}
    \begin{subfigure}[t!]{.3\columnwidth}
    \centering
    \includegraphics[width=\columnwidth] {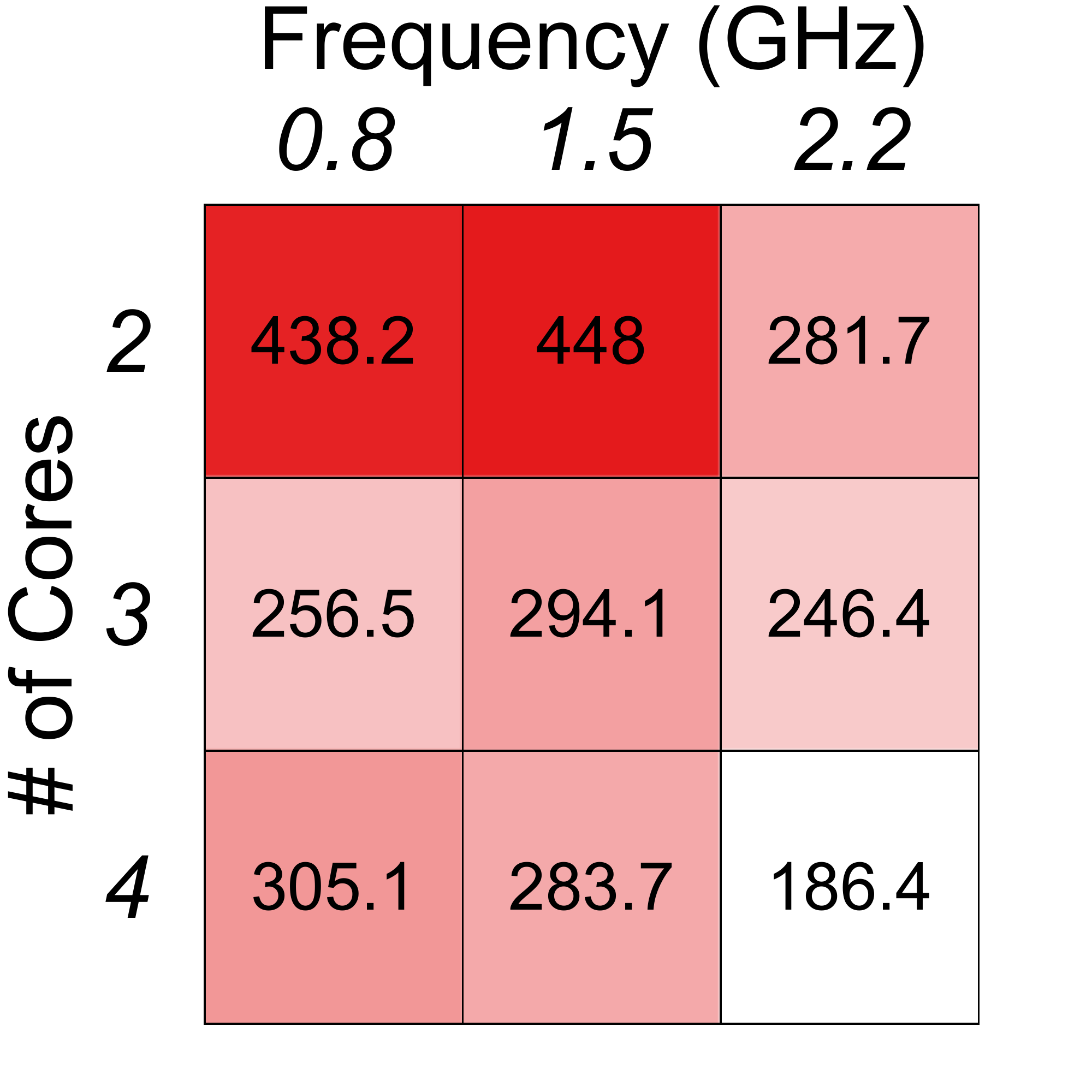}
    \caption{Energy (kJ)}
     \label{fig:benchmarks:OPA:sar:energy}
    \end{subfigure}
    \caption{Search and Rescue.}
    \label{fig:benchmarks:OPA:sar}
    \end{figure}
    \begin{figure}[t!]
    \centering
    \begin{subfigure}[t!]{.3\columnwidth}
    \centering
    \includegraphics[width=\columnwidth]{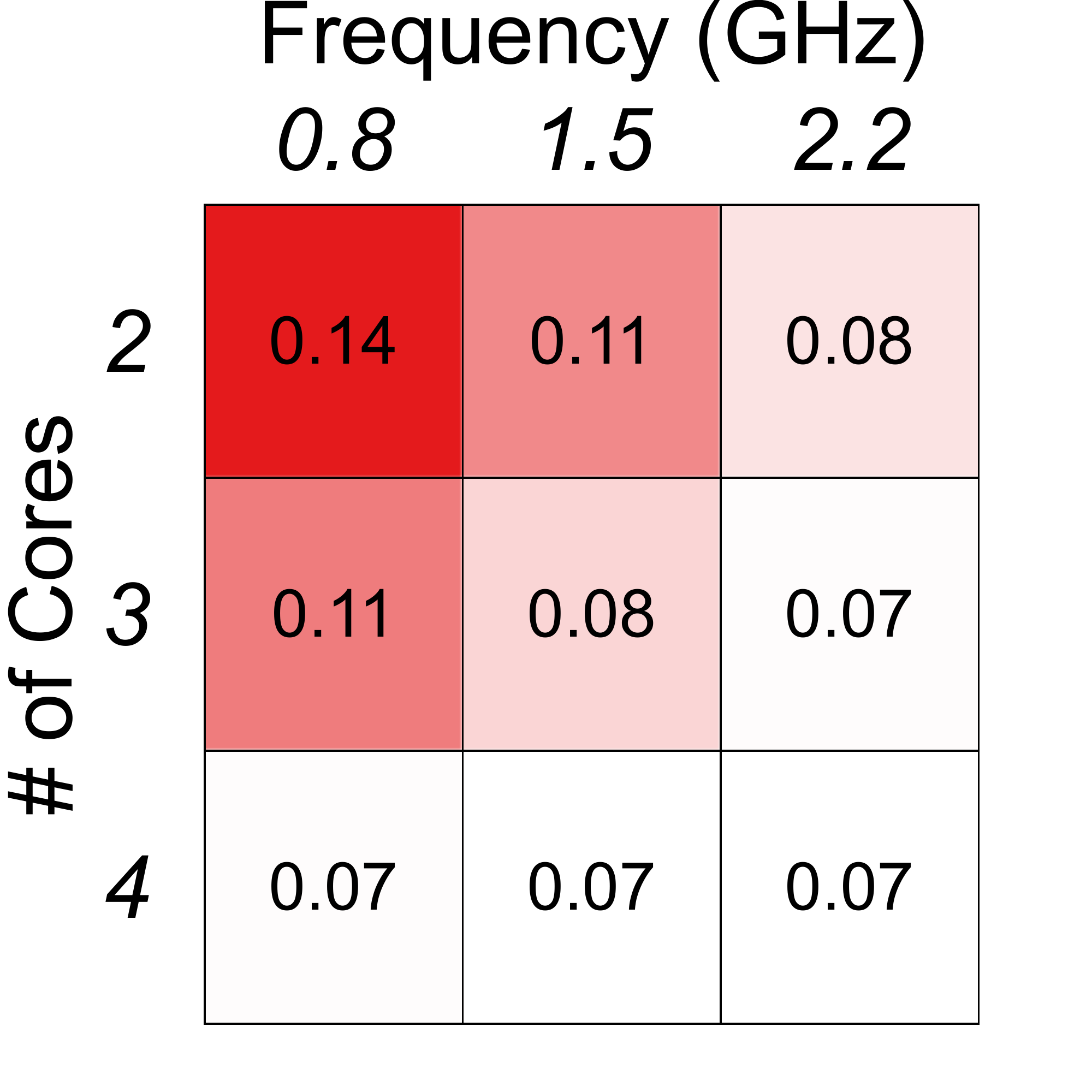}
    \caption{Error Rate(m/s)}
     \label{fig:benchmarks:OPA:ap:velocity}
    \end{subfigure}
    \begin{subfigure}[t!]{.3\columnwidth}
    \centering
    \includegraphics[width=\columnwidth]{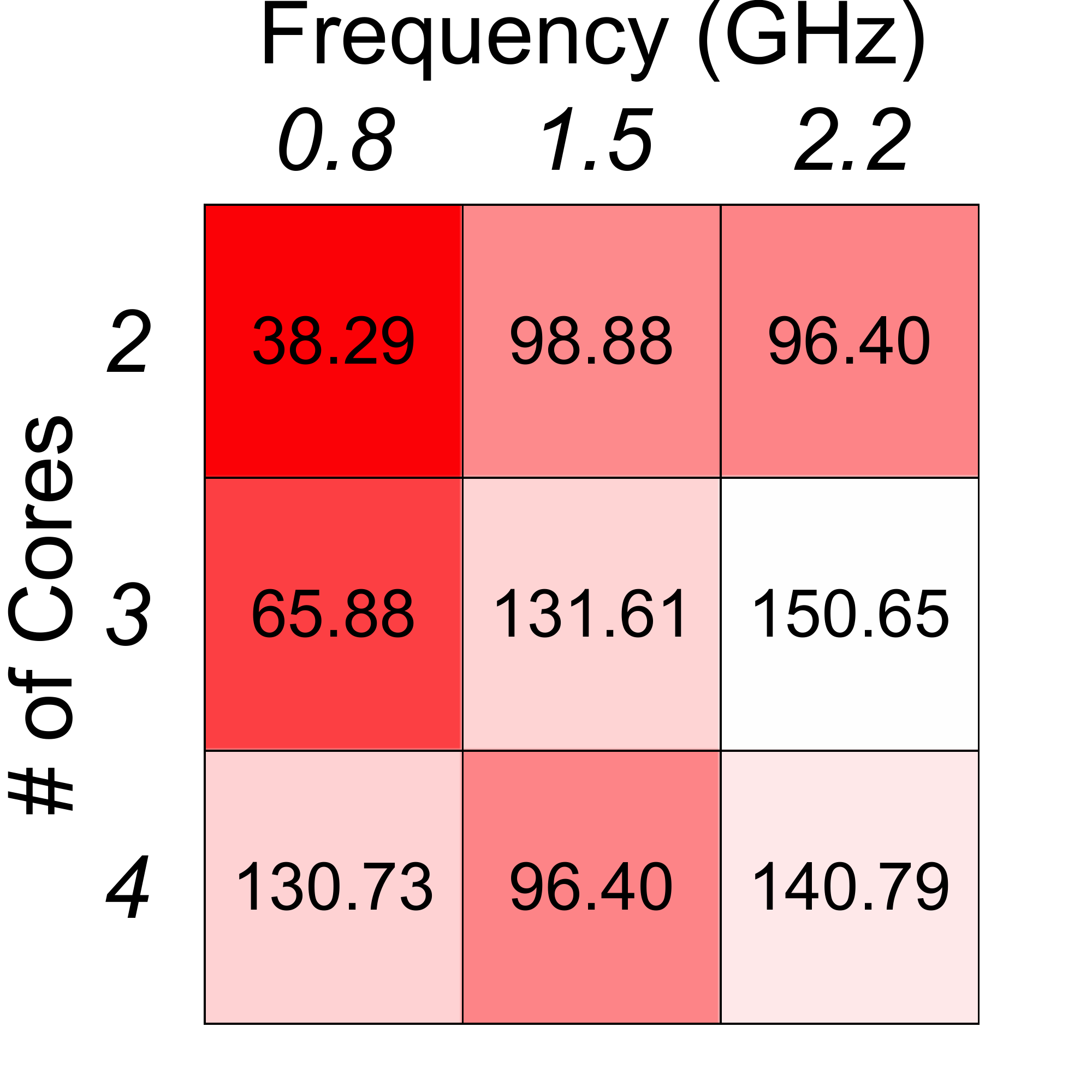}
    \caption{Mission Time(s)}
    \label{fig:benchmarks:OPA:ap:time}
    \end{subfigure}
    \begin{subfigure}[t!]{.3\columnwidth}
    \centering
    \includegraphics[width=\columnwidth] {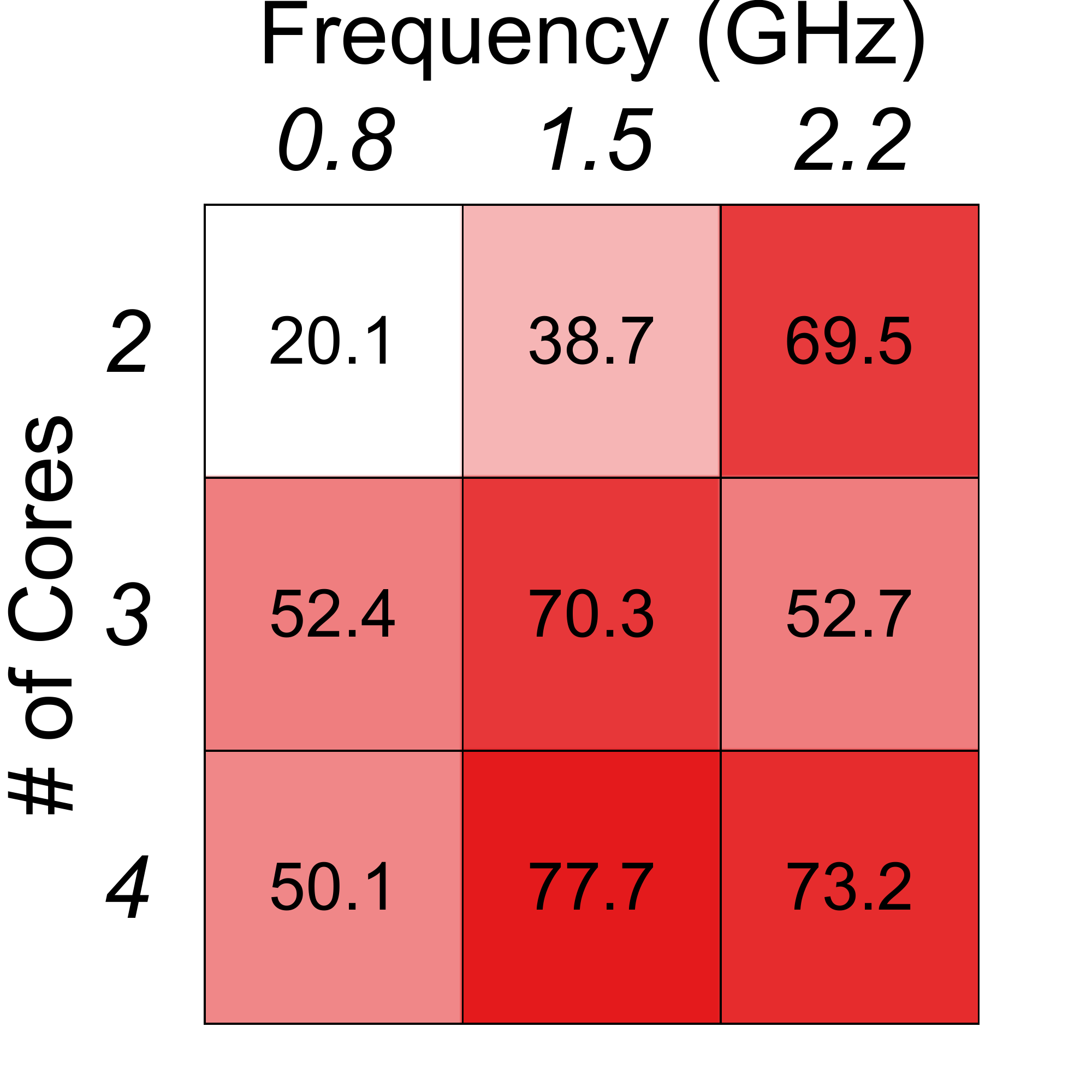}
    \caption{Energy (kJ)}
     \label{fig:benchmarks:OPA:ap:energy}
    \end{subfigure}
    \caption{Aerial Photography.}
    \label{fig:benchmarks:OPA:ap}
\end{figure}
}

\paragraph{Mapping:} We observe a reduction of up to 86\% and 83\% for the mission time and energy consumption, respectively, as compute scales with the number of cores and/or frequency values (\Fig{fig:benchmarks:OPA:mapping:velocity}, \Fig{fig:benchmarks:OPA:mapping:time}, and \Fig{fig:benchmarks:OPA:mapping:energy}). The concurrency present in this application (all nodes denoted by circles with a filled arrow connection or none at all in \Fig{fig:benchmarks:data-flow:mapping} run in parallel) justifies the performance boost from core scaling. The sequential bottlenecks, i.e. motion planning and OctoMap generation explains the frequency scaling improvements. We achieve up to 6.3X improvement in motion planning (\Fig{fig:kernel-breakdown}) and that leads to hover time reduction. We achieve a 6X improvement in OctoMap generation and that leads to maximum velocity improvement. The improvements translate to a 5.3X improvement in average velocity. Since mission time is reduced, total energy consumption reduces.

\paragraph{Search and Rescue:} We see a reduction of up to 67\% and 57\% for the mission time and the energy, respectively, as compute scales (\Fig{fig:benchmarks:OPA:sar:velocity}, \Fig{fig:benchmarks:OPA:sar:time}, and \Fig{fig:benchmarks:OPA:sar:energy}). Similar to mapping, more compute allows for the reduction of hover time and an increase in maximum velocity which contribute to the overall reduction in mission time and energy. In addition, a faster object detection kernel prevents the drone from missing sampled frames during any motion. We achieve up to 1.8X, 6.8X, and 6.6X speedup for the object detection, motion planning and OctoMap generation kernels, respectively. In aggregate, these improvements translate to 2.2X improvement in the MAV's average velocity.

\paragraph{Aerial Photography:} We observe an improvement of up to 53\% and 267\% for \emph{error} and mission time, respectively (\Fig{fig:benchmarks:OPA:ap:velocity}, \Fig{fig:benchmarks:OPA:ap:time}, and  \Fig{fig:benchmarks:OPA:ap:energy}). In aerial photography, as compared to other applications, higher mission time is more desirable than a lower mission time. The drone only flies while it can track the person, hence a longer mission time means that the target has been tracked for a longer duration. In addition to maximizing the mission time, error minimization is also desirable for this application. We define error as the distance between the person's bounding box (provided by the detection kernel) center to the image frame center.  Clock and frequency improvements translate to 2.49X and 10X speedup for the detection and tracking kernels and that allows for longer tracking with a lower error. No significant trend is observed in the energy data because energy depends on both the mission time and the velocity, and as opposed to the other applications, there is no need for the drone to minimize its velocity. Instead, it needs to successfully track the person.

\begin{figure*}[!t]
\centering
\includegraphics[width=\columnwidth]{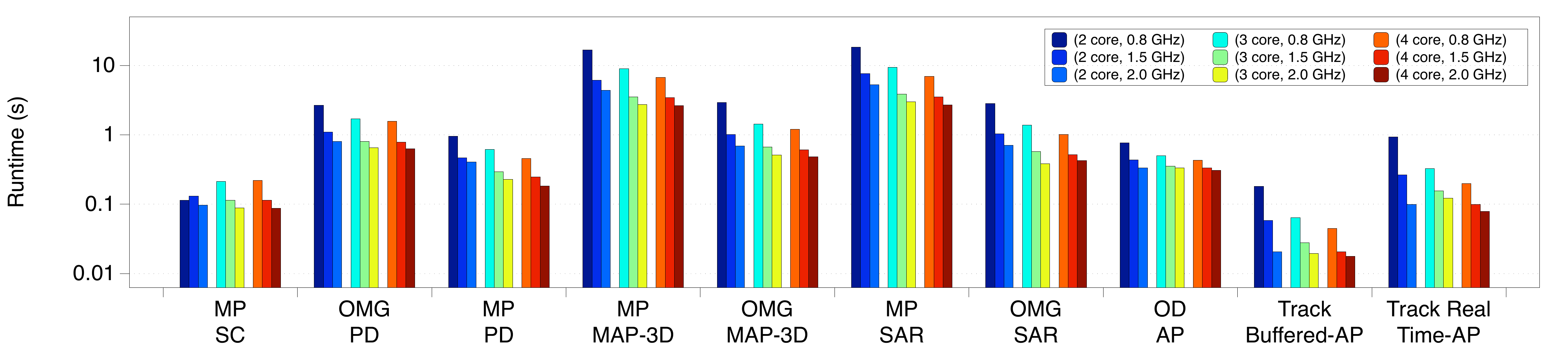}
\vspace{-18pt}
\caption{Kernel breakdown for MAVBench. The abbreviations are as follows: \emph{OD-}Object detection, \emph{MP-}Motion Planning, \emph{OMG-}OctoMap Generation for kernels and \emph{SC-}Scanning, \emph{PD-}Package Delivery,  \emph{MAP-}3D Mapping, \emph{SAR-}Search and Rescue, and \emph{AP-}Aerial Photography for applications. The $x$-axis lists the kernel-application names and $y$-Axis represents the runtime in seconds. Each bar graph represents one of the configurations used in the hardware. The cores are varied from 2 to 4 and the frequency goes from from 0.8 GHz , 1.5 GHz or 2.2 GHz.}
\label{fig:kernel-breakdown}
\end{figure*}

\label{sec:energy-case-study}
\section{Case Studies}
\label{sec:case-study}

We show how our closed-loop simulation, along with our open-source benchmark suite, can enable (1) performance, (2) energy and (3) reliability studies, both at the architecture and the holistic system-level. MAVBench simulation setup allows for inspection of intra-system, as well as system and environment interactions. Studying the intra-system interactions opens new avenues for hardware and software co-design, which we demonstrate in the form of onboard/edge-only compute versus offloading the compute to the unconstrained cloud. Studying the system and environment interactions can unlock new opportunities for trade-offs in computational complexity for energy efficiency, which we demonstrate using software-guided, hardware-assisted OctoMap resolution optimizations.

\subsection{A Performance Case Study}

As opposed to the study in Section~\ref{sec:char} where we emulated a fully-on-edge drone (i.e., a drone which all of its computation is done on the drone itself), we examine a cloud/edge drone where the computation is distributed across the edge and the cloud. We compare a fully-on-edge drone equipped with a TX2 versus a fully-in-cloud drone with a powerful cloud support. The ``cloud'' computational horsepower is composed of an Intel i7 4740 @ 4GHz with 32 GB of RAM and a GeForce GTX 1080. For network connectivity we utilize a 1Gbp/s LAN, which mimics a future 5G network~\cite{agyapong2014design, gupta2015survey}.

\begin{figure}[b!]
\centering
    \begin{subfigure}{.608\columnwidth}
    \centering
    \includegraphics[trim=5 0 20 0, clip, width=\columnwidth]{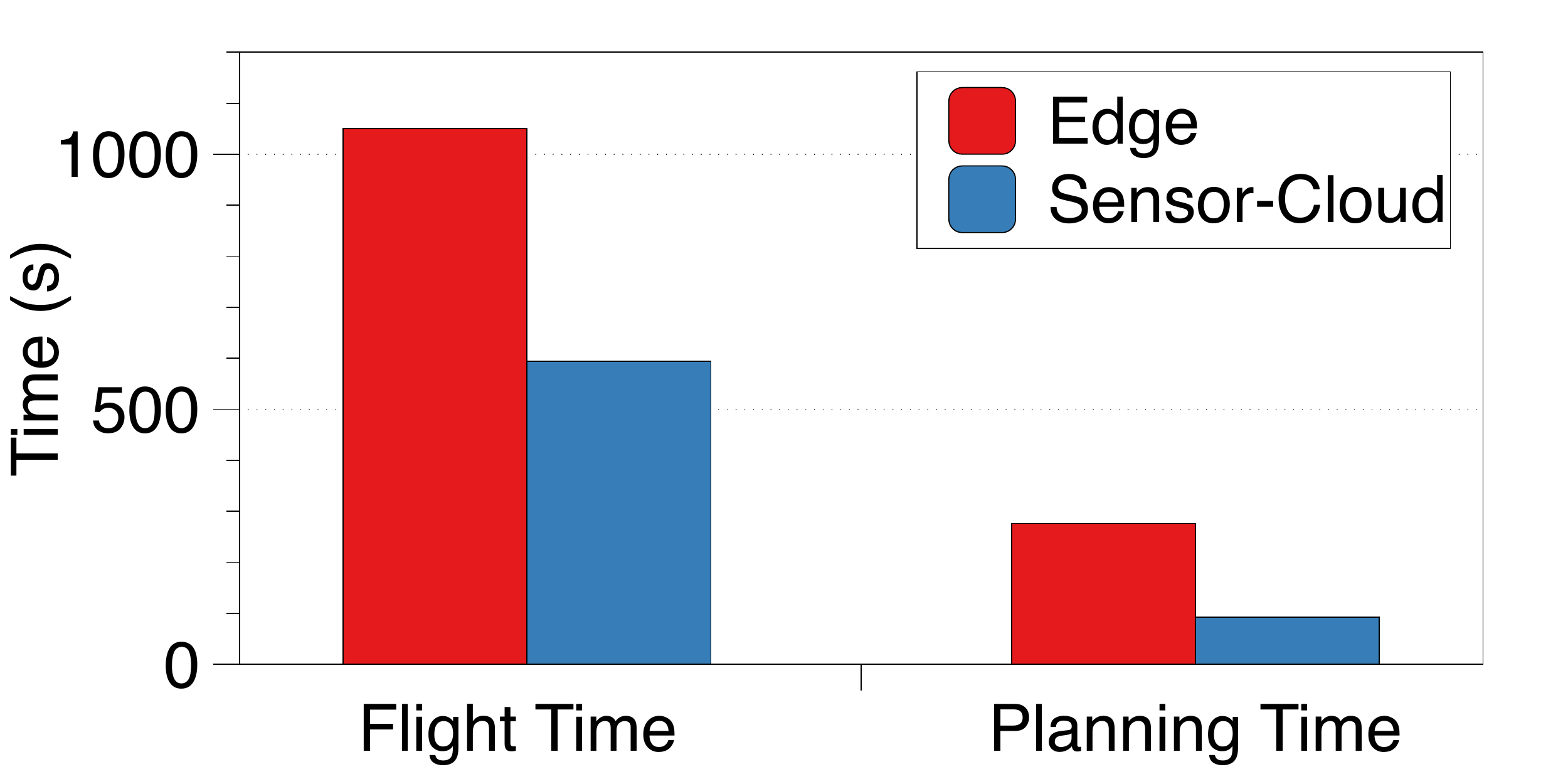}
    \caption{Performance.}
    \label{Perforamnce}
    \end{subfigure}
    \hfill
    \begin{subfigure}{.365\columnwidth}
    \centering
    \includegraphics[trim=5 10 0 0, clip, width=\columnwidth]{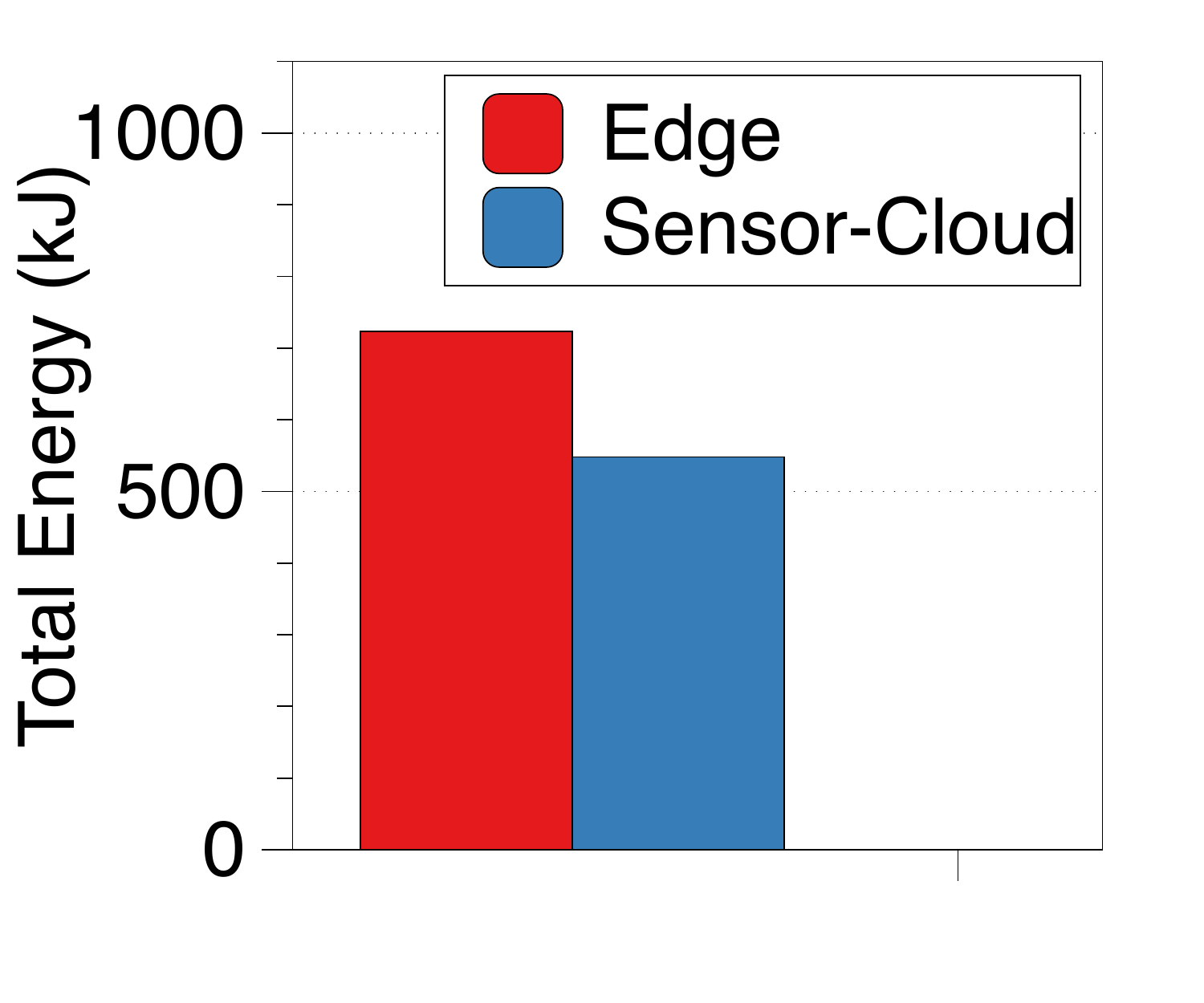}
    \caption{Energy.}
    \label{fig:drone-power-time-series}
    \end{subfigure}
\caption{Comparing a full-on-edge drone versus a full-on-cloud drone. Our system allows part or portion of the MAVBench workloads to be offloaded to the cloud (or another local co-processing agent).}
\label{fig:cloud_edge}
\end{figure}

We target the planning stage of the PPC pipeline and focus on the 3D Mapping as the application of choice to offload. As we show in \Fig{fig:cloud_edge}, a drone that can enjoy the cloud's extra compute power sees a 3X speed up in planning time. This improves the drone's average velocity due to hover time reduction, and hence reduces the drone's overall mission time by as much as 50\%, effectively doubling its endurance.

\begin{figure*}[!t]
\vspace{-12pt}
    \centering
    \begin{subfigure}[t]{.24\linewidth}
        \centering
        \includegraphics[height=1.4in]{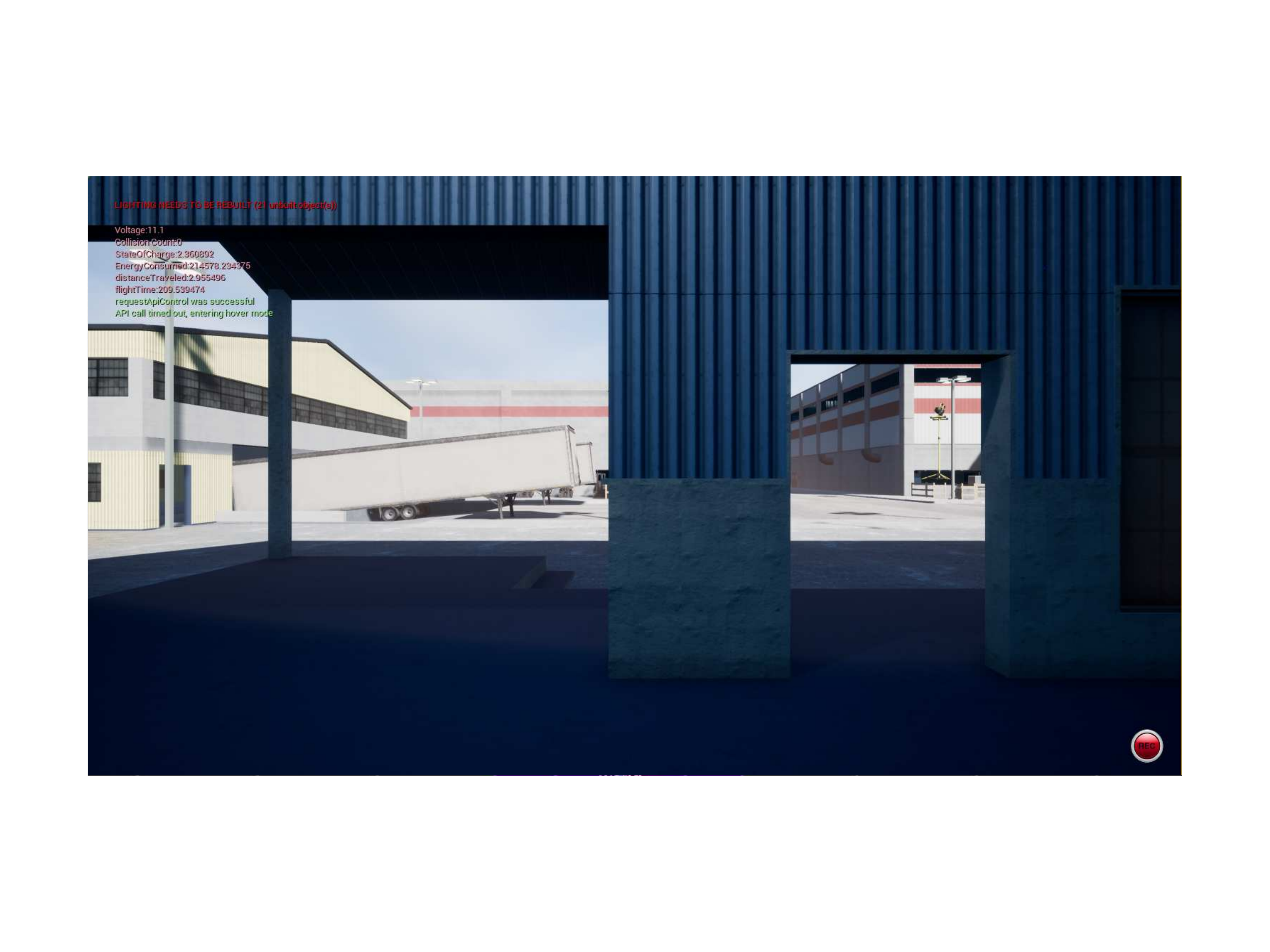}%
         \vspace{-20pt}
        \caption{Environment's map.}%
        \label{fig:garage_sim}
    \end{subfigure}
    \hfill
    \begin{subfigure}[t]{.24\linewidth}
        \centering
        \includegraphics[height=1.4in]{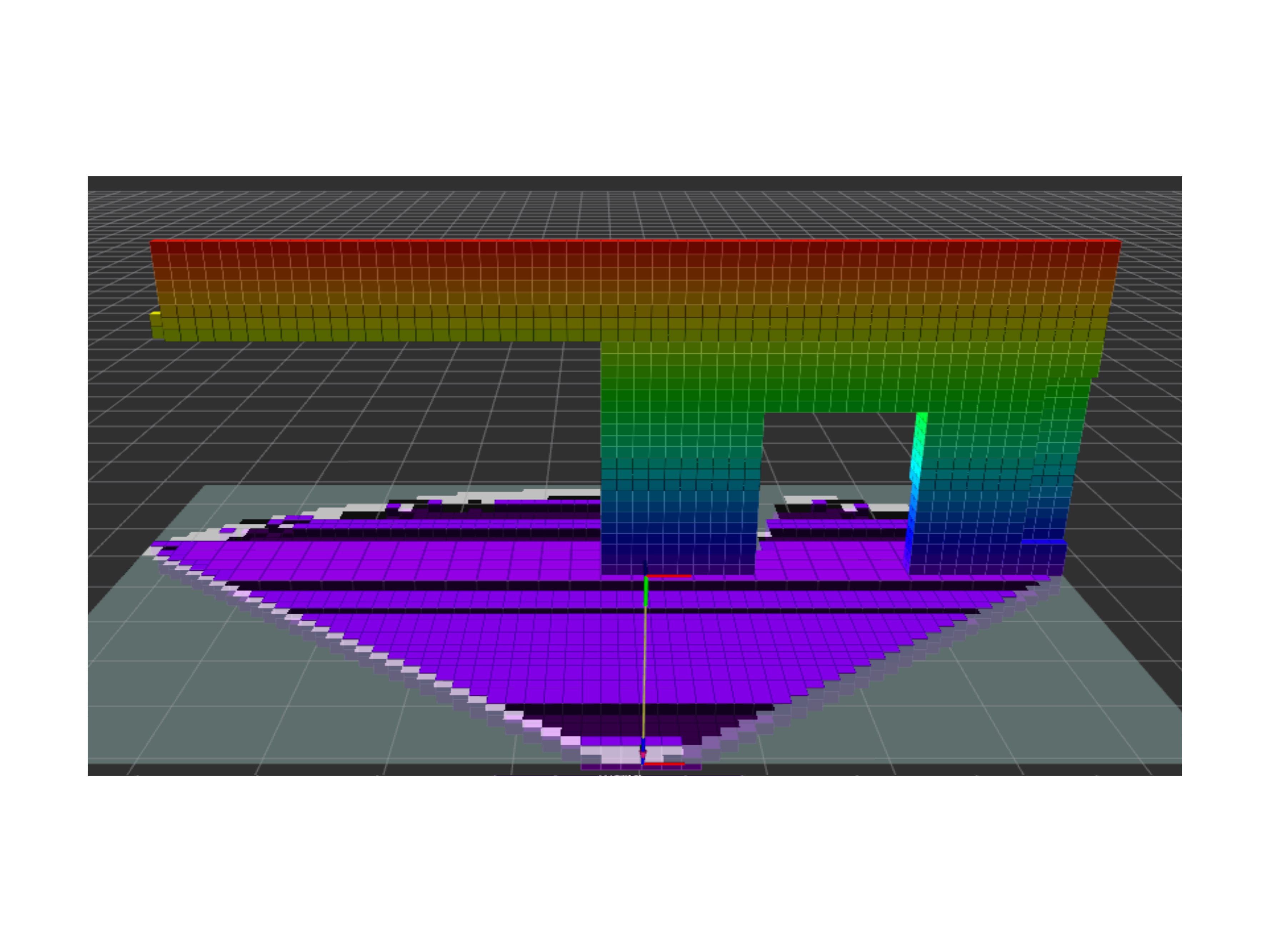}%
         \vspace{-20pt}
        \caption{Resolution of 0.15 \emph{(m)}.}%
        \label{fig:rviz__15}
    \end{subfigure}
    \hfill
     \begin{subfigure}[t]{.24\linewidth}
        \centering
        \includegraphics[height=1.4in]{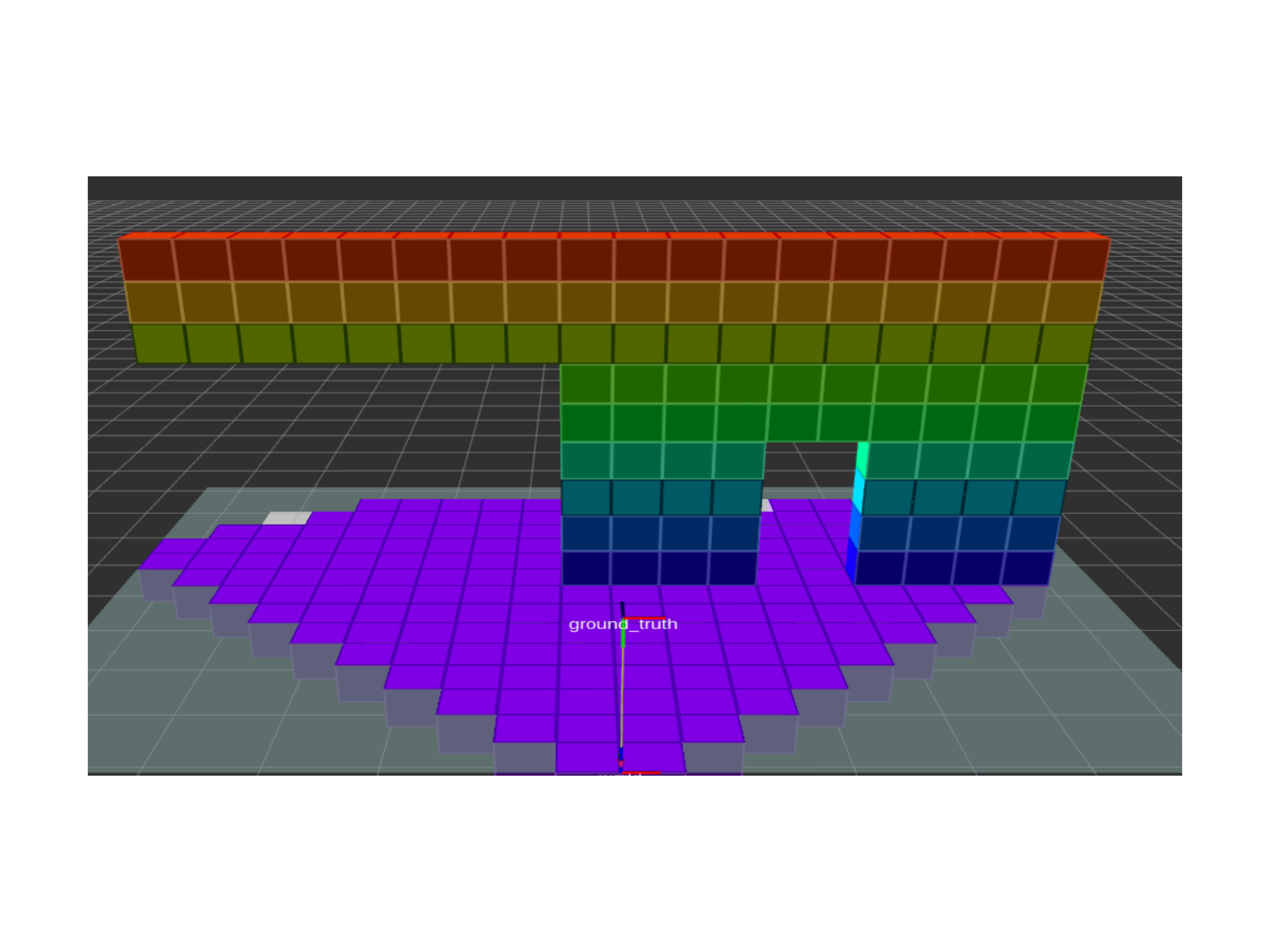}%
      \vspace{-20pt}
     \caption{Resolution of 0.5 \emph{(m)}.}%
        \label{fig:game:rviz__5}
    \end{subfigure}
       \hfill
   \begin{subfigure}[t]{.24\linewidth}
        \centering
        \includegraphics[height=1.4in]{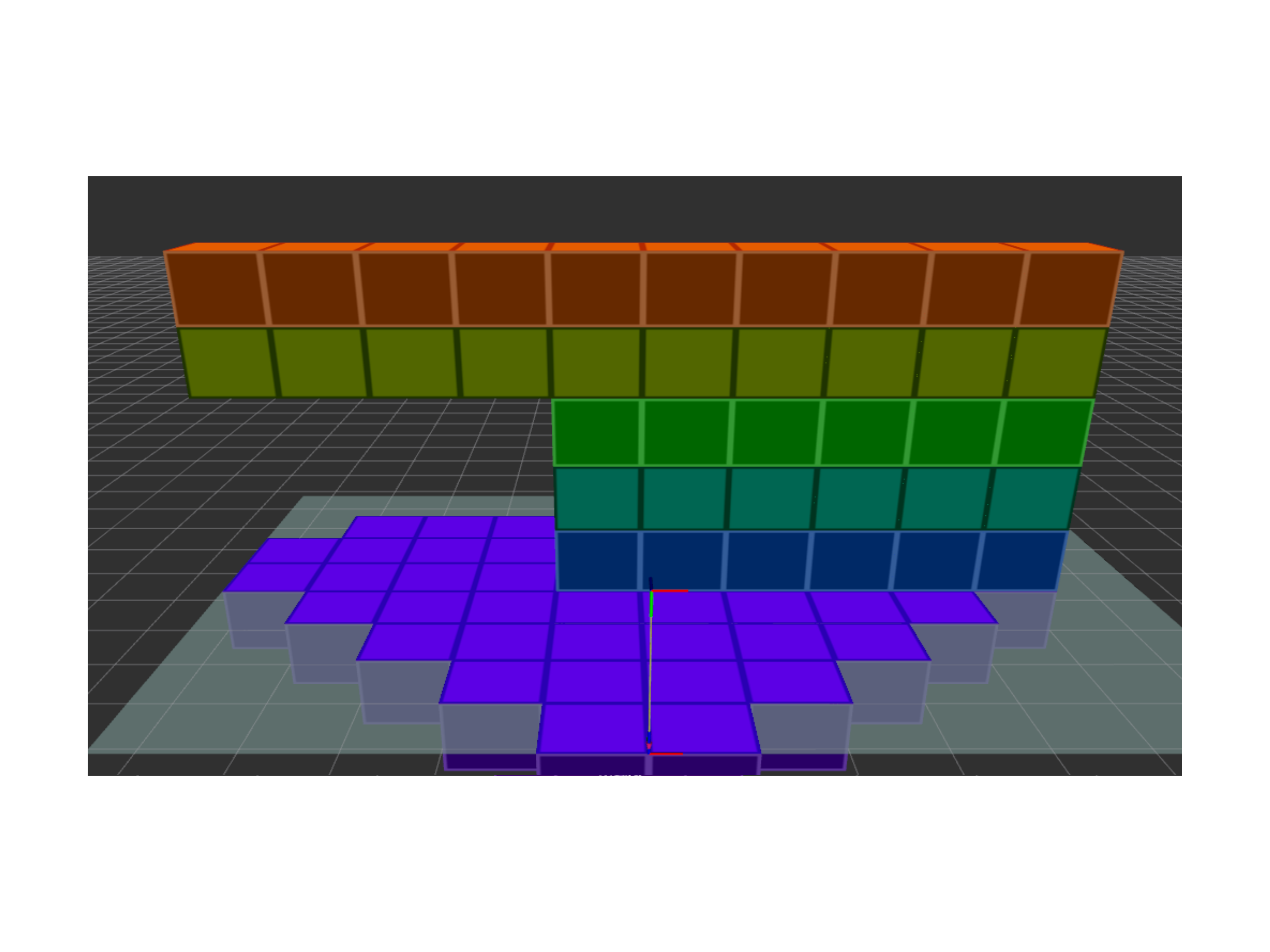}%
        \vspace{-20pt}
        \caption{Resolution of 0.80 \emph{(m)}.}%
        \label{fig:game:rviz_1}
    \end{subfigure}
    \hfill\\[15pt]
   \vspace{-25pt}
   \caption{For the environment in \emph{(a)}, OctoMap's resolution impact on the drone's perception of its environment is shown in \emph{(b)}, \emph{(c)}, \emph{(d)}.}
    \label{fig:octomap_perception}
\end{figure*}

\begin{figure}[t]
\vspace{0pt}
\centering
\includegraphics[trim=0 0 30 0, clip, width=\columnwidth]
{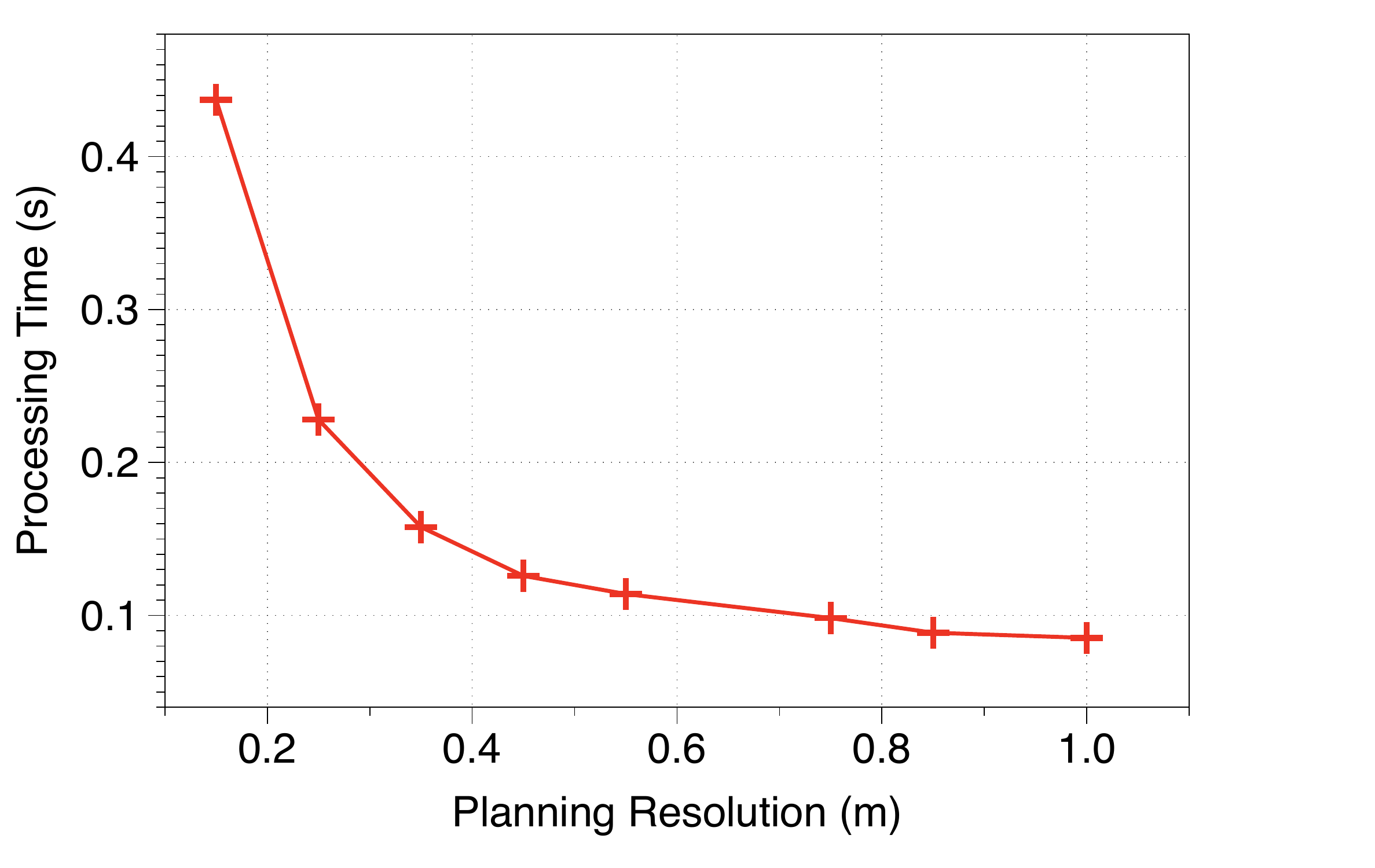}
\vspace{-23pt}
\caption{Reduction in OctoMap resolution (accuracy) can be traded off with processing time. Increasing the $x$-axis means larger voxels to represent the space more coarsely (less accurately). A 6.5X reduction in resolution results in a 4.5X improvement in processing time.}
\label{fig:octrest}
\end{figure}

\subsection{An Energy Case Study} 

Focusing on energy efficiency, we conduct a kernel/environment sensitivity analysis using the OctoMap node~\cite{octomap}, which is a major bottleneck in three of our end to end applications, namely package delivery, 3D mapping and search and rescue. OctoMap is used for the modeling of various environments without prior assumptions. The map of the environment is maintained in an efficient tree-like data structure while keeping track of the free, occupied and unknown areas. Both planning and collision avoidance kernels use OctoMap to make safe flight possible, via costly compute cycles, by only allowing navigation through free space. Due to its implementation efficiency, OctoMap is widely adopted in the robotics community. Its broad adoption and impact in two out of three stages (Perception and Planning) makes this kernel highly general and important for optimization.

The size of the voxels in OctoMap, i.e. the map's resolution, introduces accuracy versus flight-time/energy trade-off. By lowering the resolution, i.e. increasing \emph{voxel} sizes, obstacle boundaries get inflated, hence the drone's perception of the environment and the objects within it becomes inaccurate. We illustrate the impact of OctoMap resolution on the drone's perception using \Fig{fig:octomap_perception}. \Fig{fig:garage_sim} {shows the environment and  Figures~\ref{fig:rviz__15}, \ref{fig:game:rviz__5}, \ref{fig:game:rviz_1} show the drone's perception of the environment as a function of OctoMap resolution. When the resolution is lowered, the voxels size increases to the point that the drone fails to recognize the openings as possible passageways to plan through (Figure~\ref{fig:game:rviz_1}). This results in mission time inefficiency and failures depending on the environment.

To examine the accuracy versus performance trade off, we measured OctoMap kernel's processing time (running in isolation) while varying its resolution knob. \Fig{fig:octrest} shows that as planning resolution increases (i.e., voxels are larger so space is represented more coarsely and hence less accurately), performance improves dramatically because less compute is needed. Going from one extreme to another, when planning resolution goes from less than 0.2~m to 1.0~m ($x$-axis), OctoMap's processing time (or update rate) goes from more than 0.4~seconds to less than 0.1~seconds ($y$-axis). In other words, a 6.5X reduction in accuracy results in a 4.5X improvement in processing time.

\begin{figure}[t]
\centering
\includegraphics[trim=0 0 0 0, clip, width=\columnwidth]
{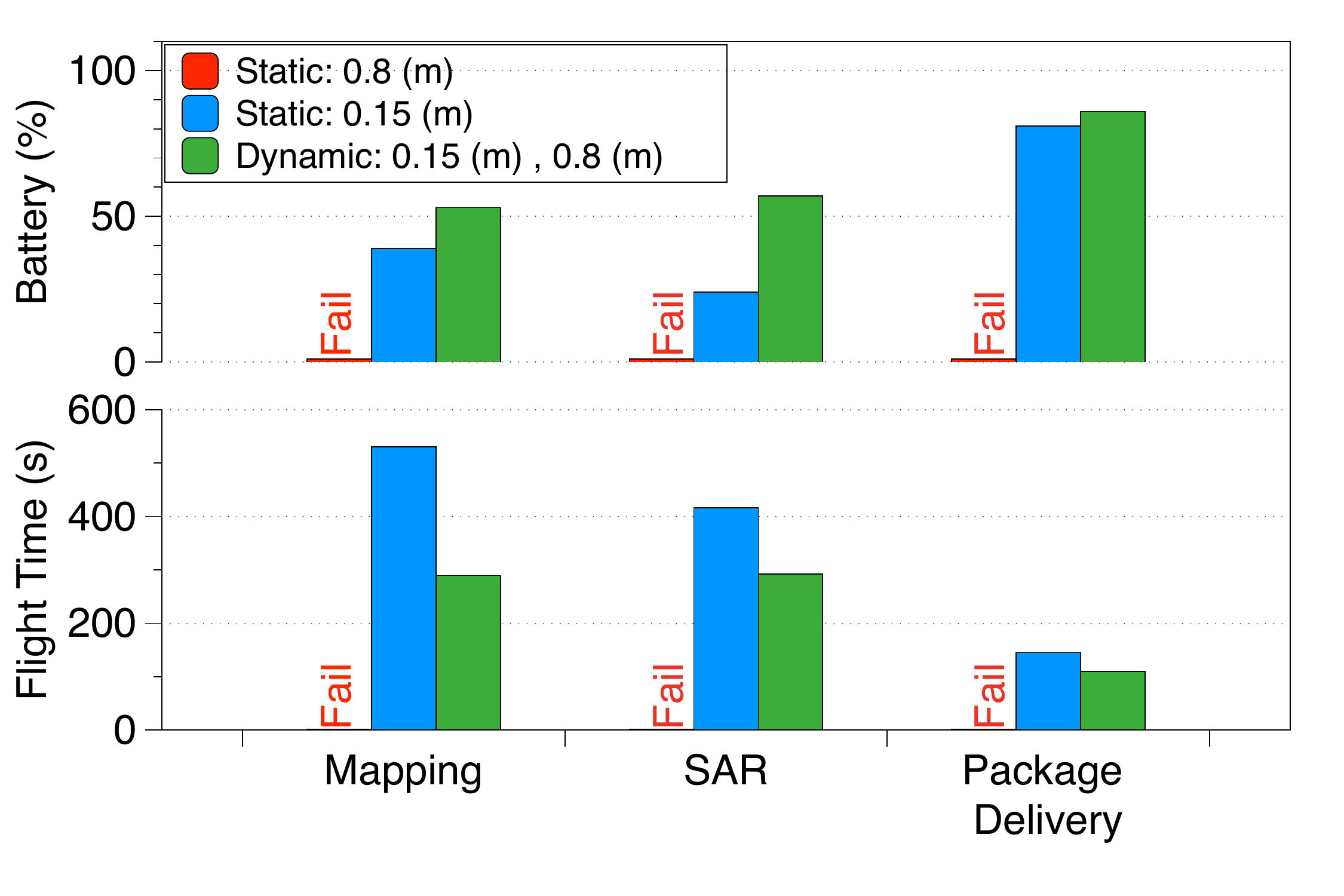}
\vspace{-23pt}
\caption{Switching between OctoMap resolutions dynamically leads to successfully finishing the mission compared to 0.80~m. It also leads to battery life improvement compared to 0.15~m. The $y$-axis in the top graph is the battery left on the drone upon mission completion.}
\label{fig:dual-res}
\end{figure}

Certain aspects like obstacle density in the environment determine the ``ideal'' OctoMap resolution. In low-density environments, where the drone has many obstacle-free paths to take, a low resolution can suffice. In dense environments, low resolutions can deprive the drone of viable obstacle-free paths because the drone perceives the obstacles to be larger than they are in the real world, and so plans to avoid them. Since the drone's environment constantly changes, a dynamic approach where a runtime sets the resolution is ideally desirable.

We study two environments during the mission, namely outdoors (low obstacle density) and indoors (high obstacle density). \Fig{fig:dual-res} shows the result of two static (predetermined) resolutions, 0.15~m and 0.80~m, and our dynamic approach that multiplexes between the two appropriately.\footnote{Resolutions are based on the environment like the door width size. A 0.15~m resolution is chosen to ensure that the drone (diagonal width of 0.65~m) considers an average door (width of 0.82~m) as an opening for planning.} The dynamic approach allows improvement of battery consumption by up to 1.8X. 
Intuitively, as compute reduces, OctoMap bottleneck eases, and therefore the drone completes its mission faster. The figure also highlights another interesting relationship that statically choosing the 0.80~m resolution to optimize for compute (only) causes the drone to fail its mission since it is unable to plan paths through narrow openings in the indoor environments. Instead, by switching between the two resolutions according to the environment's obstacle density, the dynamic approach is able to balance OctoMap computation with mission feasibility and energy, holistically. Therefore, in all cases, the dynamic approach uses less energy and retains more battery life at mission end time.


\subsection{A Reliability Case Study}

Reliability is an especially important topic in the context of autonomous vehicles~\cite{reli-1, reli-2, reli-3, reli-4}. Traditionally, it is common to study the susceptibility of execution to errors that manifest in programs and the architecture. In autonomous vehicles, errors or ``noise'' in the data can arise from sensor inputs. 

We investigate the impact of sensor noise on the performance of our package delivery application, specifically its perception stage. The data is summarized in Table~\ref{tbl:noise-data}. We inject Gaussian noise with a range of standard deviations (0 to 1.5~m) into the depth readings of the drone's RGBD camera. The sensory noise distorts the drone's perception of the obstacles in its environment, and we found such noise \emph{inflates} obstacles, making them appear larger than they are in reality. This causes the drone to re-plan its trajectories more often, as it assumes that its planned path will collide into objects that are actually further away than they seem. The more the drone re-plans its paths, the longer it takes to reach its destination, which increases it mission time by up to 90\%. It is also important to note that when noise values reach greater than a certain value, it causes the drone to fail in its mission altogether, e.g. noise with the standard deviation of 1.5~m results the drone to fail reaching its delivery destination in 10\% of its total runs. 

In addition to injecting noise in the sensor subsystem, we can also inject errors directly into the compute subsystem to ``simulate'' soft errors and transient bit flips in logic. Such a capability can be used to conduct vulnerability analysis~\cite{mukherjee2005soft}.

\section{Related Work}
\label{sec:related}

There is prior work that focuses on building simulators and benchmark suites to aid the development of autonomous MAVs. We address some shortcomings of previous approaches by providing a more integrated, end-to-end solution.

\paragraph{Simulators} Simulators are essential to the study of aerial and robotic agents. Our simulation platform is built upon Microsoft's AirSim~\cite{Airsim_paper}, a UAV simulator that uses the Unreal Game Engine to provide accurate physics models and photo-realistic environments. MAVBench uses the AirSim core and extends it with performance, power and battery models that are suited for architectural research, as well as with a gimbal, and dynamic and static obstacle creation capabilities that are not inherently part of AirSim. Another very popular simulator used in the robotics community for MAVs is Gazebo~\cite{gazebo}. However, Gazebo simulations lack photo-realism, while our work, with the help of AirSim and the Unreal Game Engine, enables more accurate visual modeling.

There are also numerous simulators widely used in industry and academia for studying autonomous agents such as ~\cite{boss-cmu,talos-mit,odin-vtech,uber-simulator, uav_benchmark_simulator}. However, they either do not provide MAV models or does not consider the architectural insights.


A recent work FlightGoggles~\cite{flight-goggles}, creates virtual reality environments for drones using the images streamed from the Unity3D game engine. However, for maximum realism, FlightGoggles requires a fully functioning drone that must fly during tests, with its sensory data being streamed in from the game engine. MAVBench, on the other hand, does not have this constraint. Our users may provide real processors for hardware-in-the-loop simulation, but they are not required to fly the MAVs physically in the real world.

\paragraph{Benchmarks} Most robot benchmark suites target individual computational kernels, such as odometry or motion-planning, rather than characterizing end-to-end applications composed of many different kernels. For example SLAMBench~\cite{slambench} and CommonRoad~\cite{common-road} solely focus on the perception and the planning stage respectively. However, our benchmarks allows for holistic studies by providing end-to-end applications. 

\renewcommand*{\arraystretch}{1.2}
\begin{table}[t!]
\vspace{10pt}
\caption{Impact of introducing depth image noise into the RGBD camera system on the drone's performance. Introducing noise into the drone's visual subsystem results in more frequent (re-)planning, which increases mission time and can also result in mission failures.}
\label{tbl:noise-data}
\resizebox{1.0\columnwidth}{!}{
\begin{tabular}{|c|c|c|c|}
\hline
\textbf{Noise Std {(m)}} & \textbf{Failure Rate {(\%)}} & \textbf{Number of Re-plans} & \textbf{Mission Time {(s)}} \\ \hline
\rowcolor{red!10}0.0             & 0                 & 2                  & 72               \\ \hline
\rowcolor{red!25}0.5            & 0                 & 3                  & 82               \\ \hline
\rowcolor{red!50}1.0             & 0                 & 4                  & 95               \\ \hline
\rowcolor{red!100}1.5           & 10                & 8                  & 137              \\ \hline
\end{tabular}
}
\end{table}
\renewcommand*{\arraystretch}{1}
\section{Conclusion}
\label{sec:conclusion}


MAVBench is a tool including a closed-loop simulation platform and a benchmark suite to probe and understand the intra-system (application data flow) and inter-system (system and environment) interactions of MAVs. This enables us to pinpoint bottlenecks and identify opportunities for hardware and software co-design and optimization. Using our setup and benchmark suite, we uncover a hidden compute to total system energy relationship where faster computers can allow drones to finish missions quickly, and hence save energy. This is because most of the drone's energy is consumed by the rotors, hence, faster compute can cut down on mission time (by increasing the max velocity and reducing the hovering time) and energy accordingly. Our insight allows us to improve MAV's battery consumption by up to 1.8X
for our OctoMap case study. 

\section*{Acknowledgements}

This material is based upon work supported by the NSF under Grant No. 1528045 and funding from Intel and Google.

\bibliographystyle{ieeetr}
\bibliography{references}

\begin{thebibliography}{10}

\bibitem{drone-sports1}
{Rachel Feltman}, ``{The Future of Sports Photography:Drones}.''
  \url{https://www.theatlantic.com/technology/archive/2014/02/the-future-of-sports-photography-drones/283896/}.

\bibitem{drone-surveillance}
{Debra R. Cohen McCullough}, ``{Unmanned Aircraft Systems(UAS) Guidebook in
  Development}.''
  \url{https://cops.usdoj.gov/html/dispatch/08-2014/UAS_Guidebook_in_Development.asp}.

\bibitem{disaster-drone}
A.~Qiantori, A.~B. Sutiono, H.~Hariyanto, H.~Suwa, and T.~Ohta, ``An emergency
  medical communications system by low altitude platform at the early stages of
  a natural disaster in indonesia,'' {\em J. Med. Syst.}, vol.~36, 2012.

\bibitem{nepal-earthquake-drone}
{James Rogers}, ``{How drones are helping the Nepal earthquake relief
  effort}.''
  \url{http://www.foxnews.com/tech/2015/04/30/how-drones-are-helping-nepal-earthquake-relief-effort.html}.

\bibitem{Amazondel:online}
``Amazon delivered its first customer package by drone.''
  \url{https://www.usatoday.com/story/tech/news/2016/12/14/amazon-delivered-its-first-customer-package-drone/95401366/}.

\bibitem{CSD-Amazon-Drone-patents}
{Arthur Holland Michel}, ``{Amazon's Drone Patents}.''
  \url{http://dronecenter.bard.edu/amazon-drone-patents/}.

\bibitem{drone-package-delivery-google}
{BBC News}, ``{Google Plans Drone Delivery Service for 2017}.''
  \url{http://www.bbc.com/news/technology-34704868}.

\bibitem{faa-2021}
{The U.S. Federal Aviation Administration (FAA)}, ``{Fact Sheet -- FAA
  Forecast--Fiscals Years 2016-37}.''
  \url{https://www.faa.gov/news/fact_sheets/news_story.cfm?newsId=21514}.

\bibitem{FAA:online}
``Faa aerospace forecasts.''
  \url{https://www.faa.gov/data_research/aviation/aerospace_forecasts/}.

\bibitem{Airsim_paper}
S.~Shah, D.~Dey, C.~Lovett, and A.~Kapoor, ``Airsim: High-fidelity visual and
  physical simulation for autonomous vehicles,'' {\em CoRR},
  vol.~abs/1705.05065, 2017.

\bibitem{yolo16}
J.~Redmon and A.~Farhadi, ``{YOLO9000:} better, faster, stronger,'' {\em CoRR},
  vol.~abs/1612.08242, 2016.

\bibitem{hog}
N.~Dalal and B.~Triggs, ``Histograms of oriented gradients for human
  detection,'' in {\em IEEE Computer Society Conference on Computer Vision and
  Pattern Recognition (CVPR)}, 2005.

\bibitem{octomap}
A.~Hornung, K.~M. Wurm, M.~Bennewitz, C.~Stachniss, and W.~Burgard,
  ``{OctoMap}: An efficient probabilistic {3D} mapping framework based on
  octrees,'' {\em Autonomous Robots}, 2013.
\newblock Software available at \url{http://octomap.github.com}.

\bibitem{ompl}
``Ompl.'' \url{https://ompl.kavrakilab.org/}.

\bibitem{orbslam2}
R.~Mur-Artal and J.~D. Tard\'os, ``{ORB-SLAM2}: an open-source {SLAM} system
  for monocular, stereo and {RGB-D} cameras,'' {\em IEEE Transactions on
  Robotics}, vol.~33, no.~5, pp.~1255--1262, 2017.

\bibitem{vins-mono}
T.~Qin, P.~Li, and S.~Shen, ``Vins-mono: A robust and versatile monocular
  visual-inertial state estimator,'' {\em arXiv preprint arXiv:1708.03852},
  2017.

\bibitem{dji-matrice}
``Matrice 100.'' \url{https://www.dji.com/matrice100/info/}.

\bibitem{solo3DR}
``3dr robotics.'' \url{https://3dr.com/solo-drone/}.

\bibitem{Zhang2017207}
A.~Vega, P.~Bose, and A.~Buyuktosunoglu, ``Reliable electrical systems for
  micro aerial vehicles and insect-scale robots: Challenges and progress,'' in
  {\em Rugged Embedded Systems}, Morgan Kaufmann, 2017.

\bibitem{GameEngi70:online}
``Game engine technology by unreal.''
  \url{https://www.unrealengine.com/en-US/what-is-unreal-engine-4}.

\bibitem{PhysicsS8:online}
``Physics simulation | unreal engine.''
  \url{https://docs.unrealengine.com/latest/INT/Engine/Physics/}.

\bibitem{Airsim:online}
``Microsoft/airsim: Open source simulator based on unreal engine for autonomous
  vehicles from microsoft ai \& research.''
  \url{https://github.com/Microsoft/AirSim}.

\bibitem{Pixhawk:online}
``Pixhawk flight controller.'' \url{https://pixhawk.org/}.

\bibitem{PX4Archi7:online}
``Px4 architectural overview - px4 developer guide.''
  \url{https://dev.px4.io/en/concept/architecture.html}.

\bibitem{mavlinkm68:online}
``Mavlink: Micro air vehicle protocol.'' \url{https://github.com/mavlink}.

\bibitem{TX2}
``Embedded systems developer kits, modules, \& sdks | nvidia jetson.''
  \url{https://www.nvidia.com/en-us/autonomous-machines/embedded-systems-dev-kits-modules/}.

\bibitem{joule:online}
``Iot.'' \url{https://software.intel.com/en-us/intel-joule-getting-started}.

\bibitem{ROSorgPo80:online}
``Ros.org | powering the world's robots.'' \url{http://www.ros.org/}.

\bibitem{energyaware}
C.~D. Franco and G.~Buttazzo, ``Energy-aware coverage path planning of uavs,''
  in {\em IEEE International Conference on Autonomous Robot Systems and
  Competitions (ICARSC)}, 2015.

\bibitem{3DR-energy-model}
C.~Tseng, C.~Chau, K.~M. Elbassioni, and M.~Khonji, ``Flight tour planning with
  recharging optimization for battery-operated autonomous drones,'' {\em CoRR},
  vol.~abs/1703.10049, 2017.

\bibitem{coulomb-counter}
K.~S.~R. Kumar, V.~V. Sastry, O.~C. Sekhar, D.~K. Mohanta, D.~Rajesh, and
  M.~P.~C. Varma, ``Design and fabrication of coulomb counter for estimation of
  soc of battery,'' in {\em IEEE International Conference on Power Electronics,
  Drives and Energy Systems (PEDES)}, 2016.

\bibitem{battery-model}
M.~Chen and G.~A. Rincon-Mora, ``Accurate electrical battery model capable of
  predicting runtime and i-v performance,'' {\em in IEEE Transactions on Energy
  Conversion}, vol.~21, pp.~504--511, June 2006.

\bibitem{unrealcv}
W.~Qiu and A.~Yuille, ``Unrealcv: Connecting computer vision to unreal
  engine,'' in {\em Computer Vision--ECCV 2016 Workshops}, Springer.

\bibitem{Handbook_robotic}
B.~Siciliano and O.~Khatib, {\em Springer Handbook of Robotics}.
\newblock Secaucus, NJ, USA: Springer-Verlag New York, Inc., 2007.

\bibitem{tech_problem}
M.~Benallegue, J.-P. Laumond, and N.~Mansard, {\em Robot Motion Planning and
  Control: Is It More than a Technological Problem?}
\newblock 05 2017.

\bibitem{kcf-c++}
J.~Faro, C.~Bailer, and J.~F. Henriques, ``Project title.''
  \url{https://github.com/charlespwd/project-title}, 2013.

\bibitem{kcf}
J.~F. Henriques, R.~Caseiro, P.~Martins, and J.~Batista, ``High-speed tracking
  with kernelized correlation filters,'' {\em IEEE Transactions on Pattern
  Analysis and Machine Intelligence}, vol.~37, no.~3, 2015.

\bibitem{rrt}
S.~M. LaValle, ``Rapidly-exploring random trees: A new tool for path
  planning,'' 1998.

\bibitem{prm}
L.~E. Kavraki, P.~Svestka, J.-C. Latombe, and M.~H. Overmars, ``Probabilistic
  roadmaps for path planning in high-dimensional configuration spaces,'' {\em
  IEEE transactions on Robotics and Automation}, vol.~12, no.~4, pp.~566--580,
  1996.

\bibitem{astar}
P.~E. Hart, N.~J. Nilsson, and B.~Raphael, ``A formal basis for the heuristic
  determination of minimum cost paths,'' {\em IEEE transactions on Systems
  Science and Cybernetics}, vol.~4, no.~2, pp.~100--107, 1968.

\bibitem{nbvplanner}
A.~Bircher, M.~Kamel, K.~Alexis, H.~Oleynikova, and R.~Siegwart, ``Receding
  horizon "next-best-view" planner for 3d exploration,'' in {\em IEEE
  International Conference on Robotics and Automation (ICRA)}, 2016.

\bibitem{high-speed-nav}
S.~Liu, M.~Watterson, S.~Tang, and V.~Kumar, ``High speed navigation for
  quadrotors with limited onboard sensing,'' in {\em IEEE International
  Conference on Robotics and Automation (ICRA)}, 2016.

\bibitem{eLoggerV4}
``Eagle tree systems.''
  \url{http://www.eagletreesystems.com/index.php?route=product/product&product_id=54}.

\bibitem{agyapong2014design}
P.~K. Agyapong, M.~Iwamura, D.~Staehle, W.~Kiess, and A.~Benjebbour, ``Design
  considerations for a 5g network architecture,'' {\em IEEE Communications
  Magazine}, vol.~52, no.~11, pp.~65--75, 2014.

\bibitem{gupta2015survey}
A.~Gupta and R.~K. Jha, ``A survey of 5g network: Architecture and emerging
  technologies,'' {\em IEEE access}, vol.~3, pp.~1206--1232, 2015.

\bibitem{reli-1}
R.~Remenyte-Prescott, J.~D. Andrews, and P.~W.~H. Chung, ``An efficient phased
  mission reliability analysis for autonomous vehicles,'' {\em Reliability
  Engineering \& System Safety}, vol.~95, no.~3, pp.~226--235, 2010.

\bibitem{reli-2}
P.~Fernandes and U.~Nunes, ``Platooning with dsrc-based ivc-enabled autonomous
  vehicles: Adding infrared communications for ivc reliability improvement,''
  in {\em IEEE Intelligent Vehicles Symposium (IV)}, 2012.

\bibitem{reli-3}
N.~Kalra and S.~M. Paddock, ``Driving to safety: How many miles of driving
  would it take to demonstrate autonomous vehicle reliability?,'' {\em
  Transportation Research Part A: Policy and Practice}, vol.~94, 2016.

\bibitem{reli-4}
M.~Alam, M.~Song, S.~Hester, and T.~Seliga, ``Reliability analysis of
  phased-mission systems: a practical approach,'' in {\em IEEE Reliability and
  Maintainability Symposium (RAMS)}, 2006.

\bibitem{mukherjee2005soft}
S.~S. Mukherjee, J.~Emer, and S.~K. Reinhardt, ``The soft error problem: An
  architectural perspective,'' in {\em IEEE High-Performance Computer
  Architecture (HPCA)}, 2005.

\bibitem{gazebo}
N.~Koenig and A.~Howard, ``Design and use paradigms for gazebo, an open-source
  multi-robot simulator,'' {\em IEEE International Conference on Intelligent
  Robots and Systems (IROS)}, 2004.

\bibitem{boss-cmu}
Urmson {\em et~al.}, ``Autonomous driving in traffic: Boss and the urban
  challenge,'' {\em AI magazine}, vol.~30, no.~2, p.~17, 2009.

\bibitem{talos-mit}
J.~Leonard, D.~Barrett, J.~How, S.~Teller, M.~Antone, S.~Campbell, A.~Epstein,
  G.~Fiore, L.~Fletcher, E.~Frazzoli, {\em et~al.}, ``Team mit urban challenge
  technical report,'' 2007.

\bibitem{odin-vtech}
Bacha {\em et~al.}, ``Odin: Team victortango's entry in the darpa urban
  challenge,'' {\em Journal of field Robotics}, vol.~25, no.~8, pp.~467--492,
  2008.

\bibitem{uber-simulator}
{Xioji Chen}, ``{Engineering Uber's Self-Driving Car Visualization Platform for
  Web}.'' \url{https://eng.uber.com/atg-dataviz/}.

\bibitem{uav_benchmark_simulator}
M.~Mueller, N.~Smith, and B.~Ghanem, ``A benchmark and simulator for uav
  tracking,'' in {\em Proc. of the European Conference on Computer Vision
  (ECCV)}, 2016.

\bibitem{flight-goggles}
Sayre-McCord {\em et~al.}, ``Visual-inertial navigation algorithm development
  using photorealistic camera simulation in the loop,'' in {\em IEEE
  International Conference on Robotics and Automation (ICRA)}, 2018.

\bibitem{slambench}
Nardi {\em et~al.}, ``Introducing slambench, a performance and accuracy
  benchmarking methodology for slam,'' in {\em IEEE International Conference on
  Robotics and Automation (ICRA)}, IEEE, 2015.

\bibitem{common-road}
M.~Althoff, M.~Koschi, and S.~Manzinger, ``Commonroad: Composable benchmarks
  for motion planning on roads,'' in {\em IEEE Intelligent Vehicles Symposium
  (IV)}, 2017.

\end{thebibliography}

\end{document}